\pdfoutput=1

\documentclass[11pt]{article}

\usepackage[]{acl}
\usepackage[acronym]{glossaries}
\usepackage{graphicx}
\usepackage{times}
\usepackage{latexsym}
\usepackage[T1]{fontenc}

\usepackage[utf8]{inputenc}

\usepackage{microtype}

\title{Fairness and Bias in Multimodal \acrshort{ai}: A Survey}

\author{First Author \\
  Affiliation / Address line 1 \\
  Affiliation / Address line 2 \\
  Affiliation / Address line 3 \\
  \texttt{email@domain} \\\And
  Second Author \\
  Affiliation / Address line 1 \\
  Affiliation / Address line 2 \\
  Affiliation / Address line 3 \\
  \texttt{email@domain} \\}

\author{\\
Tosin Adewumi\textsuperscript{*$\dag$}, Lama Alkhaled\textsuperscript{$\dag$}, Namrata Gurung\textsuperscript{1}, Goya van Boven\textsuperscript{2}, Irene Pagliai\textsuperscript{3}  \\
\\
\textsuperscript{$\dag$}Machine Learning Group, LTU, Sweden, 
\textsuperscript{1}QualityMinds GmbH, Germany, \\
\textsuperscript{2}Utrecht University, the Netherlands,
\textsuperscript{3}University of Göttingen, Germany, \\
\\
\textsuperscript{$\dag$}firstname.lastname@ltu.se,
\textsuperscript{1}namrata.gurung@qualityminds.de \\
\textsuperscript{2}j.g.vanboven@students.uu.nl,
\textsuperscript{3}irene.pagliai@uni-goettingen.de
\\
}

\makeglossaries

\newacronym{ml}{ML}{Machine Learning}
\newacronym{llm}{LLM}{Large Language Model}
\newacronym{sota}{SotA}{state-of-the-art}
\newacronym{iaa}{IAA}{inter-annotator agreement}
\newacronym{nlp}{NLP}{natural language processing}
\newacronym{ai}{AI}{artificial intelligence}
\newacronym{llama}{LLaMA}{Large Language Model Meta AI}
\newacronym{bert}{BERT}{Bidirectional Encoder Representations from Transformers}
\newacronym{svm}{SVM}{support vector machine}
\newacronym{gru}{GRU}{gated recurrent unit}
\newacronym{cnn}{CNN}{convolutional neural network}
\newacronym{t5}{T5}{Text-to-Text Transfer Transformer}
\newacronym{shi}{SHI}{Simple Hallucination Index}
\newacronym{gai}{GenAI}{generative artifical intelligence}
\newacronym{roberta}{RoBERTa}{Robustly optimized BERT pretraining Approach}
\newacronym{al}{AL}{Active Learning}
\newacronym{hf}{HF}{HuggingFace}
\newacronym{lmm}{LMM}{Large Multimodal Model}
\newacronym{mab}{MAB}{Multi-Axes Bias Dataset}
\newacronym{yfcc}{YFCC100M}{Yahoo Flickr Creative Commons 100 Million}
\newacronym{clip}{CLIP}{Contrastive Language-Image Pretraining}
\newacronym{sobem}{SOBEM}{Sexual OBjectification and EMotion Database}
\newacronym{eats}{EATs}{Embedding Association Tests}
\newacronym{wit}{WIT}{WebImageText}
\newacronym{grad}{GRAD-CAM}{Gradient-weighted Class Activation Mapping}
\newacronym{mimic}{MIMIC-IV}{Medical Information Mart for Intensive Care v. IV}
\newacronym{cleam}{CLEAM}{CLassifier Error-Aware Measurement}
\newacronym{sa}{SA}{Sensitive Attribute}
\newacronym{slip}{SLIP}{Semi-supervised Language-Image Pretraining}
\newacronym{blip}{BLIP}{Bootstapping Language-Image Pretraining}
\newacronym{coco}{MS COCO}{Microsoft Common Objects in Context}
\newacronym{cfd}{CFD}{Chicago Face Database}
\newacronym{us}{US}{United States}
\newacronym{owtc}{OWTC}{OpenWebText Corpus}
\newacronym{roots}{ROOTS}{Responsible Open-science Open-collaboration Text Sources}
\newacronym{bloom}{BLOOM}{BigScience Large Open-science Open-access Multilingual}
\newacronym{aurora}{Aurora-M}{Aurora-Multilingual}
\newacronym{pali}{PaLI}{Pathways Language and Image}
\newacronym{webli}{WebLI}{Web Language and Image}
\newacronym{llava}{LLaVA}{Large Language and Vision Assistant}
\newacronym{hcr}{HCR}{Hate Content Rate}
\newacronym{nsfw}{NSFW}{Not Safe For Work}
\newacronym{weat}{WEAT}{Word Embedding Association Test}
\newacronym{seat}{SEAT}{Sentence Embedding Association Test}
\newacronym{ripa}{RIPA}{Relational Inner Product Association}
\newacronym{ainlp}{A-INLP}{Autoregressive Iterative Nullspace Projection}
\newacronym{cda}{CDA}{Counterfactual Data Augmentation}
\newacronym{vqganc}{VQGAN-CLIP}{VQGAN-CLIP}
\newacronym{neurips}{NeurIPS}{Conference on Neural Information Processing Systems}
\newacronym{pkp}{PKP}{Public Knowledge Project}
\newacronym{mit}{MIT Press}{Massachusetts Institute of Technology Press}
\newacronym{jmlr}{JMLR}{Journal of Machine Learning Research}
\newacronym{laion}{LAION}{Large-scale Artificial Intelligence Open Network}
\newacronym{wikt}{WIT}{Wikipedia-based Image Text}
\newacronym{gpt}{GPT}{Generative Pretrained Transformer}
\newacronym{auc}{AUC}{Area Under the Curve}
\newacronym{bbq}{BBQ}{Bias Benchmark for QA}
\newacronym{cot}{CoT}{Chain of Thought}
\newacronym{wos}{WoS}{Web of Science}
\newacronym{acm}{ACM}{Association for Computing Machinery}
\newacronym{aaai}{AAAI}{Association of Advancement Artificial Intelligence}
\newacronym{ieee}{IEEE}{Institute of Electrical and Electronics Engineers}
\newacronym{mdpi}{MDPI}{Multidisciplinary Digital Publishing Institute}
\newacronym{plos}{PLoS}{Public Library of Science}
\newacronym{ameep}{EPODDER}{Association of Measurement and Evaluation in Education and Psychology}
\newacronym{nyusl}{NYUSL}{New York University School of Law}
\newacronym{acpi}{ACPI}{Academic Conferences and Publishing International}
\newacronym{aiaf}{AIAF}{AI Access Foundation}
\newacronym{hep}{HE Press}{Higher Education Press}
\newacronym{acl}{ACL}{Association of Computational Linguistics}
\newacronym{iat}{IAT}{Implicit Association Test}
\newacronym{ect}{ECT}{Embedding Coherence Test}
\newacronym{eqt}{EQT}{Embedding Quality Test}
\newacronym{aif}{AIF360}{AI Fairness 360}
\newacronym{cbow}{CBoW}{Continuous Bag-of-Words}
\newacronym{mt}{MT}{Machine Translation}
\newacronym{dam}{DAM}{Debiasing with Adapter Modules}
\newacronym{adele}{ADELE}{Adapter-based DEbiasing of LanguagE Models}
\newacronym{dear}{DeAR}{Debiasing with Additive Residuals}
\newacronym{mlrp}{MLRP}{Machine Learning Research Press}
\newacronym{geep}{GEEP}{GEnder Equality Prompt}
\newacronym{spie}{SPIE}{Society of Photographic Instrumentation Engineers}

\begin{document}
\maketitle
\begin{abstract}
The importance of addressing fairness and bias in \acrfull{ai} systems cannot be over-emphasized.
Mainstream media has been awashed with news of incidents around stereotypes and other types of bias in many of these systems in recent years.
In this survey, we fill a gap with regards to the relatively minimal study of fairness and bias in Large Multimodal Models (\acrshort{lmm}s) compared to Large Language Models (\acrshort{llm}s), providing \textbf{50 examples} of datasets and models related to both types of \acrshort{ai} along with the challenges of bias affecting them.
We discuss the less-mentioned category of mitigating bias, preprocessing (with particular attention on the first part of it, which we call \textit{preuse}).
The method is less-mentioned compared to the two well-known ones in the literature: intrinsic and extrinsic mitigation methods.
We critically discuss the various ways researchers are addressing these challenges.
Our method involved two slightly different search queries on two reputable search engines, \textit{Google Scholar} and \textit{\acrfull{wos}}, which revealed that for the queries `\textit{Fairness and bias in Large Multimodal Models}' and `\textit{Fairness and bias in Large Language Models}', 33,400 and 538,000 links are the initial results, respectively, for Scholar while 4 and 50 links are the initial results, respectively, for \acrshort{wos}.
For reproducibility and verification, we provide links to the search results and the citations to all the final reviewed papers.
We believe this work contributes to filling this gap and providing insight to researchers and other stakeholders on ways to address the challenges of fairness and bias in multimodal and language \acrshort{ai}.
\end{abstract}

\section{Introduction}
\label{intro}
\label{intro}

Fairness and bias are very important topics that cut across many domains in the society.
The rapid advancements in the research and applications of \acrfull{ai} have made them even more compelling in recent times, such that many studies have emerged on them \cite{frankel2020fair,booth2021bias,adewumi2022state,teo2024measuring}.
One important gap in the literature, however, is that there is relatively minimal study or survey on `Fairness and bias in Large Multimodal Models.'
By multimodal \acrshort{ai}, we mean the datasets or \acrshort{ai} models that can take one or more modalities as input and/or another as output.
This gap is evidenced by the fact that there are fewer works around the topic.
For example, a query search on \textit{Google Scholar} returns 33,400 links compared to 538,000 links for `Fairness and bias in Large Language Models' (where the first query search is equivalent to the boolean operation \textit{fairness} \textit{AND} \textit{and} \textit{AND} \textit{bias} \textit{AND} \textit{in} \textit{AND} \textit{large} \textit{AND} \textit{multimodal} \textit{AND} \textit{models}).\footnote{July 10, 2024}
This implies more than 16 times the result compared to the former.
However, filtering the publication year range to 2014-2024 reduces the links to 17,200 and 19,300, respectively.
We intend to contribute in filling that gap in this work.

The two terms, fairness and bias, are strongly related but fairness is concerned with equality and justice while bias is concerned with systematic error, which may arise from human prejudices \cite{booth2021bias,ALKHALED2023100030}.
For the purpose of this survey, fairness may be defined as \textit{equal representation} with regards to a given \acrfull{sa} \cite{hutchinson201950,frankel2020fair}.
Hence, we may consider a \acrfull{gai} to be fair if it generates both male and female samples with equal probabilities, with regards to the \acrshort{sa} gender \cite{teo2024measuring}.
Bias is a (non-random) systematic error in a measurement resulting in a difference in accuracy in one entity compared to another, given the ground truth \cite{booth2021bias,scheuneman1979method}.
We acknowledge there are other quantitative definitions of fairness and bias, as noted by \citet{hutchinson201950} and \citet{weidinger2022taxonomy}.

It appears the emergence of big data, which has brought rapid advancement in the \acrfull{sota}, also brought along the increase in poor quality content and prediction, such as the increased criminal prediction for Black and Latino people observed by \citet{birhane2024dark}.
Similar issues are observed across many domains, including healthcare, employment, forensics, criminal justice, credit scoring, and computational social science, among others \cite{liang2021towards,ferrara2023fairness,landers2023auditing,han2023fairness}.
According to \citet{wolfe2023contrastive}, the model \acrshort{vqganc}, similarly to Stable Diffusion, generated sexualized images for the harmless prompt "\textit{a 17 year old girl}," 73\% of the time.
The comparison to a similar prompt with the term "\textit{girl}" replaced with "\textit{boy}" shows a sharp contrast.

In \textbf{related work}, many of the relatively recent surveys on fairness and bias in \acrshort{ai} appear to have been of a general nature or focused on other areas.
\citet{pagano2023bias} focused their attention on \acrshort{ml} generally and the 5-year period between 2017 and 2022, thereby missing the nuances and some of the details related to \acrfull{nlp} and multimodal \acrshort{ai}.
\citet{mehrabi2021survey} surveyed applications that have exhibited bias in different domains, listed sources of bias in these applications and created a taxonomy for fairness definitions.
On the other hand, \citet{le2022survey} paid attention to benchmark tabular datasets for fairness, analysing relationships between different protected and class attributes.
\citet{balayn2021managing} focused on data bias in data engineering and management research, arguing for the enforcement of fairness requirements and constraints on the data for training and evaluating systems.
Their survey method is limited in that it restricted part of its literature search between 2019 and 2020.
In our work, in addition to discussing datasets that make \acrshort{ai} models biased and the datasets for evaluating bias and fairness, we discuss other important related areas, such as the mitigation strategies.

\citet{blodgett-etal-2020-language} surveyed over 140 articles about bias in \acrshort{nlp} and realized that the stated motivations are usually inconsistent and vague, and the articles do not engage with the broader applicable literature that is external to \acrshort{nlp}.
They, therefore, made 3 recommendations for those in this field: (1) establish the relationships in language and social hierarchies by referring to the broader literature outside of \acrshort{nlp}, (2) be explicitly clear on why the system described as biased is harmful, and (3) evaluate the language of people affected by the biased systems.
Their survey was restricted to only articles on text-based \acrshort{nlp}, thereby excluding speech or multimodal \acrshort{ai}, and limited to articles before May 2020.
However, we follow their recommendations and our work answers some of the research questions identified in their work.
For example, Section \ref{fairbias} and subsection \ref{curate} address the question \textit{How are datasets collected?}.
The survey by \citet{sun-etal-2019-mitigating} focused on recognizing and mitigating \acrshort{nlp} gender bias and the authors discussed this based on four forms of representation bias: denigration, stereotyping, recognition, and under-representation.
In \citet{haltaufderheide2024ethics}, they identified ethical issues around fairness and bias with \acrshort{llm}s in medicine and healthcare.
Additional works that have surveyed fairness and bias in \acrshort{llm}s include \citet{bender2021dangers,meade-etal-2022-empirical,gallegos2023bias,chang2024survey,myers2024foundation}.

In view of the foregoing gap and challenges, this work critically surveys the literature with the \textbf{primary objective} of ascertaining what the state of work is on \textit{fairness and bias in multimodal \acrshort{ai}}, thereby making the following key contributions.

\begin{enumerate}
    \item We fill the gap with a comprehensive survey of fairness and bias across a wide spectrum of \acrshort{lmm}s, \acrshort{llm}s, and multimodal datasets, providing 50 examples of datasets and models in a structured way, along with the challenges of bias affecting them.

    \item We discuss the less-mentioned category of mitigating bias in \acrshort{nlp} (i.e. preprocessing), though more common in general \acrshort{ml} \cite{mehrabi2021survey,pagano2023bias}, with particular attention on the first part of it, which we call \textit{preuse}.
    The other two well-known mitigation methods in the literature are the intrinsic (or in-processing) and extrinsic (or post-processing) methods \cite{ramesh-etal-2023-fairness,cabello-etal-2023-evaluating}.
    
    \item We critically discuss many important approaches to addressing the challenges of fairness and bias.
    
\end{enumerate}

The rest of this paper is organised as follows.
In the following Section (\ref{fairnessand}), we highlight some of the theories around fairness and bias in \acrshort{ai}.
In Section \ref{method}, we explain the method used for this survey.
In Section \ref{fairbias}, we focus on various works along the two paradigms of \acrshort{lmm}s and \acrshort{llm}s, discussing the datasets and models in the literature.
In Section \ref{discuss}, we discuss widely on the methods to evaluate fairness and bias, the datasets for evaluation, and the debiasing strategies.
We conclude the survey in Section \ref{conclude} with a summary and possible future work.

\section{Fairness and Bias in \acrshort{ai}}
\label{fairnessand}

\subsection{Implicit and Explicit Bias}
There is distinction between implicit and explicit bias such that, given two sets of terms that express the bias axis (e.g. gender or race), one set of male gender terms could be $\{S_{1}\}$ = \{dad, man\} and the other set of female terms be $\{S_{2}\}$ = \{mum, woman\}.
Implicit bias is sufficiently specified with both lists.
However, explicit bias requires two additional sets of attributes $\{A_{1}\}$ = \{engineer, doctor\} and $\{A_{2}\}$ = \{caregiver, carer\} that express the terms to which the earlier gendered terms exhibit association, albeit to different levels.
Hence, a gender biased system could result in male terms in $\{S_{1}\}$ being strongly associated with attributes of career terms $\{A_{1}\}$ compared to female terms $\{S_{2}\}$, which could be strongly associated with attributes of home-related terms $\{A_{2}\}$
\cite{friedrich-etal-2021-debie}.

Besides these two broad distinction of bias, there are many types of bias, depending on the philosophical or social perspectives that may be taken.
Hence, a discussion about every possible type of fairness or bias is beyond the scope of this study but we refer readers to \citet{mehrabi2021survey}, \citet{van2024undesirable} and \citet{navigli2023biases} for some of the many types that may be listed.

\subsection{Concepts of Fairness and Bias}

In this section, we highlight a few of the concepts around fairness and bias found in the Social Science literature.

\subsubsection{Justice Theory}
The theory is regarded as a three-part framework of  distributive, interactional, and procedural justice perceptions  \cite{greenberg1990organizational,landers2023auditing}.
Distributive justice, when outcomes are expected to be distributed equally, is the overarching aspect related to \acrshort{ai} fairness and bias, according to \citet{landers2023auditing}.
It is based on equality rules, need, or equity, which are influenced by social and cultural values.

\subsubsection{Equity Theory}
Equity theory uses a unidimensional concept of fairness instead of  multidimensional, according to \citet{leventhal1980should}.
It perceives justice solely on the merit principle and the final distribution of reward (or punishment), where reward is proportional to contribution \cite{adams1976equity}.

\subsubsection{Objectification Theory} Just as inanimate objects have no emotions or thoughts, objectification theory establishes a view of a subject as primarily without human characteristics, especially for women and girls \cite{fredrickson1997objectification,heflick2011women,andrighetto2019now}. The theory identifies sexual objectification bias, which is when the emotions or thoughts of a person are disregarded and one is treated as mere body parts for sex \cite{fredrickson1997objectification,wolfe2023contrastive}.

\subsection{Consequences of Bias}

\citet{fredrickson1997objectification} confirms that objectification victimises the subject and may result in habitual body monitoring, thereby increasing mental health risks, sexual dysfunction, eating disorders, and depression.
This unhealthy reality is also confirmed by \citet{swim2001everyday}.
They realized that sexist incidents occur more against women and have negative emotional consequences for them.
Some of these incidents are traditional gender role stereotypes, degrading remarks, and sexual objectification, which are found in the data used for training \acrshort{ai} models.
For details about the mechanics of training language models, we refer readers to \citet{radford2019language} and \citet{hoffmann2022empirical}.
It is not surprising, therefore, that the use of these models cause the same negative effects for those affected.
Hence, fairness and bias are not only ethical or moral issues but have legal implications \cite{landers2023auditing}.
In the \acrfull{us}, disparate treatment because of sensitive attributes is unlawful \cite{berry2015differential,hutchinson201950,meng2022interpretability}.
This is also the case in some other countries \cite{zafar2017fairness}.

\section{Methodology}
\label{method}

In order to have a fair and thorough survey for the stated objective in Section \ref{intro}, we followed the general guidelines recommended for conducting a systematic literature review, which is founded on a rigorous and auditable methodology \cite{kitchenham2004procedures,brereton2007lessons}.
We used two common scientific search engines: \textit{Google Scholar} and \textit{\acrfull{wos}}.
Both are advantageous because they index the main literature databases or publishers, including the \acrfull{ieee}, \acrfull{acm}, \acrfull{neurips}, \acrfull{mdpi}, \acrfull{pkp}, \acrfull{mit}, \acrfull{plos}, \acrfull{acl}, \acrfull{aiaf}, \acrfull{hep}, \acrfull{ameep}, \acrfull{nyusl}, \acrfull{acpi}, \acrfull{mlrp}, \acrfull{jmlr}, and \acrfull{spie}.

\acrshort{wos} provides a second assessment to Scholar because of the limitations of the latter, including its inclusion of non-peer-reviewed links or possible predatory journals in its results of primary studies (or papers) \cite{foltynek2019academic}.
The two search engines are identified as largely reliable, helping to mitigate the risk of incompleteness in the search \cite{foltynek2019academic}.
For both, we employed a similar multi-phase search process itemized below, which helped to bring refinement to each search and the final results.
We are confident this approach returned the most applicable results for the purpose of this study.

\begin{enumerate}
    \item In the first phase, we designed the following two search queries based on our objective. 
    \begin{itemize}
        \item \textit{Fairness and Bias in Large Multimodal Models}
        \item \textit{Fairness and Bias in Large Language Models}
    \end{itemize}

    The boolean operation, in both search engines, for the first is equivalent to \textit{fairness} \textit{AND} \textit{and} \textit{AND} \textit{bias} \textit{AND} \textit{in} \textit{AND} \textit{large} \textit{AND} \textit{multimodal} \textit{AND} \textit{models} while the second is \textit{fairness} \textit{AND} \textit{and} \textit{AND} \textit{bias} \textit{AND} \textit{in} \textit{AND} \textit{large} \textit{AND} \textit{language} \textit{AND} \textit{models}.
    
    \item In the second phase, we filtered the results based on the time period of just over 10 years (2014 to 2024) and relevance (after a review of paper titles and abstracts for over 200 papers).
    For example, one result involved an article about `\textit{executive coaching programs}' that has nothing to do with \acrshort{ai} or social bias.
    All the filtered papers are in English.
    In addition, we corrected for misplaced results (i.e. papers that turned up in one search result though they belong in the other; about 15 in multimodal belonging in language and 9 in language belonging in multimodal, for Scholar).
    We removed papers published in journals in \textit{Beall's List of Potential Predatory Journals and Publishers},\footnote{beallslist.net/standalone-journals} if present.

    \item In the third phase, we critically reviewed the papers for their contributions, including the datasets, models, possible solutions proffered on fairness and bias, and other relevant discussions.

\end{enumerate}

More concretely, for \acrshort{wos}, we searched `\textit{All Fields}' of the documents `\textit{Core Collection}' across all `\textit{Editions}' without initially restricting the year of publication.
For Scholar, the first phase returned 33,400 and 538,000 result links over many pages for the first and second queries, respectively, while returning 4 and 50 for \acrshort{wos}.\textsuperscript{1}
Filtering Scholar, based on the time period, returned 17,200 and 19,300 links, which finally reduced to 69 and 101 at the end of the second phase for the first and second terms, respectively, while returning 8 and 44 for \acrshort{wos}.
It is noteworthy that for Scholar, there were equally 100 links to start with for both queries.
The boolean search operation involved all the individual words in any arbitrary order without case sensitivity and narrowed the search to mostly relevant documents.
We compare `Multimodal Models' with only `Language Models' in the search terms because the latter, with the introduction of the Transformer architecture \cite{vaswani2017attention}, have influenced computer vision \cite{yuan2021tokens} and they serve as important components for many multimodal \acrshort{ai}.
Furthermore, we recognize that multimodal implies the combination of `language' (or `speech') and `vision' but we did not use this distinction in our multimodal search because (1) the boolean equivalent can result in false positives, (2) even when we attempted it, we had less than 50\% links in result compared to the second term of only `language' search, and (3) `multimodal' is a standard term in the field.

\begin{table}[t!]
\centering
\caption{Distribution of scientific papers on Google Scholar. (\footnotesize{Filtered total is limited to the year range 2014 - 2024 and to the first 100 links per search query, excluding irrelevant links}.)}
\label{paperlist1}
\begin{tabular}{|p{0.027\linewidth} | p{0.22\linewidth} | p{0.25\linewidth} | p{0.25\linewidth}|}
\hline
\# &  \textbf{Publisher} &  \textbf{Multimodal} &  \textbf{Language} \\
\hline
\multicolumn{2}{|l|}{Unfiltered total} & 33,400 &  \textbf{538,000} \\
\hline
\hline
1 & \acrshort{ieee} & 10 & - \\
2 & Elsevier & 3  & 3 \\
3 & \acrshort{acm} & 14 &  13 \\
4 & Springer & 7 & 5 \\
5 & \acrshort{neurips} & 4 & 11 \\
6 & Nature & 3 & 5 \\
7 & \acrshort{mdpi} & 2 & 1 \\
8 & \acrshort{mlrp} & 2 & 4 \\
9 & \acrshort{pkp} & 1 & 1 \\
10 & \acrshort{mit} & - & 4\\
11 & PubMed & 1 & 1 \\
12 & Cambridge & - & 1 \\
13 & \acrshort{plos} & 1 & 1 \\
14 & \acrshort{jmlr} & - & 2\\
15 & \acrshort{acl} & 5 & 23 \\
16 & \acrshort{spie} & 1 &- \\
17 & De Gruyter & 1 & - \\
18 & Wiley & 1 & 1 \\
19 & Sage & 1 & - \\
20 & Academic Pinnacle & 1 & 1\\
\hline
& Filtered sub-total & 58 &  \textbf{77} \\
\hline
&  & & \\
21 & arXiv & 11 & 23 \\
22 & Preprints & - & 1 \\
\hline
& Filtered total & 69 & \textbf{101} \\
\hline
\hline
\end{tabular}
\end{table}

\begin{table}[t!]
\centering
\caption{Distribution of scientific papers on \acrfull{wos}. (\footnotesize{Filtered total is limited to the year range 2014 - 2024 and removes irrelevant results.})}
\label{paperlist2}
\begin{tabular}{|p{0.02\linewidth} | p{0.22\linewidth} | p{0.23\linewidth} | p{0.22\linewidth}|}
\hline
\# &  \textbf{Publisher} & \textbf{Multimodal} &  \textbf{Language} \\
\hline
\multicolumn{2}{|l|}{Unfiltered total} & 4 &  \textbf{50} \\
\multicolumn{2}{|l|}{Corrected  total} & 8 &  \textbf{46} \\
\hline
\hline
1 & \acrshort{ieee} & 4 & 5 \\
2 & Elsevier & -  & 2 \\
3 & \acrshort{acm} & 3 & 12 \\
4 & Springer & - & 5 \\
5 & \acrshort{neurips} & - & 1 \\
6 & Nature & - & 1 \\
7 & \acrshort{mdpi} & - & 3 \\
22  & \acrshort{mlrp} & - & 1 \\
22  & \acrshort{pkp} & 1 & - \\
16  & \acrshort{acl} & - & 10 \\
13  & Wiley & - & 2 \\
17  & \acrshort{aiaf} & - & 1 \\
12  & Now & - & 1 \\
\hline
& Filtered total & 8 &  \textbf{44} \\
\hline
\hline
\end{tabular}
\end{table}



Finally, for Scholar, we captured archived papers (e.g. arXiv) because, sometimes, their peer-reviewed versions exist but may not appear in the result.
The summary statistics of the search are presented in Tables \ref{paperlist1} and \ref{paperlist2} while the references to the papers and the link to the search results are provided in the Appendix.
The useful data from the reviewed papers (or primary studies) about datasets, \acrshort{ai} models, and other contributions are then discussed in Sections \ref{lmms}, \ref{datallms}, and \ref{discuss}, respectively.
We note that the more relevant papers turn up on the first page or at the top of the search while the quality of results degrade as one progresses through the pages or list.
From the Tables, it can be observed that there are fewer pair-reviewed papers on multimodal models compared to language models.

\section{Findings on Fairness and Bias in \acrshort{lmm}s and \acrshort{llm}s}
\label{fairbias}

It is commonly agreed that \acrshort{ai} models learn much of their bias from the data they are trained on and many datasets, especially those for pretraining, are from the Internet, which contains a diverse spectrum of content \cite{wolfe2022american}.
Tables \ref{datamodels} and \ref{table4} summarize some relevant datasets and the models beset by challenges of bias, respectively.
All the 25 datasets identified have their challenges and by extension the 25 \acrshort{ai} models which train on them.
Some of these challenges include stereotypes, porn, misogyny, racial, gender, religious, cultural, age, and demographic biases.

\subsection{\acrshort{lmm}s}
\label{lmms}

\begin{table*}[h!]
\small
\centering
\caption{Summary of Some Datasets and Their Fairness \& Bias Challenges (\footnotesize{Data in the lower part of the table are usually used in downstream tasks}).}
\label{datamodels}
\begin{tabular}{| p{0.02\linewidth} | p{0.28\linewidth} | p{0.13\linewidth} | p{0.46\linewidth}|}
\hline
\# & \textbf{Dataset} & \textbf{Modality} & \textbf{Some Challenges of Bias/Fairness}\\
\hline
1 & CommonCrawl \cite{raffel2020exploring} & Text \& Vision & Fake news, hate speech, porn \& racism \cite{gehman-etal-2020-realtoxicityprompts,luccioni2021s} \\
2 & \acrshort{laion}-400M \& 5B \cite{schuhmann2021laion,schuhmann2022laion} &  Text \& Vision & Misogyny, stereotypes \& porn \cite{birhane2021multimodal,birhane2024into} \\
3 & \acrfull{wit} \cite{radford2021learning} &  Text \& Vision &  Racial, gender biases \cite{radford2021learning}\\
 & & & \\
4 & DataComp \cite{gadre2024datacomp} & Text \& Vision & Racial bias \cite{gadre2024datacomp}\\
5 & \acrshort{webli} \cite{chen2022pali} & Text \& Vision & Age, racial, gender biases \& stereotypes \cite{chen2022pali}\\
6 & CC3M-35L \cite{thapliyal-etal-2022-crossmodal} & Text \& Vision & Cultural bias \cite{thapliyal-etal-2022-crossmodal}\\ 
7 & COCO-35L \cite{thapliyal-etal-2022-crossmodal} & Text \& Vision & Cultural bias \cite{thapliyal-etal-2022-crossmodal} \\
8 & \acrshort{wikt} \cite{10.1145/3404835.3463257} & Text \& Vision & Cultural bias \cite{10.1145/3404835.3463257} \\
& & & \\
9 & Colossal Cleaned CommonCrawl (C4) \cite{raffel2020exploring} & Text & Offensive language, racial bias \cite{raffel2020exploring} \\
10 & The Pile \cite{gao2020pile} & Text & Religious, racial, gender biases \cite{gao2020pile}\\
11 & CCAligned \cite{el-kishky-etal-2020-ccaligned} & Text &  Porn, racial bias \cite{el-kishky-etal-2020-ccaligned}\\
12 & OpenAI WebText \cite{radford2019language} & Text & Gender, racial biases \cite{gehman-etal-2020-realtoxicityprompts}\\
13 & \acrfull{owtc} & Text & Gender, racial biases \cite{gehman-etal-2020-realtoxicityprompts}\\
14 & \acrshort{roots} \cite{laurenccon2022bigscience} & Text & Cultural bias \cite{laurenccon2022bigscience} \\
\hline
\hline
 & & & \\
15 & VoxCeleb 1 \cite{nagrani2020voxceleb} & Audio \& Vision & Demographic, gender biases \cite{chung2018voxceleb2} \\
16 & VoxCeleb 2 \cite{chung2018voxceleb2} & Audio \& Vision & Demographic, gender biases \cite{chung2018voxceleb2} \\
17  & First Impressions \cite{escalante2020explaining} & Audio \& Vision & Racial, gender biases \cite{yan2020mitigating}\\
 & & & \\
 
18 & XM3600 \cite{thapliyal-etal-2022-crossmodal} & Text \& Vision & Cultural bias \cite{thapliyal-etal-2022-crossmodal}\\
19 & VQA \cite{antol2015vqa} & Text \& Vision & Gender bias \cite{ruggeri-nozza-2023-multi}\\
20 & VQA 2 \cite{goyal2017making} & Text \& Vision & Gender bias \cite{ruggeri-nozza-2023-multi} \\
21 & \acrshort{coco} \cite{lin2014microsoft} & Text \& Vision & Gender bias \cite{cabello-etal-2023-evaluating,zhao-etal-2017-men}\\
22  & Multi30K \cite{elliott-etal-2016-multi30k} & Text \& Vision & Racial bias \cite{wang-etal-2022-assessing}\\

23 & \acrshort{mimic} \cite{johnson2023mimic} & Text \& Vision & Ethnic, racial, marital status biases \cite{meng2022interpretability}\\
24  & \acrshort{mab} \cite{ALKHALED2023100030} & Text & Racism, misogyny, stereotypes \cite{ALKHALED2023100030,pagliai2024data}
 \\
25  & Twitter corpus \cite{huang-etal-2020-multilingual} & Text &  Age, gender, racial biases \cite{huang-etal-2020-multilingual} \\
\hline
\end{tabular}
\end{table*}

\citet{liang2021multibench} acknowledged that multimodal representations are challenging because they seek to integrate information from multiple areas in applications like robotics, finance, healthcare and more.
However, multimodal \acrshort{ai} has not enjoyed enough resources to study generalization across different modalities and the complexities of training.
With regards to bias, \citet{wolfe2023contrastive} found evidence of sexual objectification bias in models based on \acrfull{clip}.
The 9 \acrshort{clip} models that were investigated were trained on internet-wide web crawls.
\acrshort{clip} is known to be quite accurate \cite{radford2021learning}, however, it also appears to have scaled the biases inherent in its training data.
Also, \citet{wolfe2022american} found that more than Latino, Asian or Black, White persons are more associated with collective in-group words in embeddings from \acrshort{clip} \cite{radford2021learning}, \acrshort{slip} \cite{norman2022slip}, and \acrshort{blip} \cite{li2022blip}, as measured with \acrfull{eats}.
For a definitive assessment, their work would have benefited from additional experiments involving data of people outside the \acrfull{us}, since they used the \acrfull{cfd} \cite{ma2015chicago}.
\acrshort{cfd} is a dataset of 597 people recruited in the U.S.
Similarly, \citet{teo2024measuring} found that Stable Diffusion exhibits gender bias on slight changes to its prompts.

Besides the work on text and text-visual multimodal systems or data, there are some work on audio-visual systems or data \cite{fenu2022demographic}.
In the work by \citet{fenu2022demographic}, they perform comparative analysis on audio-visual speaker recognition systems, using fusion at the model step.
They found that the highest accuracy and the lowest disparity across groups are achieved compared to unimodal systems.

In other works, \citet{pena2023human} evaluated AI-based recruitment for multimodal data but in a fictitious case study, which may limit its generalizability in real-world applications.
\citet{booth2021bias} performed a case study of automated video interviews and found that combining more than one modality increases bias and reduces fairness, similarly to what happens when scaling crawled data for training models.
Other researchers investigated the impact of multimodal data/models on personlity assessment \cite{yan2020mitigating}, cyberbullying \cite{alasadi2020fairness}, health records \cite{meng2022interpretability} and more \cite{birhane2021multimodal,zhang2022counterfactually,ferrara2023fairness,cabello-etal-2023-evaluating}
The obvious challenges of bias in data has motivated some researchers for more attention in the ethics of data collection \cite{weinberg2022rethinking}.


\begin{table*}[h!]
\small
\centering
\caption{Summary of Models and Some of Their Fairness \& Bias Challenges. (\footnotesize{*modified for audio-visual})}
\label{table4}
\begin{tabular}{|p{0.02\linewidth} | p{0.19\linewidth} | p{0.13\linewidth} | p{0.15\linewidth} | p{0.38\linewidth}|}
\hline
\# & \textbf{\acrshort{lmm}s} & \textbf{Modality} & \textbf{Training Data}  & \textbf{Some Challenges of Bias/Fairness} \\
\hline
1 & VQGAN-CLIP \cite{crowson2022vqgan} & Text \& Vision & \acrshort{wit} \& ImageNet & Misogyny, racial bias \cite{pagliai2024data,wolfe2022american}\\
2 & DALL-E 2 \cite{ramesh2021zero}&  Text \& Vision & Conceptual Captions & Occupational stereotypes, gender bias \cite{ramesh2022hierarchical,mandal2023measuring} \\
3 & GLIDE \cite{pmlr-v162-nichol22a} & Text \& Vision & \acrshort{wit} \& Conceptual Captions & Gender stereotypes \cite{pmlr-v162-nichol22a} \\
4 & Stable Diffusion \cite{rombach2022high} & Text \& Vision &  \acrshort{laion}-5B & Gender bias \cite{teo2024measuring} \\
5 & \acrshort{slip} \cite{norman2022slip} & Text \& Vision & \acrshort{yfcc}  & Racial bias \cite{wolfe2022american}\\
 & & & & \\
6 & \acrshort{clip} \cite{radford2021learning} & Text \& Vision & \acrshort{wit} & Racial bias \cite{radford2021learning,wolfe2022american}\\
7 & \acrshort{blip} \cite{li2022blip} & Text \& Vision & \acrshort{coco}, Conceptual Captions \& \acrshort{laion}-400M & Racial, gender \& age biases \cite{wolfe2022american,ruggeri-nozza-2023-multi}\\
8 & PaliGemma & Text \& Vision & \acrshort{webli}, CC3M-35L , \acrshort{wikt} & Porn, offensive language\footnote{ai.google.dev/gemma/docs/paligemma/model-card}\\
9 & \acrshort{pali}-3 \cite{chen2022pali} & Text \& Vision & \acrshort{webli} & Age, racial, gender biases \& stereotypes \cite{chen2022pali} \\
 & & & & \\
10 & Falcon 2 \cite{almazrouei2023falcon} & Text \& Vision & RefinedWeb & Harmful content, cultural bias \cite{almazrouei2023falcon} \\
11 & BEiT \cite{wang2023image} & Text \& Vision  & Conceptual 12M, ImageNet-21K, Wikipedia & Gender, cultural biases \cite{brinkmann2023multidimensional} \\
12 & \acrshort{llava} \cite{liu2024visual,liu2024improved} & Text \& Vision  & Conceptual Captions & Cultural bias \cite{liu2024visual} \\
13 & ResNet-50* \cite{he2016deep} & Audio \& Vision & ImageNet & Gender bias \cite{fenu2022demographic}\\
14 & GPT4o \cite{achiam2023gpt} & Text, Audio \& Vision & WebText, Github, etc & Stereotypes, racial bias \cite{aich2024vernacular} \\
15 & GPT3 \cite{brown2020language} & Text & CommonCrawl \& WebText & Gender, racial, religious biases \cite{brown2020language,gehman-etal-2020-realtoxicityprompts} \\
16 & PaLM \cite{10.5555/3648699.3648939} & Text & Wikipedia, social media, Github & Occupation, gender, sexual, religious biases \cite{10.5555/3648699.3648939}
 \\
17 & LaMDA \cite{thoppilan2022lamda} & Text & Social media \& Wikipedia & Gender bias \cite{thoppilan2022lamda} \\
18 & GLaM \cite{du2022glam} & Text & Wikipedia \& social media & Toxicity, gender bias \cite{du2022glam}\\
 & & & & \\
19 & GPT2 \cite{radford2019language} & Text & WebText & Sexual, racial, gender biases \cite{sheng-etal-2019-woman,gehman-etal-2020-realtoxicityprompts}
 \\
20 & \acrshort{llama}-3 & Text & web text \& Github & Stereotypes, gender, racial, sexual, religious,  biases\cite{aich2024vernacular,touvron2023llama}\\
21 & \acrshort{llama}-2 \cite{touvron2023llama} & Text & web text & Toxicity, gender, racial, sexual, religious,  biases \cite{touvron2023llama}\\
22 & CTRL \cite{keskar2019ctrl} & Text & Wikipedia, Project Gutenberg, OpenWebText &  Gender, racial biases \cite{gehman-etal-2020-realtoxicityprompts} \\
 & & & & \\
23 & \acrshort{aurora} \cite{nakamura2024aurora} & Text & The Pile & Offensive language, religious, racial, gender biases \cite{gao2020pile,nakamura2024aurora} \\
24 & Mixtral-8x7B \cite{jiang2024mixtral} & Text & web text & Stereotypes, racial, gender, occupational biases \cite{jiang2024mixtral,aich2024vernacular}\\
25 & \acrshort{bloom} \cite{le2023bloom} & Text & \acrshort{roots} & Toxicity, gender, religious, disability, age biases \cite{le2023bloom}\\
\hline
\hline
\end{tabular}
\end{table*}

\subsection{\acrshort{llm}s}
\label{datallms}

The relevant datasets, models, and challenges of bias affecting \acrshort{llm}s, as discussed in the literature, are summarized in Tables \ref{datamodels} and \ref{table4}.
The introduction of the \acrlong{gpt} (\acrshort{gpt}-2) \cite{radford2019language} was a turning point in the language model landscape with its 1.5B parameters.
As pointed out earlier, training such a large model required a lot of data and the Internet-sourced WebText was used for this purpose.
Updated versions of the dataset have also been used for its recent successors, including \acrshort{gpt}-3 \cite{brown2020language} and \acrshort{gpt}-4 \cite{achiam2023gpt}.
Unfortunately, the attendant problems of bias witnessed in \acrshort{gpt}-2 \cite{gehman-etal-2020-realtoxicityprompts} have followed successive versions.
This is also the case with other models, as expressed in Table \ref{table4}, for the reasons that the datasets used in pretraining, some of which are given in Table \ref{datamodels}, are largely from Internet sources.
Furthermore, the architectures of the models, many of which are based on the Transformer architecture \cite{vaswani2017attention}, share similarity.

In other works, \citet{schramowski2022large} show that \textit{moral directions}, i.e. what is morally right or wrong to do, are present in \acrshort{llm}s.
It is, however, highly debatable if models can be considered moral agents.
In a collaborative effort, \citet{srivastava2023beyond} probed \acrshort{llm}s in the BIG-Bench of 200 tasks, including many that are related to bias.
Also, \citet{santurkar2023whose} explored whose opinions \acrshort{llm}s reflect while \citet{harrer2023attention,nashwan2023harnessing} investigated bias in \acrshort{llm}s within healthcare.

\subsection{Mitigation Categories}
Quantitative bias measurement and mitigation in \acrshort{nlp} may be placed into 3 categories: preprocessing, as applicable in general \acrshort{ml} \cite{hort2024bias}, 
intrinsic \cite{caliskan2017semantics,may-etal-2019-measuring}, and extrinsic, as observed by \citet{ramesh-etal-2023-fairness}. 
The first, second and third involve quantifying and mitigating bias in the training data, in the trained model's representation, and in the outputs of the downstream task of the model, respectively.
More work has focused on the latter two than the first and gender bias than other dimensions \cite{ramesh-etal-2023-fairness}.
For example, \citet{delobelle-etal-2022-measuring} and \citet{welbl-etal-2021-challenges-detoxifying}  measured bias in pre-trained language models.
One main reason why more work has been on the latter, according to \citet{sun-etal-2019-mitigating}, is that debiasing an existing model to adjust the outputs usually only requires `patching' the model instead of retraining with modified (debiased) or new data, which is usually costly.

\paragraph{Preuse} This may be considered the first step of the \acrshort{nlp} preprocessing method for debiasing.
It involves quantifying the amount of bias (or toxicity) in a given dataset by using a model, after the model has been trained on a different dedicated dataset, without necessarily attempting to mitigate the bias in the given dataset.
It can be defined by Equation \ref{preuse}, where $B(d)$ is a bias metric that takes data as input and returns a scalar, $s$, as the score.
Examples of works involving this are \cite{ALKHALED2023100030,adewumi-etal-2023-bipol}.
HateBERT \cite{caselli-etal-2021-hatebert} has been used in similar settings.
Equation \ref{preuse} is similar to that in \citet{brunet2019understanding} of \textit{differential bias}, where they approximate the effect of removing small parts of training data
on bias.
Gender bias is one example of the kind of bias that can be estimated, as it has been observed that there are more male than female terms in many \acrshort{nlp} datasets \cite{sun-etal-2019-mitigating,ALKHALED2023100030,pagliai2024data}.
The ability to first estimate quantitatively the bias in a given dataset provides the basis to be able to determine the level of success of mitigation.

\begin{equation}
\mathit{B{\left( d \right)}} = {s}
\label{preuse}
\end{equation}

\subsection{Bias in multilingual \acrshort{ai}:}

Some of the multimodal data in Table \ref{datamodels} contain multilingual data.
These result in multilingual models and embeddings.
For example, CC3M-35L, COCO-35L, and \acrshort{webli}, which was used to train \acrshort{pali}.
\acrshort{webli} is a mix of pre-training tasks with texts in 109 languages \cite{chen2022pali}.
Some of these languages fall in the category of low-resource languages \cite{adewumi2023afriwoz}.
\citet{kurpicz2020cultural} reports statistically significant bias in German word embeddings based on the origin of a name in relation to pleasant and unpleasant words using \acrshort{weat} \cite{caliskan2017semantics}.
In the study by \citet{wambsganss2022bias}, they found that the pretrained German language models, GermanBERT, GermanT5, and German \acrshort{gpt}-2, had substantial conceptual, racial, and gender bias.
This was also confirmed by \citet{kraft2022measuring}, who observed sexist stereotypes in some of the models (e.g. family- and care-related terms were associated with female while crime and perpetrators were associated with male).
Similarly, for Dutch, \citet{delobelle-etal-2020-robbert} investigated gender and occupation biases in RobBERT (a Dutch RoBERTa \cite{liu2019roberta}), through a template-based association test \cite{kurita-etal-2019-measuring,may-etal-2019-measuring}.
\citet{huang-etal-2020-multilingual} also identified biases related to people's origin and age in Italian, English, Polish, Portuguese, and Spanish, using a Twitter corpus.

\section{Discussion}
\label{discuss}

Although there are many limitations or risks of multimodal \acrshort{ai} or \acrshort{llm}s \cite{acosta2022multimodal,adewumi2024limitations,pettersson2024generative,adewumi2024instruction}, perhaps the issue of fairness and bias rank among the topmost \cite{mehrabi2021survey}.
In addition, in the taxonomy of 21 risks of language models provided by
\citet{weidinger2022taxonomy}, the first category is `\textit{Discrimination, Hate speech and Exclusion}'.
Given the recurring challenges in this regard, we are of the view that existing tools for evaluating or handling fairness and bias in these systems need to be improved \cite{zhao-etal-2018-gender,rudinger-etal-2018-gender}.
It may be almost impossible to automatically filter a dataset or debias a model to be 100\% free of unfair, bias or toxic content but the research community and other stakeholders may need to determine what levels are acceptable and if it should be a requirement to have human-in-the-loop methods.
In this section, we discuss methods to audit or evaluate fairness and bias, datasets for such evaluation, and debiasing strategies.
We hope that such discussion will spur more researchers and stakeholders to see the critical importance of \acrshort{ai} that is fair and free from bias, as much as possible. 

\subsection{Methods to audit, measure, and evaluate fairness and bias}

\citet{caliskan2017semantics} introduced the \acrfull{weat}, which is based on the \acrfull{iat} \cite{nosek2002harvesting}.
The \acrshort{iat} was designed to measure attitudes towards social groups.
It showed implicit preference for White and young people over Black and old people, respectively.
Furthermore, it showed the association of male terms with science while female terms were with family and arts.
\acrfull{eats} have been used in several studies \cite{kurpicz2020cultural,wolfe2023contrastive}
and adapted with improvements in \acrfull{seat} \cite{may-etal-2019-measuring} and \acrfull{ripa} \cite{ethayarajh-etal-2019-understanding}.
Despite its widespread use, \acrshort{weat} has the disadvantage that it may systematically overestimate the bias in a model.
In vision models, \citet{10.1145/3577190.3614156} used \acrshort{weat} to audit \acrshort{clip} by detecting and quantifying bias.
Also, along the lines of the \acrshort{weat}, \citet{pmlr-v89-dev19a} introduced the \acrfull{ect} and \acrfull{eqt} and proposed methods for eliminating explicit bias.
However, their method has the weakness that it is not able to remove implicit bias \cite{friedrich-etal-2021-debie}.

Another embedding evaluation method is cosine similarity.
It was used for zero-shot classification by \citet{radford2021learning}.
It may also be used to audit fairness and bias by evaluating the similarity in image and text embeddings \cite{wolfe2023contrastive}.
The visual tool, \acrfull{grad}, generates a saliency map, which shows the most relevant regions of an image for given attributes \cite{8237336,wolfe2023contrastive}.
In an evaluation carried out by \citet{wolfe2023contrastive}, they discovered that the computed average saliency maps included only face regions for non-objectified images but both face and chest regions for objectified images, in a possible indication of sexual objectification bias.
The tool \acrfull{nsfw} detector uses a tag alongside each image for filtering undesirable content \cite{schuhmann2021laion,birhane2024into}

A recent metric introduced by \citet{ALKHALED2023100030} is \textit{bipol}.
It uses a two-step procedure in estimating bias in data \cite{adewumi-etal-2023-bipol,pagliai2024data}.
Bipol has the weakness that if the bias classifier is not accurate enough, false positives will weaken the evaluation score.
Another measure is 
\acrfull{auc}, as used by \citet{meng2022interpretability} in the investigation of algorithmic fairness of mortality prediction, where they noted that \acrshort{ml} methods obtain lower scores, usually, when it involves groups with higher mortality rates.
\citet{teo2024measuring} proposed \acrfull{cleam}, a framework for better performance in bias estimation, while
\citet{booth2021bias} measured gender bias using accuracy of Spearman rank-based correlation ($\rho$).
Furthermore, \citet{nozza2021honest} introduced the score “\textit{HONEST}”, which was tested with respect to gender bias in text generation in 6 languages: Italian, French, Portuguese, Romanian, Spanish and English.
It measures the probability that a language model will output hurtful text given a certain template and lexicon.

\paragraph{Datasets for Bias Evaluation}

Different datasets have been introduced for bias evaluation.
\citet{liang2021multibench} introduced MultiBench, a unified multimodal benchmark that spans 15 datasets, 10 modalities, and 20 prediction tasks.
FairFace was introduced by \citet{karkkainen2021fairface}.
The dataset was designed to mitigate racial bias in multimodal \acrshort{ai}, collected from the \acrfull{yfcc} dataset \cite{thomee2016yfcc100m} and contains 108,501 images, balanced across the following races: Black, White, Indian, Southeast Asian, East Asian, Middle Eastern, and Latino.
\citet{esiobu-etal-2023-robbie} introduced 2 novel datasets \textit{AdvPromptSet} and \textit{HolisticBiasR}, with which they evaluated 12 demographic dimensions for different \acrshort{llm}s.
\citet{ruzzante2022sexual} introduced
\acrfull{sobem} for sexual objectification bias studies.
It consists of 280 pictures of objectified and non-objectified female models with 3 different emotions and a neutral face.
\acrfull{bbq} is a question-set dataset that employs templates crafted to reflect specific biases identified in society.
It was introduced by \citet{parrish-etal-2022-bbq} and aims to expose implicit prejudices that may exist against individuals from legally protected categories.

\textit{BEAVERTAILS} was introduced by \citet{ji2024beavertails}.
It assesses question-answer pairs, tested on \acrshort{llm}s, with regards to 14 different harm categories, which are not exhaustive.
`\textit{Discrimination, Stereotype, Injustice}' makes up the second category in the list.
Additional datasets for bias evaluation include \textit{RedditBias} by \citet{barikeri-etal-2021-redditbias}, \textit{RealToxicityPrompts}, which  comprises of 100K English sentences \cite{gehman-etal-2020-realtoxicityprompts}, \textit{HarmfulQ} for zero-shot \acrfull{cot} across stereotype benchmarks and harmful questions \cite{shaikh-etal-2023-second}, and \textit{BOLD} \cite{dhamala2021bold}.
\citet{10208759} also introduced the Casual Conversations dataset (version 2) containing 26,467 videos of 5,567 unique participants from 7 different countries, representing a wide range of demographics, for bias and robustness evaluation of \acrshort{lmm}s that are vision and audio models.

\subsection{Debiasing strategies}
Although there's no silver bullet to solving the challenges of fairness and bias in the data and models of multimodal \acrshort{ai}, we believe a combination of two or more of the following strategies on the relevant datasets or models in Tables \ref{datamodels} and \ref{table4} will go a long way in mitigating bias in \acrshort{ai} generally.

\subsubsection{Curate over Crawl}
\label{curate}
One important method to address bias in datasets will be to \textit{curate} rather than crawl.
This is especially so because web crawling has been the popular approach to getting Internet data in the shortest possible time \cite{birhane2021multimodal}.
The assumption of \textit{scale beats noise} is the rationale for this approach by some researchers \cite{jia2021scaling}.
Unfortunately, this misconception about scaling does not only scale the quality part of the dataset but the noise along with it, no matter how small, as shown by \citet{birhane2024into} when they observed 12\% increase in hate content due to scaling.
On the other hand, clearly, despite the advantage of curation, one hurdle to overcome with the \textit{curate over crawl} approach will be the issue of scaling.

In addition, for better quality data collection, researchers like \citet{jo2020lessons} 
 advocate that \acrshort{ai} practitioners should build on the practices of those in the field of archives and libraries and have a public mission statement to guide their data collection practice \cite{weinberg2022rethinking}.
Furthermore, \citet{gebru2021datasheets} encourage, through `datasheets for datasets', standardized processes for documenting important aspects of the dataset creation, including motivation, composition, funding, collection and use cases, with the potential to mitigate unwanted biases, though this approach has its limitations because individuals included or affected by the datasets are not necessarily empowered to influence them  \cite{weinberg2022rethinking}.

\subsubsection{\acrfull{cda}}
\citet{zhao-etal-2018-gender} used \acrshort{cda} to show that it removes bias with minimal performance degradation on coreference benchmarks when combined with existing word-embedding debiasing methods.
It involves generating alternate examples of what exists (counterfactual) of data points to counter or mitigate bias.
This method has gained attention in the field \cite{meade-etal-2022-empirical,barikeri-etal-2021-redditbias}.
It is sometimes called `gender-swapping' in the specific case of gender bias \cite{sun-etal-2019-mitigating}.
\acrshort{cda} has been shown to be effective in tackling bias in coreference resolution, as mentioned earlier, as it reduced the difference in F1 scores between pro-stereotypical and antistereotypical evaluations \cite{sun-etal-2019-mitigating}.
\citet{pmlr-v89-dev19a} also embraced this neutralizing approach in their work by flipping gendered terms with their counterparts.
For example, a sentence like `he was a doctor' would be flipped to `she was a doctor'.
Despite the advantages, as pointed out earlier about the less-discussed preprocessing bias mitigation method, completely debiasing textual data can be complicated and intricate.
Some of the challenges are that the size of the data increases significantly, thereby increasing model training time, and naive term-swapping can create more challenges of unrealistic or nonsensical samples in the data, e.g. `she has hot flashes because of menopause' to `he has hot flashes because of menopause' \cite{sun-etal-2019-mitigating}.

\subsubsection{Improved Filtering}

It has been shown that poor filtering during the data creation process allows low quality data in the final dataset \cite{birhane2024dark}.
It may not be possible to automatically filter large data to be 100\% fit for purpose but improving the existing methods of filtering will go a long way in mitigating bias.

\subsubsection{Linear Projection}

This method projects all words $w \in W$ orthogonally to the bias vector, ensuring the updated set has no component along the bias vector, $v_{B}$, as given in Equation \ref{linearproj} \cite{bolukbasi2016man,pmlr-v89-dev19a}.

\begin{equation}
\mathit{w\textsuperscript{´}} = \mathit{w - \pi_{B}{\left( w \right)}}
= \mathit{w - \langle w, v_{B} \rangle v_{B}} 
\label{linearproj}
\end{equation}

The span that results becomes less by 1 in the total dimensions, say from 300 to 299, which will have negligible effects on the generalizability of the embeddings.
As an example with gender bias, subtraction with linear projection of gender terms from embeddings will make them close.
Gendered word-pairs that have few word sense (e.g. he - she) and (him - her) can have close enough identifcal positions in the vector space after debiasing.

\subsubsection{Debiasing Word Embeddings}
Removing gender bias from word embeddings can take one of two approaches: (1) removing gender subspace \cite{bolukbasi2016man} and (2) learning gender-neutral embeddings \cite{zhao-etal-2018-learning}.
The two approaches may not be adequate for embeddings that are not based on Euclidean space since cosine similarity will no longer apply \cite{sun-etal-2019-mitigating}.
The first approach modifies an embedding based on the combined properties of word embeddings that gender bias can be captured by a direction and neutral words are linearly separable from gendered words \cite{bolukbasi2016man} while the second approach preserves gender information in some dimensions but compels other dimensions of the word embedding to be free of such \cite{zhao-etal-2018-learning}.

\subsubsection{Adapters}
A post-processing method for bias mitigation that is based on Adapter modules \cite{pmlr-v97-houlsby19a}, called \acrfull{dam}, was introduced by
\citet{kumar-etal-2023-parameter}.
They encapsulate different bias mitigation functionalities and can be integrated when desired in a model, similarly to how AdapterFusion \cite{pfeiffer-etal-2021-adapterfusion} is carried out in multi-task learning.
\acrshort{dam} trains the main adapter and the bias mititgation adapters independently before combining them.
\acrshort{dam} follows an earlier adapter-based debiasing method, called \acrfull{adele}, performed in the work by \citet{lauscher-etal-2021-sustainable-modular}.
Their approach involved the additional use of \acrshort{cda}.

\subsubsection{Additive Residuals}

To address the skewed distribution of different identity groups in the training data used in \acrshort{lmm}s, \citet{10204258} introduced \acrfull{dear} to learn additive residual image representations.
This minimizes the representations' capacity to distinguish among different identity groups, thereby offering fairer output.

\subsubsection{Continued Pretraining}

\citet{fatemi-etal-2023-improving} built on the continued pretraining concept, which is sometimes used for gender bias mitigation with a small gender-neutral dataset \cite{de-vassimon-manela-etal-2021-stereotype}, by introducing \acrfull{geep} such that it reduces catastrophic forgetting, which is a likely event in continued pretraining \cite{kirkpatrick2017overcoming}.
\acrshort{geep} achieves this by freezing the entire model before updating the embeddings.
Furthermore, \citet{cabello-etal-2023-evaluating} showed that continued pretraining on gender-neutral data improves fairness by reducing group disparities in some language-vision tasks.

\subsubsection{Adversarial Learning}
\citet{yan2020mitigating} used adversarial learning for bias mitigation as proposed by \citet{zhang2018mitigating}.
They added a discriminator to jointly learn with the predictor for the sensitive attributes.
In this approach, the generator prevents the discriminator from identifying gender in a task.

\subsubsection{Gender Tagging}
In \acrfull{mt}, gender-tagging may be used \cite{vanmassenhove-etal-2018-getting}.
It involves the addition of gender tags to the beginning of data samples to identify the gender of the source data, e.g. `FEMALE I'm travelling tomorrow'.
Apparently, for more complex sentences this approach may become more challenging \cite{sun-etal-2019-mitigating}.

\subsubsection{FairDistillation}
To address bias across languages, \citet{delobelle2022fairdistillation} introduced \textit{FairDistillation}.
It is a cross-lingual method that is based on knowledge distillation \cite{hinton2015distilling} by creating smaller language models from large ones to control for stereotypical and representational biases.

\subsubsection{DEBIE}
\citet{friedrich-etal-2021-debie} introduced DEBIE as a platform for measuring and mitigating implicit and explicit bias in word embeddings.
The mitigation methods are more specific to \acrshort{nlp} and not available in general purpose library such as \acrfull{aif} \cite{8843908}.
DEBIE is a collection of commonly used bias data tools and word embeddings, including fastText \cite{bojanowski2017enriching}, GloVe \cite{pennington2014glove}, \acrfull{cbow} \cite{mikolov2013distributed}, and the \acrshort{weat} test.

\subsubsection{Other strategies}

Furthermore, there are generation detoxifying methods with the potential to reduce bias \cite{gehman-etal-2020-realtoxicityprompts}.
These include the earlier-mentioned continued pretraining and decoding-based generation.
\citet{buolamwini2018gender} advocated oversampling under-represented groups in data to mitigate bias.
\citet{zayed2024fairness} addressed fairness by pruning in \acrshort{llm}s while \citet{guo-etal-2022-auto} introduced auto-bias, by directly probing the biases in pretrained models through prompts.
\citet{liang2021towards} introduced the \acrfull{ainlp} method to carry out post-hoc debiasing on \acrshort{llm}s.
Additionally, there are Self-Debias \cite{schick2021self}, Hard-Debias \cite{bolukbasi2016man}, SentenceDebias \cite{liang-etal-2020-towards} and Dropout methods \cite{webster2020measuring,meade-etal-2022-empirical}.

\section{Conclusion}
\label{conclude}

Fairness and bias are very important considerations in multimodal \acrshort{ai}.
In this work, we presented the challenges of fairness and bias in multimodal data, \acrshort{lmm}s, and \acrshort{llm}s, defining what both terms mean within the scope of this survey, while acknowledging other definitions in the literature.
We discussed the concepts of fairness and bias from the perspective of the Social Science and showed the distribution of scientific publications across many publishers, which reveals the gap in the study of large multimodal \acrshort{ai} compared to \acrshort{llm}s, which this work contributes to filling.
Our discussions around the methods to measure fairness and bias, datasets for evaluation, and debiasing strategies will provide researchers and other stakeholders with insight on how to approach these issues.

For future work, it will be worthwhile to re-evaluate the progress made with the metrics and tools for quantifying and mitigating the challenges of fairness and bias because despite the positive effects of debiasing in \acrshort{ai}, researchers like \citet{west2019discriminating} argue that such research ought to do more than technical debiasing and include the social analysis of the use of such \acrshort{ai}, as this will account more for the impact of bias overall
\cite{weinberg2022rethinking}.
Even \citet{10336315} showed with the example of DALLE-2 that current efforts still have their limitations.
In their work on the \acrshort{lmm} DALLE-2, they revealed that, despite the guardrails for the model by OpenAI, it generated 40 more images of women for a stereotypical female-dominant administrative task, in clear gender bias, when prompted.


\section*{Acknowledgements}
This work is supported by the European Commission-funded project "Humane AI: Toward AI Systems That Augment and Empower Humans by Understanding Us, Our Society and the World Around Us."
The authors wish to thank our colleagues at the \textit{Responsible Artificial Intelligence Group at Umeå University}.
The work is partially supported by the Wallenberg AI, Autonomous Systems and Software Program (WASP), funded by the Knut and Alice Wallenberg Foundation and counterpart funding from Luleå University of Technology (LTU).

\bibliography{anthology,custom}

\begin{thebibliography}{358}
\expandafter\ifx\csname natexlab\endcsname\relax\def\natexlab#1{#1}\fi

\bibitem[{Achiam et~al.(2023)Achiam, Adler, Agarwal, Ahmad, Akkaya, Aleman, Almeida, Altenschmidt, Altman, Anadkat et~al.}]{achiam2023gpt}
Josh Achiam, Steven Adler, Sandhini Agarwal, Lama Ahmad, Ilge Akkaya, Florencia~Leoni Aleman, Diogo Almeida, Janko Altenschmidt, Sam Altman, Shyamal Anadkat, et~al. 2023.
\newblock Gpt-4 technical report.
\newblock \emph{arXiv preprint arXiv:2303.08774}.

\bibitem[{Acosta et~al.(2022)Acosta, Falcone, Rajpurkar, and Topol}]{acosta2022multimodal}
Juli{\'a}n~N Acosta, Guido~J Falcone, Pranav Rajpurkar, and Eric~J Topol. 2022.
\newblock Multimodal biomedical ai.
\newblock \emph{Nature Medicine}, 28(9):1773--1784.

\bibitem[{Adams and Freedman(1976)}]{adams1976equity}
J~Stacy Adams and Sara Freedman. 1976.
\newblock Equity theory revisited: Comments and annotated bibliography.
\newblock \emph{Advances in experimental social psychology}, 9:43--90.

\bibitem[{Adewumi et~al.(2023{\natexlab{a}})Adewumi, Adeyemi, Anuoluwapo, Peters, Buzaaba, Samuel, Rufai, Ajibade, Gwadabe, Traore et~al.}]{adewumi2023afriwoz}
Tosin Adewumi, Mofetoluwa Adeyemi, Aremu Anuoluwapo, Bukola Peters, Happy Buzaaba, Oyerinde Samuel, Amina~Mardiyyah Rufai, Benjamin Ajibade, Tajudeen Gwadabe, Mory Moussou~Koulibaly Traore, et~al. 2023{\natexlab{a}}.
\newblock Afriwoz: Corpus for exploiting cross-lingual transfer for dialogue generation in low-resource, african languages.
\newblock In \emph{2023 International Joint Conference on Neural Networks (IJCNN)}, pages 1--8. IEEE.

\bibitem[{Adewumi et~al.(2024{\natexlab{a}})Adewumi, Habib, Alkhaled, and Barney}]{adewumi2024instruction}
Tosin Adewumi, Nudrat Habib, Lama Alkhaled, and Elisa Barney. 2024{\natexlab{a}}.
\newblock Instruction makes a difference.
\newblock \emph{arXiv preprint arXiv:2402.00453}.

\bibitem[{Adewumi et~al.(2024{\natexlab{b}})Adewumi, Habib, Alkhaled, and Barney}]{adewumi2024limitations}
Tosin Adewumi, Nudrat Habib, Lama Alkhaled, and Elisa Barney. 2024{\natexlab{b}}.
\newblock On the limitations of large language models (llms): False attribution.
\newblock \emph{arXiv preprint arXiv:2404.04631}.

\bibitem[{Adewumi et~al.(2022)Adewumi, Liwicki, and Liwicki}]{adewumi2022state}
Tosin Adewumi, Foteini Liwicki, and Marcus Liwicki. 2022.
\newblock State-of-the-art in open-domain conversational ai: A survey.
\newblock \emph{Information}, 13(6):298.

\bibitem[{Adewumi et~al.(2023{\natexlab{b}})Adewumi, S{\"o}dergren, Alkhaled, Al-azzawi, Simistira~Liwicki, and Liwicki}]{adewumi-etal-2023-bipol}
Tosin Adewumi, Isabella S{\"o}dergren, Lama Alkhaled, Sana Al-azzawi, Foteini Simistira~Liwicki, and Marcus Liwicki. 2023{\natexlab{b}}.
\newblock \href {https://aclanthology.org/2023.ranlp-1.1} {Bipol: Multi-axes evaluation of bias with explainability in benchmark datasets}.
\newblock In \emph{Proceedings of the 14th International Conference on Recent Advances in Natural Language Processing}, pages 1--10, Varna, Bulgaria. INCOMA Ltd., Shoumen, Bulgaria.

\bibitem[{Agrawal et~al.(2022)Agrawal, Hegselmann, Lang, Kim, and Sontag}]{agrawal-etal-2022-large}
Monica Agrawal, Stefan Hegselmann, Hunter Lang, Yoon Kim, and David Sontag. 2022.
\newblock \href {https://doi.org/10.18653/v1/2022.emnlp-main.130} {Large language models are few-shot clinical information extractors}.
\newblock In \emph{Proceedings of the 2022 Conference on Empirical Methods in Natural Language Processing}, pages 1998--2022, Abu Dhabi, United Arab Emirates. Association for Computational Linguistics.

\bibitem[{Aher et~al.(2023)Aher, Arriaga, and Kalai}]{aher2023using}
Gati~V Aher, Rosa~I Arriaga, and Adam~Tauman Kalai. 2023.
\newblock Using large language models to simulate multiple humans and replicate human subject studies.
\newblock In \emph{International Conference on Machine Learning}, pages 337--371. PMLR.

\bibitem[{Aich et~al.(2024)Aich, Liu, Giorgi, Isman, Ungar, and Curtis}]{aich2024vernacular}
Ankit Aich, Tingting Liu, Salvatore Giorgi, Kelsey Isman, Lyle Ungar, and Brenda Curtis. 2024.
\newblock Vernacular? i barely know her: Challenges with style control and stereotyping.
\newblock \emph{arXiv preprint arXiv:2406.12679}.

\bibitem[{Alam(2022)}]{alam2022college}
Mohammad Arif~Ul Alam. 2022.
\newblock College student retention risk analysis from educational database using multi-task multi-modal neural fusion.
\newblock In \emph{Proceedings of the AAAI Conference on Artificial Intelligence}, volume~36, pages 12689--12697.

\bibitem[{Alasadi et~al.(2020)Alasadi, Arunachalam, Atrey, and Singh}]{alasadi2020fairness}
Jamal Alasadi, Ramanathan Arunachalam, Pradeep~K Atrey, and Vivek~K Singh. 2020.
\newblock A fairness-aware fusion framework for multimodal cyberbullying detection.
\newblock In \emph{2020 IEEE Sixth International Conference on Multimedia Big Data (BigMM)}, pages 166--173. IEEE.

\bibitem[{Alkhaled et~al.(2023)Alkhaled, Adewumi, and Sabry}]{ALKHALED2023100030}
Lama Alkhaled, Tosin Adewumi, and Sana~Sabah Sabry. 2023.
\newblock \href {https://doi.org/https://doi.org/10.1016/j.nlp.2023.100030} {Bipol: A novel multi-axes bias evaluation metric with explainability for nlp}.
\newblock \emph{Natural Language Processing Journal}, 4:100030.

\bibitem[{Almazrouei et~al.(2023)Almazrouei, Alobeidli, Alshamsi, Cappelli, Cojocaru, Debbah, Goffinet, Hesslow, Launay, Malartic et~al.}]{almazrouei2023falcon}
Ebtesam Almazrouei, Hamza Alobeidli, Abdulaziz Alshamsi, Alessandro Cappelli, Ruxandra Cojocaru, M{\'e}rouane Debbah, {\'E}tienne Goffinet, Daniel Hesslow, Julien Launay, Quentin Malartic, et~al. 2023.
\newblock The falcon series of open language models.
\newblock \emph{arXiv preprint arXiv:2311.16867}.

\bibitem[{Alwahaby et~al.(2022)Alwahaby, Cukurova, Papamitsiou, and Giannakos}]{alwahaby2022evidence}
Haifa Alwahaby, Mutlu Cukurova, Zacharoula Papamitsiou, and Michail Giannakos. 2022.
\newblock The evidence of impact and ethical considerations of multimodal learning analytics: A systematic literature review.
\newblock \emph{The multimodal learning analytics handbook}, pages 289--325.

\bibitem[{Andrighetto et~al.(2019)Andrighetto, Bracco, Chiorri, Masini, Passarelli, and Piccinno}]{andrighetto2019now}
Luca Andrighetto, Fabrizio Bracco, Carlo Chiorri, Michele Masini, Marcello Passarelli, and Tommaso~Francesco Piccinno. 2019.
\newblock Now you see me, now you don’t: Detecting sexual objectification through a change blindness paradigm.
\newblock \emph{Cognitive Processing}, 20:419--429.

\bibitem[{Antol et~al.(2015)Antol, Agrawal, Lu, Mitchell, Batra, Zitnick, and Parikh}]{antol2015vqa}
Stanislaw Antol, Aishwarya Agrawal, Jiasen Lu, Margaret Mitchell, Dhruv Batra, C~Lawrence Zitnick, and Devi Parikh. 2015.
\newblock Vqa: Visual question answering.
\newblock In \emph{Proceedings of the IEEE international conference on computer vision}, pages 2425--2433.

\bibitem[{Argyle et~al.(2023)Argyle, Busby, Fulda, Gubler, Rytting, and Wingate}]{argyle2023out}
Lisa~P Argyle, Ethan~C Busby, Nancy Fulda, Joshua~R Gubler, Christopher Rytting, and David Wingate. 2023.
\newblock Out of one, many: Using language models to simulate human samples.
\newblock \emph{Political Analysis}, 31(3):337--351.

\bibitem[{Balayn et~al.(2021)Balayn, Lofi, and Houben}]{balayn2021managing}
Agathe Balayn, Christoph Lofi, and Geert-Jan Houben. 2021.
\newblock Managing bias and unfairness in data for decision support: a survey of machine learning and data engineering approaches to identify and mitigate bias and unfairness within data management and analytics systems.
\newblock \emph{The VLDB Journal}, 30(5):739--768.

\bibitem[{Barikeri et~al.(2021)Barikeri, Lauscher, Vuli{\'c}, and Glava{\v{s}}}]{barikeri-etal-2021-redditbias}
Soumya Barikeri, Anne Lauscher, Ivan Vuli{\'c}, and Goran Glava{\v{s}}. 2021.
\newblock \href {https://doi.org/10.18653/v1/2021.acl-long.151} {{R}eddit{B}ias: A real-world resource for bias evaluation and debiasing of conversational language models}.
\newblock In \emph{Proceedings of the 59th Annual Meeting of the Association for Computational Linguistics and the 11th International Joint Conference on Natural Language Processing (Volume 1: Long Papers)}, pages 1941--1955, Online. Association for Computational Linguistics.

\bibitem[{Bellamy et~al.(2019)Bellamy, Dey, Hind, Hoffman, Houde, Kannan, Lohia, Martino, Mehta, Mojsilović, Nagar, Ramamurthy, Richards, Saha, Sattigeri, Singh, Varshney, and Zhang}]{8843908}
R.~K.~E. Bellamy, K.~Dey, M.~Hind, S.~C. Hoffman, S.~Houde, K.~Kannan, P.~Lohia, J.~Martino, S.~Mehta, A.~Mojsilović, S.~Nagar, K.~Natesan Ramamurthy, J.~Richards, D.~Saha, P.~Sattigeri, M.~Singh, K.~R. Varshney, and Y.~Zhang. 2019.
\newblock \href {https://doi.org/10.1147/JRD.2019.2942287} {Ai fairness 360: An extensible toolkit for detecting and mitigating algorithmic bias}.
\newblock \emph{IBM Journal of Research and Development}, 63(4/5):4:1--4:15.

\bibitem[{bench authors(2023)}]{srivastava2023beyond}
BIG bench authors. 2023.
\newblock \href {https://openreview.net/forum?id=uyTL5Bvosj} {Beyond the imitation game: Quantifying and extrapolating the capabilities of language models}.
\newblock \emph{Transactions on Machine Learning Research}.

\bibitem[{Bender et~al.(2021)Bender, Gebru, McMillan-Major, and Shmitchell}]{bender2021dangers}
Emily~M Bender, Timnit Gebru, Angelina McMillan-Major, and Shmargaret Shmitchell. 2021.
\newblock On the dangers of stochastic parrots: Can language models be too big?
\newblock In \emph{Proceedings of the 2021 ACM conference on fairness, accountability, and transparency}, pages 610--623.

\bibitem[{Berry(2015)}]{berry2015differential}
Christopher~M Berry. 2015.
\newblock Differential validity and differential prediction of cognitive ability tests: Understanding test bias in the employment context.
\newblock \emph{Annu. Rev. Organ. Psychol. Organ. Behav.}, 2(1):435--463.

\bibitem[{Bevara et~al.(2024)Bevara, Mannuru, Karedla, and Xiao}]{bevara2024scaling}
Ravi Varma~Kumar Bevara, Nishith~Reddy Mannuru, Sai~Pranathi Karedla, and Ting Xiao. 2024.
\newblock Scaling implicit bias analysis across transformer-based language models through embedding association test and prompt engineering.
\newblock \emph{Applied Sciences}, 14(8):3483.

\bibitem[{Biderman et~al.(2023)Biderman, Schoelkopf, Anthony, Bradley, O’Brien, Hallahan, Khan, Purohit, Prashanth, Raff et~al.}]{biderman2023pythia}
Stella Biderman, Hailey Schoelkopf, Quentin~Gregory Anthony, Herbie Bradley, Kyle O’Brien, Eric Hallahan, Mohammad~Aflah Khan, Shivanshu Purohit, USVSN~Sai Prashanth, Edward Raff, et~al. 2023.
\newblock Pythia: A suite for analyzing large language models across training and scaling.
\newblock In \emph{International Conference on Machine Learning}, pages 2397--2430. PMLR.

\bibitem[{Birhane et~al.(2024{\natexlab{a}})Birhane, Dehdashtian, Prabhu, and Boddeti}]{birhane2024dark}
Abeba Birhane, Sepehr Dehdashtian, Vinay Prabhu, and Vishnu Boddeti. 2024{\natexlab{a}}.
\newblock The dark side of dataset scaling: Evaluating racial classification in multimodal models.
\newblock In \emph{The 2024 ACM Conference on Fairness, Accountability, and Transparency}, pages 1229--1244.

\bibitem[{Birhane et~al.(2024{\natexlab{b}})Birhane, Han, Boddeti, Luccioni et~al.}]{birhane2024into}
Abeba Birhane, Sanghyun Han, Vishnu Boddeti, Sasha Luccioni, et~al. 2024{\natexlab{b}}.
\newblock Into the laion’s den: Investigating hate in multimodal datasets.
\newblock \emph{Advances in Neural Information Processing Systems}, 36.

\bibitem[{Birhane et~al.(2021)Birhane, Prabhu, and Kahembwe}]{birhane2021multimodal}
Abeba Birhane, Vinay~Uday Prabhu, and Emmanuel Kahembwe. 2021.
\newblock Multimodal datasets: misogyny, pornography, and malignant stereotypes.
\newblock \emph{arXiv preprint arXiv:2110.01963}.

\bibitem[{Blodgett et~al.(2020)Blodgett, Barocas, Daum{\'e}~III, and Wallach}]{blodgett-etal-2020-language}
Su~Lin Blodgett, Solon Barocas, Hal Daum{\'e}~III, and Hanna Wallach. 2020.
\newblock \href {https://doi.org/10.18653/v1/2020.acl-main.485} {Language (technology) is power: A critical survey of {``}bias{''} in {NLP}}.
\newblock In \emph{Proceedings of the 58th Annual Meeting of the Association for Computational Linguistics}, pages 5454--5476, Online. Association for Computational Linguistics.

\bibitem[{Blodgett et~al.(2021)Blodgett, Lopez, Olteanu, Sim, and Wallach}]{blodgett2021stereotyping}
Su~Lin Blodgett, Gilsinia Lopez, Alexandra Olteanu, Robert Sim, and Hanna Wallach. 2021.
\newblock Stereotyping norwegian salmon: An inventory of pitfalls in fairness benchmark datasets.
\newblock In \emph{Proceedings of the 59th Annual Meeting of the Association for Computational Linguistics and the 11th International Joint Conference on Natural Language Processing (Volume 1: Long Papers)}, pages 1004--1015.

\bibitem[{Bojanowski et~al.(2017)Bojanowski, Grave, Joulin, and Mikolov}]{bojanowski2017enriching}
Piotr Bojanowski, Edouard Grave, Armand Joulin, and Tomas Mikolov. 2017.
\newblock Enriching word vectors with subword information.
\newblock \emph{Transactions of the association for computational linguistics}, 5:135--146.

\bibitem[{Bolukbasi et~al.(2016)Bolukbasi, Chang, Zou, Saligrama, and Kalai}]{bolukbasi2016man}
Tolga Bolukbasi, Kai-Wei Chang, James~Y Zou, Venkatesh Saligrama, and Adam~T Kalai. 2016.
\newblock Man is to computer programmer as woman is to homemaker? debiasing word embeddings.
\newblock \emph{Advances in neural information processing systems}, 29.

\bibitem[{Booth et~al.(2021)Booth, Hickman, Subburaj, Tay, Woo, and D'Mello}]{booth2021bias}
Brandon~M Booth, Louis Hickman, Shree~Krishna Subburaj, Louis Tay, Sang~Eun Woo, and Sidney~K D'Mello. 2021.
\newblock Bias and fairness in multimodal machine learning: A case study of automated video interviews.
\newblock In \emph{Proceedings of the 2021 International Conference on Multimodal Interaction}, pages 268--277.

\bibitem[{Brereton et~al.(2007)Brereton, Kitchenham, Budgen, Turner, and Khalil}]{brereton2007lessons}
Pearl Brereton, Barbara~A Kitchenham, David Budgen, Mark Turner, and Mohamed Khalil. 2007.
\newblock Lessons from applying the systematic literature review process within the software engineering domain.
\newblock \emph{Journal of systems and software}, 80(4):571--583.

\bibitem[{Brinkmann et~al.(2023)Brinkmann, Swoboda, and Bartelt}]{brinkmann2023multidimensional}
Jannik Brinkmann, Paul Swoboda, and Christian Bartelt. 2023.
\newblock A multidimensional analysis of social biases in vision transformers.
\newblock In \emph{Proceedings of the IEEE/CVF International Conference on Computer Vision}, pages 4914--4923.

\bibitem[{Brown et~al.(2020)Brown, Mann, Ryder, Subbiah, Kaplan, Dhariwal, Neelakantan, Shyam, Sastry, Askell et~al.}]{brown2020language}
Tom Brown, Benjamin Mann, Nick Ryder, Melanie Subbiah, Jared~D Kaplan, Prafulla Dhariwal, Arvind Neelakantan, Pranav Shyam, Girish Sastry, Amanda Askell, et~al. 2020.
\newblock Language models are few-shot learners.
\newblock \emph{Advances in neural information processing systems}, 33:1877--1901.

\bibitem[{Brunet et~al.(2019)Brunet, Alkalay-Houlihan, Anderson, and Zemel}]{brunet2019understanding}
Marc-Etienne Brunet, Colleen Alkalay-Houlihan, Ashton Anderson, and Richard Zemel. 2019.
\newblock Understanding the origins of bias in word embeddings.
\newblock In \emph{International conference on machine learning}, pages 803--811. PMLR.

\bibitem[{Buolamwini and Gebru(2018)}]{buolamwini2018gender}
Joy Buolamwini and Timnit Gebru. 2018.
\newblock Gender shades: Intersectional accuracy disparities in commercial gender classification.
\newblock In \emph{Conference on fairness, accountability and transparency}, pages 77--91. PMLR.

\bibitem[{Cabello et~al.(2023)Cabello, Bugliarello, Brandl, and Elliott}]{cabello-etal-2023-evaluating}
Laura Cabello, Emanuele Bugliarello, Stephanie Brandl, and Desmond Elliott. 2023.
\newblock \href {https://doi.org/10.18653/v1/2023.emnlp-main.525} {Evaluating bias and fairness in gender-neutral pretrained vision-and-language models}.
\newblock In \emph{Proceedings of the 2023 Conference on Empirical Methods in Natural Language Processing}, pages 8465--8483, Singapore. Association for Computational Linguistics.

\bibitem[{Cai et~al.(2024)Cai, Cao, Guo, Wen, Liu, and Chen}]{cai2024locating}
Yuchen Cai, Ding Cao, Rongxi Guo, Yaqin Wen, Guiquan Liu, and Enhong Chen. 2024.
\newblock Locating and mitigating gender bias in large language models.
\newblock In \emph{International Conference on Intelligent Computing}, pages 471--482. Springer.

\bibitem[{Caliskan et~al.(2017)Caliskan, Bryson, and Narayanan}]{caliskan2017semantics}
Aylin Caliskan, Joanna~J Bryson, and Arvind Narayanan. 2017.
\newblock Semantics derived automatically from language corpora contain human-like biases.
\newblock \emph{Science}, 356(6334):183--186.

\bibitem[{Caselli et~al.(2021)Caselli, Basile, Mitrovi{\'c}, and Granitzer}]{caselli-etal-2021-hatebert}
Tommaso Caselli, Valerio Basile, Jelena Mitrovi{\'c}, and Michael Granitzer. 2021.
\newblock \href {https://doi.org/10.18653/v1/2021.woah-1.3} {{H}ate{BERT}: Retraining {BERT} for abusive language detection in {E}nglish}.
\newblock In \emph{Proceedings of the 5th Workshop on Online Abuse and Harms (WOAH 2021)}, pages 17--25, Online. Association for Computational Linguistics.

\bibitem[{Chang et~al.(2024)Chang, Wang, Wang, Wu, Yang, Zhu, Chen, Yi, Wang, Wang et~al.}]{chang2024survey}
Yupeng Chang, Xu~Wang, Jindong Wang, Yuan Wu, Linyi Yang, Kaijie Zhu, Hao Chen, Xiaoyuan Yi, Cunxiang Wang, Yidong Wang, et~al. 2024.
\newblock A survey on evaluation of large language models.
\newblock \emph{ACM Transactions on Intelligent Systems and Technology}, 15(3):1--45.

\bibitem[{Chen and Mueller(2024)}]{chen2024quantifying}
Jiuhai Chen and Jonas Mueller. 2024.
\newblock Quantifying uncertainty in answers from any language model and enhancing their trustworthiness.
\newblock In \emph{Proceedings of the 62nd Annual Meeting of the Association for Computational Linguistics (Volume 1: Long Papers)}, pages 5186--5200.

\bibitem[{Chen et~al.(2021)Chen, Tworek, Jun, Yuan, Pinto, Kaplan, Edwards, Burda, Joseph, Brockman et~al.}]{chen2021evaluating}
Mark Chen, Jerry Tworek, Heewoo Jun, Qiming Yuan, Henrique Ponde De~Oliveira Pinto, Jared Kaplan, Harri Edwards, Yuri Burda, Nicholas Joseph, Greg Brockman, et~al. 2021.
\newblock Evaluating large language models trained on code.
\newblock \emph{arXiv preprint arXiv:2107.03374}.

\bibitem[{Chen et~al.(2022)Chen, Wang, Changpinyo, Piergiovanni, Padlewski, Salz, Goodman, Grycner, Mustafa, Beyer et~al.}]{chen2022pali}
Xi~Chen, Xiao Wang, Soravit Changpinyo, AJ~Piergiovanni, Piotr Padlewski, Daniel Salz, Sebastian Goodman, Adam Grycner, Basil Mustafa, Lucas Beyer, et~al. 2022.
\newblock Pali: A jointly-scaled multilingual language-image model.
\newblock \emph{arXiv preprint arXiv:2209.06794}.

\bibitem[{Chen et~al.(2024)Chen, Du, Wen, Zhou, Cui, Weng, Tu, Wang, Tong, Huang et~al.}]{chen2024mj}
Zhaorun Chen, Yichao Du, Zichen Wen, Yiyang Zhou, Chenhang Cui, Zhenzhen Weng, Haoqin Tu, Chaoqi Wang, Zhengwei Tong, Qinglan Huang, et~al. 2024.
\newblock Mj-bench: Is your multimodal reward model really a good judge for text-to-image generation?
\newblock \emph{arXiv preprint arXiv:2407.04842}.

\bibitem[{Cheong et~al.(2024)Cheong, Kalkan, and Gunes}]{cheong2024fairrefuse}
Jiaee Cheong, Sinan Kalkan, and Hatice Gunes. 2024.
\newblock Fairrefuse: Referee-guided fusion for multi-modal causal fairness in depression detection.

\bibitem[{Cho et~al.(2023)Cho, Zala, and Bansal}]{cho2023dall}
Jaemin Cho, Abhay Zala, and Mohit Bansal. 2023.
\newblock Dall-eval: Probing the reasoning skills and social biases of text-to-image generation models.
\newblock In \emph{Proceedings of the IEEE/CVF International Conference on Computer Vision}, pages 3043--3054.

\bibitem[{Chowdhery et~al.(2024)Chowdhery, Narang, Devlin, Bosma, Mishra, Roberts, Barham, Chung, Sutton, Gehrmann, Schuh, Shi, Tsvyashchenko, Maynez, Rao, Barnes, Tay, Shazeer, Prabhakaran, Reif, Du, Hutchinson, Pope, Bradbury, Austin, Isard, Gur-Ari, Yin, Duke, Levskaya, Ghemawat, Dev, Michalewski, Garcia, Misra, Robinson, Fedus, Zhou, Ippolito, Luan, Lim, Zoph, Spiridonov, Sepassi, Dohan, Agrawal, Omernick, Dai, Pillai, Pellat, Lewkowycz, Moreira, Child, Polozov, Lee, Zhou, Wang, Saeta, Diaz, Firat, Catasta, Wei, Meier-Hellstern, Eck, Dean, Petrov, and Fiedel}]{10.5555/3648699.3648939}
Aakanksha Chowdhery, Sharan Narang, Jacob Devlin, Maarten Bosma, Gaurav Mishra, Adam Roberts, Paul Barham, Hyung~Won Chung, Charles Sutton, Sebastian Gehrmann, Parker Schuh, Kensen Shi, Sashank Tsvyashchenko, Joshua Maynez, Abhishek Rao, Parker Barnes, Yi~Tay, Noam Shazeer, Vinodkumar Prabhakaran, Emily Reif, Nan Du, Ben Hutchinson, Reiner Pope, James Bradbury, Jacob Austin, Michael Isard, Guy Gur-Ari, Pengcheng Yin, Toju Duke, Anselm Levskaya, Sanjay Ghemawat, Sunipa Dev, Henryk Michalewski, Xavier Garcia, Vedant Misra, Kevin Robinson, Liam Fedus, Denny Zhou, Daphne Ippolito, David Luan, Hyeontaek Lim, Barret Zoph, Alexander Spiridonov, Ryan Sepassi, David Dohan, Shivani Agrawal, Mark Omernick, Andrew~M. Dai, Thanumalayan~Sankaranarayana Pillai, Marie Pellat, Aitor Lewkowycz, Erica Moreira, Rewon Child, Oleksandr Polozov, Katherine Lee, Zongwei Zhou, Xuezhi Wang, Brennan Saeta, Mark Diaz, Orhan Firat, Michele Catasta, Jason Wei, Kathy Meier-Hellstern, Douglas Eck, Jeff Dean, Slav Petrov, and Noah Fiedel.
  2024.
\newblock Palm: scaling language modeling with pathways.
\newblock \emph{J. Mach. Learn. Res.}, 24(1).

\bibitem[{Chowdhery et~al.(2023)Chowdhery, Narang, Devlin, Bosma, Mishra, Roberts, Barham, Chung, Sutton, Gehrmann et~al.}]{chowdhery2023palm}
Aakanksha Chowdhery, Sharan Narang, Jacob Devlin, Maarten Bosma, Gaurav Mishra, Adam Roberts, Paul Barham, Hyung~Won Chung, Charles Sutton, Sebastian Gehrmann, et~al. 2023.
\newblock Palm: Scaling language modeling with pathways.
\newblock \emph{Journal of Machine Learning Research}, 24(240):1--113.

\bibitem[{Chung et~al.(2024)Chung, Hou, Longpre, Zoph, Tay, Fedus, Li, Wang, Dehghani, Brahma et~al.}]{chung2024scaling}
Hyung~Won Chung, Le~Hou, Shayne Longpre, Barret Zoph, Yi~Tay, William Fedus, Yunxuan Li, Xuezhi Wang, Mostafa Dehghani, Siddhartha Brahma, et~al. 2024.
\newblock Scaling instruction-finetuned language models.
\newblock \emph{Journal of Machine Learning Research}, 25(70):1--53.

\bibitem[{Chung et~al.(2018)Chung, Nagrani, and Zisserman}]{chung2018voxceleb2}
Joon~Son Chung, Arsha Nagrani, and Andrew Zisserman. 2018.
\newblock Voxceleb2: Deep speaker recognition.
\newblock \emph{arXiv preprint arXiv:1806.05622}.

\bibitem[{Clusmann et~al.(2023)Clusmann, Kolbinger, Muti, Carrero, Eckardt, Laleh, L{\"o}ffler, Schwarzkopf, Unger, Veldhuizen et~al.}]{clusmann2023future}
Jan Clusmann, Fiona~R Kolbinger, Hannah~Sophie Muti, Zunamys~I Carrero, Jan-Niklas Eckardt, Narmin~Ghaffari Laleh, Chiara Maria~Lavinia L{\"o}ffler, Sophie-Caroline Schwarzkopf, Michaela Unger, Gregory~P Veldhuizen, et~al. 2023.
\newblock The future landscape of large language models in medicine.
\newblock \emph{Communications medicine}, 3(1):141.

\bibitem[{Crowson et~al.(2022)Crowson, Biderman, Kornis, Stander, Hallahan, Castricato, and Raff}]{crowson2022vqgan}
Katherine Crowson, Stella Biderman, Daniel Kornis, Dashiell Stander, Eric Hallahan, Louis Castricato, and Edward Raff. 2022.
\newblock \href {https://doi.org/10.1007/978-3-031-19836-6_6} {Vqgan-clip: Open domain image generation and editing with natural language guidance}.
\newblock In \emph{Computer Vision – ECCV 2022: 17th European Conference, Tel Aviv, Israel, October 23–27, 2022, Proceedings, Part XXXVII}, page 88–105, Berlin, Heidelberg. Springer-Verlag.

\bibitem[{Da et~al.(2024)Da, Bossa, Berenguer, and Sahli}]{10388308}
Yifei Da, Matías~Nicolás Bossa, Abel~Díaz Berenguer, and Hichem Sahli. 2024.
\newblock \href {https://doi.org/10.1109/ACCESS.2024.3353056} {Reducing bias in sentiment analysis models through causal mediation analysis and targeted counterfactual training}.
\newblock \emph{IEEE Access}, 12:10120--10134.

\bibitem[{da~Silva et~al.(2021)da~Silva, Louro, Goncalves, Marques, Dias, da~Cunha, and Tasinaffo}]{da2021could}
Daniela~America da~Silva, Henrique Duarte~Borges Louro, Gildarcio~Sousa Goncalves, Johnny~Cardoso Marques, Luiz Alberto~Vieira Dias, Adilson~Marques da~Cunha, and Paulo~Marcelo Tasinaffo. 2021.
\newblock Could a conversational ai identify offensive language?
\newblock \emph{Information}, 12(10):418.

\bibitem[{Dacon and Liu(2021)}]{10.1145/3442442.3452325}
Jamell Dacon and Haochen Liu. 2021.
\newblock \href {https://doi.org/10.1145/3442442.3452325} {Does gender matter in the news? detecting and examining gender bias in news articles}.
\newblock In \emph{Companion Proceedings of the Web Conference 2021}, WWW '21, page 385–392, New York, NY, USA. Association for Computing Machinery.

\bibitem[{de~Vassimon~Manela et~al.(2021)de~Vassimon~Manela, Errington, Fisher, van Breugel, and Minervini}]{de-vassimon-manela-etal-2021-stereotype}
Daniel de~Vassimon~Manela, David Errington, Thomas Fisher, Boris van Breugel, and Pasquale Minervini. 2021.
\newblock \href {https://doi.org/10.18653/v1/2021.eacl-main.190} {Stereotype and skew: Quantifying gender bias in pre-trained and fine-tuned language models}.
\newblock In \emph{Proceedings of the 16th Conference of the European Chapter of the Association for Computational Linguistics: Main Volume}, pages 2232--2242, Online. Association for Computational Linguistics.

\bibitem[{Delobelle and Berendt(2022)}]{delobelle2022fairdistillation}
Pieter Delobelle and Bettina Berendt. 2022.
\newblock Fairdistillation: mitigating stereotyping in language models.
\newblock In \emph{Joint European Conference on Machine Learning and Knowledge Discovery in Databases}, pages 638--654. Springer.

\bibitem[{Delobelle et~al.(2022)Delobelle, Tokpo, Calders, and Berendt}]{delobelle-etal-2022-measuring}
Pieter Delobelle, Ewoenam Tokpo, Toon Calders, and Bettina Berendt. 2022.
\newblock \href {https://doi.org/10.18653/v1/2022.naacl-main.122} {Measuring fairness with biased rulers: A comparative study on bias metrics for pre-trained language models}.
\newblock In \emph{Proceedings of the 2022 Conference of the North American Chapter of the Association for Computational Linguistics: Human Language Technologies}, pages 1693--1706, Seattle, United States. Association for Computational Linguistics.

\bibitem[{Delobelle et~al.(2020)Delobelle, Winters, and Berendt}]{delobelle-etal-2020-robbert}
Pieter Delobelle, Thomas Winters, and Bettina Berendt. 2020.
\newblock \href {https://doi.org/10.18653/v1/2020.findings-emnlp.292} {{R}ob{BERT}: a {D}utch {R}o{BERT}a-based {L}anguage {M}odel}.
\newblock In \emph{Findings of the Association for Computational Linguistics: EMNLP 2020}, pages 3255--3265, Online. Association for Computational Linguistics.

\bibitem[{Desai et~al.(2023)Desai, Patil, Patil, and Mehta}]{desai2023large}
Bhavin Desai, Kapil Patil, Asit Patil, and Ishita Mehta. 2023.
\newblock Large language models: A comprehensive exploration of modern ai's potential and pitfalls.
\newblock \emph{Journal of Innovative Technologies}, 6(1).

\bibitem[{Dev and Phillips(2019)}]{pmlr-v89-dev19a}
Sunipa Dev and Jeff Phillips. 2019.
\newblock \href {https://proceedings.mlr.press/v89/dev19a.html} {Attenuating bias in word vectors}.
\newblock In \emph{Proceedings of the Twenty-Second International Conference on Artificial Intelligence and Statistics}, volume~89 of \emph{Proceedings of Machine Learning Research}, pages 879--887. PMLR.

\bibitem[{Dhamala et~al.(2021)Dhamala, Sun, Kumar, Krishna, Pruksachatkun, Chang, and Gupta}]{dhamala2021bold}
Jwala Dhamala, Tony Sun, Varun Kumar, Satyapriya Krishna, Yada Pruksachatkun, Kai-Wei Chang, and Rahul Gupta. 2021.
\newblock Bold: Dataset and metrics for measuring biases in open-ended language generation.
\newblock In \emph{Proceedings of the 2021 ACM conference on fairness, accountability, and transparency}, pages 862--872.

\bibitem[{Dinan et~al.(2020)Dinan, Fan, Williams, Urbanek, Kiela, and Weston}]{dinan-etal-2020-queens}
Emily Dinan, Angela Fan, Adina Williams, Jack Urbanek, Douwe Kiela, and Jason Weston. 2020.
\newblock \href {https://doi.org/10.18653/v1/2020.emnlp-main.656} {Queens are powerful too: Mitigating gender bias in dialogue generation}.
\newblock In \emph{Proceedings of the 2020 Conference on Empirical Methods in Natural Language Processing (EMNLP)}, pages 8173--8188, Online. Association for Computational Linguistics.

\bibitem[{Doan et~al.(2024)Doan, Chu, Wang, and Zhang}]{doan2024fairness}
Thang~Viet Doan, Zhibo Chu, Zichong Wang, and Wenbin Zhang. 2024.
\newblock Fairness definitions in language models explained.
\newblock \emph{arXiv preprint arXiv:2407.18454}.

\bibitem[{Dodge et~al.(2019)Dodge, Liao, Zhang, Bellamy, and Dugan}]{dodge2019explaining}
Jonathan Dodge, Q~Vera Liao, Yunfeng Zhang, Rachel~KE Bellamy, and Casey Dugan. 2019.
\newblock Explaining models: an empirical study of how explanations impact fairness judgment.
\newblock In \emph{Proceedings of the 24th international conference on intelligent user interfaces}, pages 275--285.

\bibitem[{Dolci et~al.(2023)Dolci, Azzalini, and Tanelli}]{dolci2023improving}
Tommaso Dolci, Fabio Azzalini, and Mara Tanelli. 2023.
\newblock Improving gender-related fairness in sentence encoders: A semantics-based approach.
\newblock \emph{Data Science and Engineering}, 8(2):177--195.

\bibitem[{Drukker et~al.(2023)Drukker, Chen, Gichoya, Gruszauskas, Kalpathy-Cramer, Koyejo, Myers, S{\'a}, Sahiner, Whitney et~al.}]{drukker2023toward}
Karen Drukker, Weijie Chen, Judy Gichoya, Nicholas Gruszauskas, Jayashree Kalpathy-Cramer, Sanmi Koyejo, Kyle Myers, Rui~C S{\'a}, Berkman Sahiner, Heather Whitney, et~al. 2023.
\newblock Toward fairness in artificial intelligence for medical image analysis: identification and mitigation of potential biases in the roadmap from data collection to model deployment.
\newblock \emph{Journal of Medical Imaging}, 10(6):061104--061104.

\bibitem[{Du et~al.(2022)Du, Huang, Dai, Tong, Lepikhin, Xu, Krikun, Zhou, Yu, Firat et~al.}]{du2022glam}
Nan Du, Yanping Huang, Andrew~M Dai, Simon Tong, Dmitry Lepikhin, Yuanzhong Xu, Maxim Krikun, Yanqi Zhou, Adams~Wei Yu, Orhan Firat, et~al. 2022.
\newblock Glam: Efficient scaling of language models with mixture-of-experts.
\newblock In \emph{International Conference on Machine Learning}, pages 5547--5569. PMLR.

\bibitem[{Edenberg and Wood(2023)}]{edenberg2023disambiguating}
Elizabeth Edenberg and Alexandra Wood. 2023.
\newblock Disambiguating algorithmic bias: from neutrality to justice.
\newblock In \emph{Proceedings of the 2023 AAAI/ACM Conference on AI, Ethics, and Society}, pages 691--704.

\bibitem[{El-Kishky et~al.(2020)El-Kishky, Chaudhary, Guzm{\'a}n, and Koehn}]{el-kishky-etal-2020-ccaligned}
Ahmed El-Kishky, Vishrav Chaudhary, Francisco Guzm{\'a}n, and Philipp Koehn. 2020.
\newblock \href {https://doi.org/10.18653/v1/2020.emnlp-main.480} {{CCA}ligned: A massive collection of cross-lingual web-document pairs}.
\newblock In \emph{Proceedings of the 2020 Conference on Empirical Methods in Natural Language Processing (EMNLP)}, pages 5960--5969, Online. Association for Computational Linguistics.

\bibitem[{Elliott et~al.(2016)Elliott, Frank, Sima{'}an, and Specia}]{elliott-etal-2016-multi30k}
Desmond Elliott, Stella Frank, Khalil Sima{'}an, and Lucia Specia. 2016.
\newblock \href {https://doi.org/10.18653/v1/W16-3210} {{M}ulti30{K}: Multilingual {E}nglish-{G}erman image descriptions}.
\newblock In \emph{Proceedings of the 5th Workshop on Vision and Language}, pages 70--74, Berlin, Germany. Association for Computational Linguistics.

\bibitem[{Eloundou et~al.(2023)Eloundou, Manning, Mishkin, and Rock}]{eloundou2023gpts}
Tyna Eloundou, Sam Manning, Pamela Mishkin, and Daniel Rock. 2023.
\newblock Gpts are gpts: An early look at the labor market impact potential of large language models.
\newblock \emph{arXiv preprint arXiv:2303.10130}.

\bibitem[{Escalante et~al.(2020)Escalante, Kaya, Salah, Escalera, Gucluturk, G{\"u}{\c{c}}l{\"u}, Bar{\'o}, Guyon, Junior, Madadi et~al.}]{escalante2020explaining}
Hugo~Jair Escalante, Heysem Kaya, Albert~Ali Salah, Sergio Escalera, Yagmur Gucluturk, Umut G{\"u}{\c{c}}l{\"u}, Xavier Bar{\'o}, Isabelle Guyon, Julio~Jacques Junior, Meysam Madadi, et~al. 2020.
\newblock Explaining first impressions: Modeling, recognizing, and explaining apparent personality from videos.
\newblock \emph{IEEE Transactions on Affective Computing}.

\bibitem[{Escud{\'e}~Font and Costa-juss{\`a}(2019)}]{escude-font-costa-jussa-2019-equalizing}
Joel Escud{\'e}~Font and Marta~R. Costa-juss{\`a}. 2019.
\newblock \href {https://doi.org/10.18653/v1/W19-3821} {Equalizing gender bias in neural machine translation with word embeddings techniques}.
\newblock In \emph{Proceedings of the First Workshop on Gender Bias in Natural Language Processing}, pages 147--154, Florence, Italy. Association for Computational Linguistics.

\bibitem[{Esiobu et~al.(2023)Esiobu, Tan, Hosseini, Ung, Zhang, Fernandes, Dwivedi-Yu, Presani, Williams, and Smith}]{esiobu-etal-2023-robbie}
David Esiobu, Xiaoqing Tan, Saghar Hosseini, Megan Ung, Yuchen Zhang, Jude Fernandes, Jane Dwivedi-Yu, Eleonora Presani, Adina Williams, and Eric Smith. 2023.
\newblock \href {https://doi.org/10.18653/v1/2023.emnlp-main.230} {{ROBBIE}: Robust bias evaluation of large generative language models}.
\newblock In \emph{Proceedings of the 2023 Conference on Empirical Methods in Natural Language Processing}, pages 3764--3814, Singapore. Association for Computational Linguistics.

\bibitem[{Ethayarajh et~al.(2019)Ethayarajh, Duvenaud, and Hirst}]{ethayarajh-etal-2019-understanding}
Kawin Ethayarajh, David Duvenaud, and Graeme Hirst. 2019.
\newblock \href {https://doi.org/10.18653/v1/P19-1166} {Understanding undesirable word embedding associations}.
\newblock In \emph{Proceedings of the 57th Annual Meeting of the Association for Computational Linguistics}, pages 1696--1705, Florence, Italy. Association for Computational Linguistics.

\bibitem[{Fan et~al.(2020)Fan, Esparza, Dargin, Wu, Oztekin, and Mostafavi}]{fan2020spatial}
Chao Fan, Miguel Esparza, Jennifer Dargin, Fangsheng Wu, Bora Oztekin, and Ali Mostafavi. 2020.
\newblock Spatial biases in crowdsourced data: Social media content attention concentrates on populous areas in disasters.
\newblock \emph{Computers, Environment and Urban Systems}, 83:101514.

\bibitem[{Fatemi et~al.(2023)Fatemi, Xing, Liu, and Xiong}]{fatemi-etal-2023-improving}
Zahra Fatemi, Chen Xing, Wenhao Liu, and Caimming Xiong. 2023.
\newblock \href {https://doi.org/10.18653/v1/2023.acl-short.108} {Improving gender fairness of pre-trained language models without catastrophic forgetting}.
\newblock In \emph{Proceedings of the 61st Annual Meeting of the Association for Computational Linguistics (Volume 2: Short Papers)}, pages 1249--1262, Toronto, Canada. Association for Computational Linguistics.

\bibitem[{Fei et~al.(2022)Fei, Lu, Gao, Yang, Huo, Wen, Lu, Song, Gao, Xiang et~al.}]{fei2022towards}
Nanyi Fei, Zhiwu Lu, Yizhao Gao, Guoxing Yang, Yuqi Huo, Jingyuan Wen, Haoyu Lu, Ruihua Song, Xin Gao, Tao Xiang, et~al. 2022.
\newblock Towards artificial general intelligence via a multimodal foundation model.
\newblock \emph{Nature Communications}, 13(1):3094.

\bibitem[{Felkner et~al.(2023)Felkner, Chang, Jang, and May}]{felkner-etal-2023-winoqueer}
Virginia Felkner, Ho-Chun~Herbert Chang, Eugene Jang, and Jonathan May. 2023.
\newblock \href {https://doi.org/10.18653/v1/2023.acl-long.507} {{W}ino{Q}ueer: A community-in-the-loop benchmark for anti-{LGBTQ}+ bias in large language models}.
\newblock In \emph{Proceedings of the 61st Annual Meeting of the Association for Computational Linguistics (Volume 1: Long Papers)}, pages 9126--9140, Toronto, Canada. Association for Computational Linguistics.

\bibitem[{Fenu and Marras(2022)}]{fenu2022demographic}
Gianni Fenu and Mirko Marras. 2022.
\newblock Demographic fairness in multimodal biometrics: A comparative analysis on audio-visual speaker recognition systems.
\newblock \emph{Procedia Computer Science}, 198:249--254.

\bibitem[{Ferrara(2023{\natexlab{a}})}]{ferrara2023fairness}
Emilio Ferrara. 2023{\natexlab{a}}.
\newblock Fairness and bias in artificial intelligence: A brief survey of sources, impacts, and mitigation strategies.
\newblock \emph{Sci}, 6(1):3.

\bibitem[{Ferrara(2023{\natexlab{b}})}]{ferrara2023should}
Emilio Ferrara. 2023{\natexlab{b}}.
\newblock Should chatgpt be biased? challenges and risks of bias in large language models.
\newblock \emph{arXiv preprint arXiv:2304.03738}.

\bibitem[{Folt{\`y}nek et~al.(2019)Folt{\`y}nek, Meuschke, and Gipp}]{foltynek2019academic}
Tom{\'a}{\v{s}} Folt{\`y}nek, Norman Meuschke, and Bela Gipp. 2019.
\newblock Academic plagiarism detection: a systematic literature review.
\newblock \emph{ACM Computing Surveys (CSUR)}, 52(6):1--42.

\bibitem[{Frankel and Vendrow(2020)}]{frankel2020fair}
Eric Frankel and Edward Vendrow. 2020.
\newblock Fair generation through prior modification.
\newblock In \emph{32nd Conference on Neural Information Processing Systems (NeurIPS 2018)}.

\bibitem[{Fredrickson and Roberts(1997)}]{fredrickson1997objectification}
Barbara~L Fredrickson and Tomi-Ann Roberts. 1997.
\newblock Objectification theory: Toward understanding women's lived experiences and mental health risks.
\newblock \emph{Psychology of women quarterly}, 21(2):173--206.

\bibitem[{Freiberger and Buchmann(2024)}]{freiberger2024fairness}
Vincent Freiberger and Erik Buchmann. 2024.
\newblock Fairness certification for natural language processing and large language models.
\newblock In \emph{Intelligent Systems Conference}, pages 606--624. Springer.

\bibitem[{Friedrich et~al.(2023)Friedrich, Brack, Struppek, Hintersdorf, Schramowski, Luccioni, and Kersting}]{friedrich2023fair}
Felix Friedrich, Manuel Brack, Lukas Struppek, Dominik Hintersdorf, Patrick Schramowski, Sasha Luccioni, and Kristian Kersting. 2023.
\newblock Fair diffusion: Instructing text-to-image generation models on fairness.
\newblock \emph{arXiv preprint arXiv:2302.10893}.

\bibitem[{Friedrich et~al.(2021)Friedrich, Lauscher, Ponzetto, and Glava{\v{s}}}]{friedrich-etal-2021-debie}
Niklas Friedrich, Anne Lauscher, Simone~Paolo Ponzetto, and Goran Glava{\v{s}}. 2021.
\newblock \href {https://doi.org/10.18653/v1/2021.eacl-demos.11} {{D}eb{IE}: A platform for implicit and explicit debiasing of word embedding spaces}.
\newblock In \emph{Proceedings of the 16th Conference of the European Chapter of the Association for Computational Linguistics: System Demonstrations}, pages 91--98, Online. Association for Computational Linguistics.

\bibitem[{Gadiraju et~al.(2023)Gadiraju, Kane, Dev, Taylor, Wang, Denton, and Brewer}]{gadiraju2023wouldn}
Vinitha Gadiraju, Shaun Kane, Sunipa Dev, Alex Taylor, Ding Wang, Emily Denton, and Robin Brewer. 2023.
\newblock " i wouldn’t say offensive but...": Disability-centered perspectives on large language models.
\newblock In \emph{Proceedings of the 2023 ACM Conference on Fairness, Accountability, and Transparency}, pages 205--216.

\bibitem[{Gadre et~al.(2024)Gadre, Ilharco, Fang, Hayase, Smyrnis, Nguyen, Marten, Wortsman, Ghosh, Zhang et~al.}]{gadre2024datacomp}
Samir~Yitzhak Gadre, Gabriel Ilharco, Alex Fang, Jonathan Hayase, Georgios Smyrnis, Thao Nguyen, Ryan Marten, Mitchell Wortsman, Dhruba Ghosh, Jieyu Zhang, et~al. 2024.
\newblock Datacomp: In search of the next generation of multimodal datasets.
\newblock \emph{Advances in Neural Information Processing Systems}, 36.

\bibitem[{Gallegos et~al.(2023)Gallegos, Rossi, Barrow, Tanjim, Kim, Dernoncourt, Yu, Zhang, and Ahmed}]{gallegos2023bias}
Isabel~O Gallegos, Ryan~A Rossi, Joe Barrow, Md~Mehrab Tanjim, Sungchul Kim, Franck Dernoncourt, Tong Yu, Ruiyi Zhang, and Nesreen~K Ahmed. 2023.
\newblock Bias and fairness in large language models: A survey.
\newblock \emph{arXiv preprint arXiv:2309.00770}.

\bibitem[{Ganguli et~al.(2022)Ganguli, Hernandez, Lovitt, Askell, Bai, Chen, Conerly, Dassarma, Drain, Elhage et~al.}]{ganguli2022predictability}
Deep Ganguli, Danny Hernandez, Liane Lovitt, Amanda Askell, Yuntao Bai, Anna Chen, Tom Conerly, Nova Dassarma, Dawn Drain, Nelson Elhage, et~al. 2022.
\newblock Predictability and surprise in large generative models.
\newblock In \emph{Proceedings of the 2022 ACM Conference on Fairness, Accountability, and Transparency}, pages 1747--1764.

\bibitem[{Gao et~al.(2020{\natexlab{a}})Gao, Biderman, Black, Golding, Hoppe, Foster, Phang, He, Thite, Nabeshima et~al.}]{gao2020pile}
Leo Gao, Stella Biderman, Sid Black, Laurence Golding, Travis Hoppe, Charles Foster, Jason Phang, Horace He, Anish Thite, Noa Nabeshima, et~al. 2020{\natexlab{a}}.
\newblock The pile: An 800gb dataset of diverse text for language modeling.
\newblock \emph{arXiv preprint arXiv:2101.00027}.

\bibitem[{Gao et~al.(2020{\natexlab{b}})Gao, Fisch, and Chen}]{gao2020making}
Tianyu Gao, Adam Fisch, and Danqi Chen. 2020{\natexlab{b}}.
\newblock Making pre-trained language models better few-shot learners.
\newblock \emph{arXiv preprint arXiv:2012.15723}.

\bibitem[{Garg et~al.(2019)Garg, Perot, Limtiaco, Taly, Chi, and Beutel}]{garg2019counterfactual}
Sahaj Garg, Vincent Perot, Nicole Limtiaco, Ankur Taly, Ed~H Chi, and Alex Beutel. 2019.
\newblock Counterfactual fairness in text classification through robustness.
\newblock In \emph{Proceedings of the 2019 AAAI/ACM Conference on AI, Ethics, and Society}, pages 219--226.

\bibitem[{Garrido-Mu{\~n}oz et~al.(2021)Garrido-Mu{\~n}oz, Montejo-R{\'a}ez, Mart{\'\i}nez-Santiago, and Ure{\~n}a-L{\'o}pez}]{garrido2021survey}
Ismael Garrido-Mu{\~n}oz, Arturo Montejo-R{\'a}ez, Fernando Mart{\'\i}nez-Santiago, and L~Alfonso Ure{\~n}a-L{\'o}pez. 2021.
\newblock A survey on bias in deep nlp.
\newblock \emph{Applied Sciences}, 11(7):3184.

\bibitem[{Gebru et~al.(2021)Gebru, Morgenstern, Vecchione, Vaughan, Wallach, Iii, and Crawford}]{gebru2021datasheets}
Timnit Gebru, Jamie Morgenstern, Briana Vecchione, Jennifer~Wortman Vaughan, Hanna Wallach, Hal~Daum{\'e} Iii, and Kate Crawford. 2021.
\newblock Datasheets for datasets.
\newblock \emph{Communications of the ACM}, 64(12):86--92.

\bibitem[{Gehman et~al.(2020)Gehman, Gururangan, Sap, Choi, and Smith}]{gehman-etal-2020-realtoxicityprompts}
Samuel Gehman, Suchin Gururangan, Maarten Sap, Yejin Choi, and Noah~A. Smith. 2020.
\newblock \href {https://doi.org/10.18653/v1/2020.findings-emnlp.301} {{R}eal{T}oxicity{P}rompts: Evaluating neural toxic degeneration in language models}.
\newblock In \emph{Findings of the Association for Computational Linguistics: EMNLP 2020}, pages 3356--3369, Online. Association for Computational Linguistics.

\bibitem[{Georgopoulos et~al.(2021)Georgopoulos, Oldfield, Nicolaou, Panagakis, and Pantic}]{georgopoulos2021mitigating}
Markos Georgopoulos, James Oldfield, Mihalis~A Nicolaou, Yannis Panagakis, and Maja Pantic. 2021.
\newblock Mitigating demographic bias in facial datasets with style-based multi-attribute transfer.
\newblock \emph{International Journal of Computer Vision}, 129(7):2288--2307.

\bibitem[{Giner-Miguelez et~al.(2023)Giner-Miguelez, G{\'o}mez, and Cabot}]{giner2023datadoc}
Joan Giner-Miguelez, Abel G{\'o}mez, and Jordi Cabot. 2023.
\newblock Datadoc analyzer: A tool for analyzing the documentation of scientific datasets.
\newblock In \emph{Proceedings of the 32nd ACM International Conference on Information and Knowledge Management}, pages 5046--5050.

\bibitem[{Goyal et~al.(2022{\natexlab{a}})Goyal, Duval, Seessel, Caron, Misra, Sagun, Joulin, and Bojanowski}]{goyal2022vision}
Priya Goyal, Quentin Duval, Isaac Seessel, Mathilde Caron, Ishan Misra, Levent Sagun, Armand Joulin, and Piotr Bojanowski. 2022{\natexlab{a}}.
\newblock Vision models are more robust and fair when pretrained on uncurated images without supervision.
\newblock \emph{arXiv preprint arXiv:2202.08360}.

\bibitem[{Goyal et~al.(2022{\natexlab{b}})Goyal, Soriano, Hazirbas, Sagun, and Usunier}]{goyal2022fairness}
Priya Goyal, Adriana~Romero Soriano, Caner Hazirbas, Levent Sagun, and Nicolas Usunier. 2022{\natexlab{b}}.
\newblock Fairness indicators for systematic assessments of visual feature extractors.
\newblock In \emph{Proceedings of the 2022 ACM Conference on Fairness, Accountability, and Transparency}, pages 70--88.

\bibitem[{Goyal et~al.(2017)Goyal, Khot, Summers-Stay, Batra, and Parikh}]{goyal2017making}
Yash Goyal, Tejas Khot, Douglas Summers-Stay, Dhruv Batra, and Devi Parikh. 2017.
\newblock Making the v in vqa matter: Elevating the role of image understanding in visual question answering.
\newblock In \emph{Proceedings of the IEEE conference on computer vision and pattern recognition}, pages 6904--6913.

\bibitem[{Greenberg(1990)}]{greenberg1990organizational}
Jerald Greenberg. 1990.
\newblock Organizational justice: Yesterday, today, and tomorrow.
\newblock \emph{Journal of management}, 16(2):399--432.

\bibitem[{Guo and Caliskan(2021)}]{guo2021detecting}
Wei Guo and Aylin Caliskan. 2021.
\newblock Detecting emergent intersectional biases: Contextualized word embeddings contain a distribution of human-like biases.
\newblock In \emph{Proceedings of the 2021 AAAI/ACM Conference on AI, Ethics, and Society}, pages 122--133.

\bibitem[{Guo et~al.(2022)Guo, Yang, and Abbasi}]{guo-etal-2022-auto}
Yue Guo, Yi~Yang, and Ahmed Abbasi. 2022.
\newblock \href {https://doi.org/10.18653/v1/2022.acl-long.72} {Auto-debias: Debiasing masked language models with automated biased prompts}.
\newblock In \emph{Proceedings of the 60th Annual Meeting of the Association for Computational Linguistics (Volume 1: Long Papers)}, pages 1012--1023, Dublin, Ireland. Association for Computational Linguistics.

\bibitem[{Hager et~al.(2024)Hager, Jungmann, Holland, Bhagat, Hubrecht, Knauer, Vielhauer, Makowski, Braren, Kaissis et~al.}]{hager2024evaluation}
Paul Hager, Friederike Jungmann, Robbie Holland, Kunal Bhagat, Inga Hubrecht, Manuel Knauer, Jakob Vielhauer, Marcus Makowski, Rickmer Braren, Georgios Kaissis, et~al. 2024.
\newblock Evaluation and mitigation of the limitations of large language models in clinical decision-making.
\newblock \emph{Nature medicine}, pages 1--10.

\bibitem[{Halevy et~al.(2021)Halevy, Harris, Bruckman, Yang, and Howard}]{10.1145/3465416.3483299}
Matan Halevy, Camille Harris, Amy Bruckman, Diyi Yang, and Ayanna Howard. 2021.
\newblock \href {https://doi.org/10.1145/3465416.3483299} {Mitigating racial biases in toxic language detection with an equity-based ensemble framework}.
\newblock In \emph{Proceedings of the 1st ACM Conference on Equity and Access in Algorithms, Mechanisms, and Optimization}, EAAMO '21, New York, NY, USA. Association for Computing Machinery.

\bibitem[{Haltaufderheide and Ranisch(2024)}]{haltaufderheide2024ethics}
Joschka Haltaufderheide and Robert Ranisch. 2024.
\newblock The ethics of chatgpt in medicine and healthcare: a systematic review on large language models (llms).
\newblock \emph{NPJ Digital Medicine}, 7(1):183.

\bibitem[{Han(2023)}]{han2023fairness}
Yuchen Han. 2023.
\newblock Fairness evaluation within large language models through the lens of depression.
\newblock In \emph{Proceedings of the 2023 4th International Conference on Machine Learning and Computer Application}, pages 108--112.

\bibitem[{Hao et~al.(2024)Hao, von Davier, Yaneva, Lottridge, von Davier, and Harris}]{hao2024transforming}
Jiangang Hao, Alina~A von Davier, Victoria Yaneva, Susan Lottridge, Matthias von Davier, and Deborah~J Harris. 2024.
\newblock Transforming assessment: The impacts and implications of large language models and generative ai.
\newblock \emph{Educational Measurement: Issues and Practice}, 43(2):16--29.

\bibitem[{Harrer(2023)}]{harrer2023attention}
Stefan Harrer. 2023.
\newblock Attention is not all you need: the complicated case of ethically using large language models in healthcare and medicine.
\newblock \emph{EBioMedicine}, 90.

\bibitem[{He et~al.(2016)He, Zhang, Ren, and Sun}]{he2016deep}
Kaiming He, Xiangyu Zhang, Shaoqing Ren, and Jian Sun. 2016.
\newblock Deep residual learning for image recognition.
\newblock In \emph{Proceedings of the IEEE conference on computer vision and pattern recognition}, pages 770--778.

\bibitem[{Heflick et~al.(2011)Heflick, Goldenberg, Cooper, and Puvia}]{heflick2011women}
Nathan~A Heflick, Jamie~L Goldenberg, Douglas~P Cooper, and Elisa Puvia. 2011.
\newblock From women to objects: Appearance focus, target gender, and perceptions of warmth, morality and competence.
\newblock \emph{Journal of Experimental Social Psychology}, 47(3):572--581.

\bibitem[{Hinton et~al.(2015)Hinton, Vinyals, and Dean}]{hinton2015distilling}
Geoffrey Hinton, Oriol Vinyals, and Jeff Dean. 2015.
\newblock Distilling the knowledge in a neural network.
\newblock \emph{arXiv preprint arXiv:1503.02531}.

\bibitem[{Hirota et~al.(2022{\natexlab{a}})Hirota, Nakashima, and Garcia}]{hirota2022gender}
Yusuke Hirota, Yuta Nakashima, and Noa Garcia. 2022{\natexlab{a}}.
\newblock Gender and racial bias in visual question answering datasets.
\newblock In \emph{Proceedings of the 2022 ACM Conference on Fairness, Accountability, and Transparency}, pages 1280--1292.

\bibitem[{Hirota et~al.(2022{\natexlab{b}})Hirota, Nakashima, and Garcia}]{Hirota_2022_CVPR}
Yusuke Hirota, Yuta Nakashima, and Noa Garcia. 2022{\natexlab{b}}.
\newblock Quantifying societal bias amplification in image captioning.
\newblock In \emph{Proceedings of the IEEE/CVF Conference on Computer Vision and Pattern Recognition (CVPR)}, pages 13450--13459.

\bibitem[{Hoffmann et~al.(2022)Hoffmann, Borgeaud, Mensch, Buchatskaya, Cai, Rutherford, de~Las~Casas, Hendricks, Welbl, Clark et~al.}]{hoffmann2022empirical}
Jordan Hoffmann, Sebastian Borgeaud, Arthur Mensch, Elena Buchatskaya, Trevor Cai, Eliza Rutherford, Diego de~Las~Casas, Lisa~Anne Hendricks, Johannes Welbl, Aidan Clark, et~al. 2022.
\newblock An empirical analysis of compute-optimal large language model training.
\newblock \emph{Advances in Neural Information Processing Systems}, 35:30016--30030.

\bibitem[{Hort et~al.(2024)Hort, Chen, Zhang, Harman, and Sarro}]{hort2024bias}
Max Hort, Zhenpeng Chen, Jie~M Zhang, Mark Harman, and Federica Sarro. 2024.
\newblock Bias mitigation for machine learning classifiers: A comprehensive survey.
\newblock \emph{ACM Journal on Responsible Computing}, 1(2):1--52.

\bibitem[{Hou et~al.(2024)Hou, Zhang, Lin, Lu, Xie, McAuley, and Zhao}]{hou2024large}
Yupeng Hou, Junjie Zhang, Zihan Lin, Hongyu Lu, Ruobing Xie, Julian McAuley, and Wayne~Xin Zhao. 2024.
\newblock Large language models are zero-shot rankers for recommender systems.
\newblock In \emph{European Conference on Information Retrieval}, pages 364--381. Springer.

\bibitem[{Houlsby et~al.(2019)Houlsby, Giurgiu, Jastrzebski, Morrone, De~Laroussilhe, Gesmundo, Attariyan, and Gelly}]{pmlr-v97-houlsby19a}
Neil Houlsby, Andrei Giurgiu, Stanislaw Jastrzebski, Bruna Morrone, Quentin De~Laroussilhe, Andrea Gesmundo, Mona Attariyan, and Sylvain Gelly. 2019.
\newblock \href {https://proceedings.mlr.press/v97/houlsby19a.html} {Parameter-efficient transfer learning for {NLP}}.
\newblock In \emph{Proceedings of the 36th International Conference on Machine Learning}, volume~97 of \emph{Proceedings of Machine Learning Research}, pages 2790--2799. PMLR.

\bibitem[{Hovy and Prabhumoye(2021)}]{hovy2021five}
Dirk Hovy and Shrimai Prabhumoye. 2021.
\newblock Five sources of bias in natural language processing.
\newblock \emph{Language and linguistics compass}, 15(8):e12432.

\bibitem[{Hu et~al.(2021)Hu, Shen, Wallis, Allen-Zhu, Li, Wang, Wang, and Chen}]{hu2021lora}
Edward~J Hu, Yelong Shen, Phillip Wallis, Zeyuan Allen-Zhu, Yuanzhi Li, Shean Wang, Lu~Wang, and Weizhu Chen. 2021.
\newblock Lora: Low-rank adaptation of large language models.
\newblock \emph{arXiv preprint arXiv:2106.09685}.

\bibitem[{Huang et~al.(2020{\natexlab{a}})Huang, Zhang, Jiang, Stanforth, Welbl, Rae, Maini, Yogatama, and Kohli}]{huang-etal-2020-reducing}
Po-Sen Huang, Huan Zhang, Ray Jiang, Robert Stanforth, Johannes Welbl, Jack Rae, Vishal Maini, Dani Yogatama, and Pushmeet Kohli. 2020{\natexlab{a}}.
\newblock \href {https://doi.org/10.18653/v1/2020.findings-emnlp.7} {Reducing sentiment bias in language models via counterfactual evaluation}.
\newblock In \emph{Findings of the Association for Computational Linguistics: EMNLP 2020}, pages 65--83, Online. Association for Computational Linguistics.

\bibitem[{Huang et~al.(2020{\natexlab{b}})Huang, Xing, Dernoncourt, and Paul}]{huang-etal-2020-multilingual}
Xiaolei Huang, Linzi Xing, Franck Dernoncourt, and Michael~J. Paul. 2020{\natexlab{b}}.
\newblock \href {https://aclanthology.org/2020.lrec-1.180} {Multilingual {T}witter corpus and baselines for evaluating demographic bias in hate speech recognition}.
\newblock In \emph{Proceedings of the Twelfth Language Resources and Evaluation Conference}, pages 1440--1448, Marseille, France. European Language Resources Association.

\bibitem[{Huang et~al.(2023)Huang, Song, Wang, Zhao, Chen, Juefei-Xu, and Ma}]{huang2023look}
Yuheng Huang, Jiayang Song, Zhijie Wang, Shengming Zhao, Huaming Chen, Felix Juefei-Xu, and Lei Ma. 2023.
\newblock Look before you leap: An exploratory study of uncertainty measurement for large language models.
\newblock \emph{arXiv preprint arXiv:2307.10236}.

\bibitem[{Hutchinson and Mitchell(2019)}]{hutchinson201950}
Ben Hutchinson and Margaret Mitchell. 2019.
\newblock 50 years of test (un) fairness: Lessons for machine learning.
\newblock In \emph{Proceedings of the conference on fairness, accountability, and transparency}, pages 49--58.

\bibitem[{Hutchinson et~al.(2020)Hutchinson, Prabhakaran, Denton, Webster, Zhong, and Denuyl}]{hutchinson-etal-2020-social}
Ben Hutchinson, Vinodkumar Prabhakaran, Emily Denton, Kellie Webster, Yu~Zhong, and Stephen Denuyl. 2020.
\newblock \href {https://doi.org/10.18653/v1/2020.acl-main.487} {Social biases in {NLP} models as barriers for persons with disabilities}.
\newblock In \emph{Proceedings of the 58th Annual Meeting of the Association for Computational Linguistics}, pages 5491--5501, Online. Association for Computational Linguistics.

\bibitem[{Jaiswal and Provost(2020)}]{jaiswal2020privacy}
Mimansa Jaiswal and Emily~Mower Provost. 2020.
\newblock Privacy enhanced multimodal neural representations for emotion recognition.
\newblock In \emph{Proceedings of the AAAI Conference on Artificial Intelligence}, volume~34, pages 7985--7993.

\bibitem[{Jamil(2024)}]{jamil2024equity}
Hasan Jamil. 2024.
\newblock Equity and fairness challenges in online learning in the age of chatgpt.
\newblock In \emph{Proceedings of the 39th ACM/SIGAPP Symposium on Applied Computing}, pages 91--92.

\bibitem[{Jenks(2024)}]{jenks2024communicating}
Christopher~J Jenks. 2024.
\newblock Communicating the cultural other: Trust and bias in generative ai and large language models.
\newblock \emph{Applied Linguistics Review}, (0).

\bibitem[{Jeong et~al.(2024)Jeong, Yang, Choi, and Lee}]{10459082}
Yongwoo Jeong, Jiseon Yang, In~Ho Choi, and Juyeon Lee. 2024.
\newblock \href {https://doi.org/10.1109/ACCESS.2024.3373470} {Feature-based text search engine mitigating data diversity problem using pre-trained large language model for fast deployment services}.
\newblock \emph{IEEE Access}, 12:48145--48157.

\bibitem[{Ji et~al.(2024)Ji, Liu, Dai, Pan, Zhang, Bian, Chen, Sun, Wang, and Yang}]{ji2024beavertails}
Jiaming Ji, Mickel Liu, Josef Dai, Xuehai Pan, Chi Zhang, Ce~Bian, Boyuan Chen, Ruiyang Sun, Yizhou Wang, and Yaodong Yang. 2024.
\newblock Beavertails: Towards improved safety alignment of llm via a human-preference dataset.
\newblock \emph{Advances in Neural Information Processing Systems}, 36.

\bibitem[{Jia et~al.(2021)Jia, Yang, Xia, Chen, Parekh, Pham, Le, Sung, Li, and Duerig}]{jia2021scaling}
Chao Jia, Yinfei Yang, Ye~Xia, Yi-Ting Chen, Zarana Parekh, Hieu Pham, Quoc Le, Yun-Hsuan Sung, Zhen Li, and Tom Duerig. 2021.
\newblock Scaling up visual and vision-language representation learning with noisy text supervision.
\newblock In \emph{International conference on machine learning}, pages 4904--4916. PMLR.

\bibitem[{Jiang et~al.(2024{\natexlab{a}})Jiang, Sablayrolles, Roux, Mensch, Savary, Bamford, Chaplot, Casas, Hanna, Bressand et~al.}]{jiang2024mixtral}
Albert~Q Jiang, Alexandre Sablayrolles, Antoine Roux, Arthur Mensch, Blanche Savary, Chris Bamford, Devendra~Singh Chaplot, Diego de~las Casas, Emma~Bou Hanna, Florian Bressand, et~al. 2024{\natexlab{a}}.
\newblock Mixtral of experts.
\newblock \emph{arXiv preprint arXiv:2401.04088}.

\bibitem[{Jiang et~al.(2021)Jiang, Araki, Ding, and Neubig}]{jiang2021can}
Zhengbao Jiang, Jun Araki, Haibo Ding, and Graham Neubig. 2021.
\newblock How can we know when language models know? on the calibration of language models for question answering.
\newblock \emph{Transactions of the Association for Computational Linguistics}, 9:962--977.

\bibitem[{Jiang et~al.(2024{\natexlab{b}})Jiang, Seyedi, Griner, Abbasi, Rad, Kwon, Cotes, and Clifford}]{jiang2024evaluating}
Zifan Jiang, Salman Seyedi, Emily Griner, Ahmed Abbasi, Ali~Bahrami Rad, Hyeokhyen Kwon, Robert~O Cotes, and Gari~D Clifford. 2024{\natexlab{b}}.
\newblock Evaluating and mitigating unfairness in multimodal remote mental health assessments.
\newblock \emph{PLOS Digital Health}, 3(7):e0000413.

\bibitem[{Jo and Gebru(2020)}]{jo2020lessons}
Eun~Seo Jo and Timnit Gebru. 2020.
\newblock Lessons from archives: Strategies for collecting sociocultural data in machine learning.
\newblock In \emph{Proceedings of the 2020 conference on fairness, accountability, and transparency}, pages 306--316.

\bibitem[{Johnson et~al.(2023)Johnson, Bulgarelli, Shen, Gayles, Shammout, Horng, Pollard, Hao, Moody, Gow et~al.}]{johnson2023mimic}
Alistair~EW Johnson, Lucas Bulgarelli, Lu~Shen, Alvin Gayles, Ayad Shammout, Steven Horng, Tom~J Pollard, Sicheng Hao, Benjamin Moody, Brian Gow, et~al. 2023.
\newblock Mimic-iv, a freely accessible electronic health record dataset.
\newblock \emph{Scientific data}, 10(1):1.

\bibitem[{Joshi et~al.(2021)Joshi, Walambe, and Kotecha}]{joshi2021review}
Gargi Joshi, Rahee Walambe, and Ketan Kotecha. 2021.
\newblock A review on explainability in multimodal deep neural nets.
\newblock \emph{IEEE Access}, 9:59800--59821.

\bibitem[{Karabacak and Margetis(2023)}]{karabacak2023embracing}
Mert Karabacak and Konstantinos Margetis. 2023.
\newblock Embracing large language models for medical applications: opportunities and challenges.
\newblock \emph{Cureus}, 15(5).

\bibitem[{Karkkainen and Joo(2021)}]{karkkainen2021fairface}
Kimmo Karkkainen and Jungseock Joo. 2021.
\newblock Fairface: Face attribute dataset for balanced race, gender, and age for bias measurement and mitigation.
\newblock In \emph{Proceedings of the IEEE/CVF winter conference on applications of computer vision}, pages 1548--1558.

\bibitem[{Kasneci et~al.(2023)Kasneci, Se{\ss}ler, K{\"u}chemann, Bannert, Dementieva, Fischer, Gasser, Groh, G{\"u}nnemann, H{\"u}llermeier et~al.}]{kasneci2023chatgpt}
Enkelejda Kasneci, Kathrin Se{\ss}ler, Stefan K{\"u}chemann, Maria Bannert, Daryna Dementieva, Frank Fischer, Urs Gasser, Georg Groh, Stephan G{\"u}nnemann, Eyke H{\"u}llermeier, et~al. 2023.
\newblock Chatgpt for good? on opportunities and challenges of large language models for education.
\newblock \emph{Learning and individual differences}, 103:102274.

\bibitem[{Kehrenberg et~al.(2020)Kehrenberg, Bartlett, Thomas, and Quadrianto}]{10.1007/978-3-030-58574-7_34}
Thomas Kehrenberg, Myles Bartlett, Oliver Thomas, and Novi Quadrianto. 2020.
\newblock Null-sampling for interpretable and fair representations.
\newblock In \emph{Computer Vision -- ECCV 2020}, pages 565--580, Cham. Springer International Publishing.

\bibitem[{Keskar et~al.(2019)Keskar, McCann, Varshney, Xiong, and Socher}]{keskar2019ctrl}
Nitish~Shirish Keskar, Bryan McCann, Lav~R Varshney, Caiming Xiong, and Richard Socher. 2019.
\newblock Ctrl: A conditional transformer language model for controllable generation.
\newblock \emph{arXiv preprint arXiv:1909.05858}.

\bibitem[{Khanuja et~al.(2023)Khanuja, Ruder, and Talukdar}]{52742}
Simran Khanuja, Sebastian Ruder, and Partha Talukdar. 2023.
\newblock \href {https://aclanthology.org/2023.findings-eacl.131/} {Evaluating inclusivity, equity, and accessibility of nlp technology: A case study for indian languages}.
\newblock In \emph{Findings of the Association for Computational Linguistics: EACL 2023}.

\bibitem[{Kirkpatrick et~al.(2017)Kirkpatrick, Pascanu, Rabinowitz, Veness, Desjardins, Rusu, Milan, Quan, Ramalho, Grabska-Barwinska et~al.}]{kirkpatrick2017overcoming}
James Kirkpatrick, Razvan Pascanu, Neil Rabinowitz, Joel Veness, Guillaume Desjardins, Andrei~A Rusu, Kieran Milan, John Quan, Tiago Ramalho, Agnieszka Grabska-Barwinska, et~al. 2017.
\newblock Overcoming catastrophic forgetting in neural networks.
\newblock \emph{Proceedings of the national academy of sciences}, 114(13):3521--3526.

\bibitem[{Kiros et~al.(2014)Kiros, Salakhutdinov, and Zemel}]{pmlr-v32-kiros14}
Ryan Kiros, Ruslan Salakhutdinov, and Rich Zemel. 2014.
\newblock \href {https://proceedings.mlr.press/v32/kiros14.html} {Multimodal neural language models}.
\newblock In \emph{Proceedings of the 31st International Conference on Machine Learning}, volume~32 of \emph{Proceedings of Machine Learning Research}, pages 595--603, Bejing, China. PMLR.

\bibitem[{Kitchenham(2004)}]{kitchenham2004procedures}
Barbara Kitchenham. 2004.
\newblock Procedures for performing systematic reviews.
\newblock \emph{Keele, UK, Keele University}, 33(2004):1--26.

\bibitem[{Koh et~al.(2024)Koh, Fried, and Salakhutdinov}]{koh2024generating}
Jing~Yu Koh, Daniel Fried, and Russ~R Salakhutdinov. 2024.
\newblock Generating images with multimodal language models.
\newblock \emph{Advances in Neural Information Processing Systems}, 36.

\bibitem[{Kojima et~al.(2022)Kojima, Gu, Reid, Matsuo, and Iwasawa}]{kojima2022large}
Takeshi Kojima, Shixiang~Shane Gu, Machel Reid, Yutaka Matsuo, and Yusuke Iwasawa. 2022.
\newblock Large language models are zero-shot reasoners.
\newblock \emph{Advances in neural information processing systems}, 35:22199--22213.

\bibitem[{Kotek et~al.(2023)Kotek, Dockum, and Sun}]{kotek2023gender}
Hadas Kotek, Rikker Dockum, and David Sun. 2023.
\newblock Gender bias and stereotypes in large language models.
\newblock In \emph{Proceedings of the ACM collective intelligence conference}, pages 12--24.

\bibitem[{Kraft et~al.(2022)Kraft, Zorn, Fecht, Simon, Biemann, and Usbeck}]{kraft2022measuring}
Angelie Kraft, Hans-Peter Zorn, Pascal Fecht, Judith Simon, Chris Biemann, and Ricardo Usbeck. 2022.
\newblock Measuring gender bias in german language generation.
\newblock \emph{INFORMATIK 2022}.

\bibitem[{Kumar et~al.(2023)Kumar, Lesota, Zerveas, Cohen, Eickhoff, Schedl, and Rekabsaz}]{kumar-etal-2023-parameter}
Deepak Kumar, Oleg Lesota, George Zerveas, Daniel Cohen, Carsten Eickhoff, Markus Schedl, and Navid Rekabsaz. 2023.
\newblock \href {https://doi.org/10.18653/v1/2023.eacl-main.201} {Parameter-efficient modularised bias mitigation via {A}dapter{F}usion}.
\newblock In \emph{Proceedings of the 17th Conference of the European Chapter of the Association for Computational Linguistics}, pages 2738--2751, Dubrovnik, Croatia. Association for Computational Linguistics.

\bibitem[{Kumar et~al.(2024)Kumar, Huang, Perez, Yang, Li, Morreale, Kruger, and Jiang}]{kumar2024bias}
Yulia Kumar, Kuan Huang, Angelo Perez, Guohao Yang, J~Jenny Li, Patricia Morreale, Dov Kruger, and Raymond Jiang. 2024.
\newblock Bias and cyberbullying detection and data generation with transformer ai models and top llms.

\bibitem[{Kumari and Ekbal(2021)}]{KUMARI2021115412}
Rina Kumari and Asif Ekbal. 2021.
\newblock \href {https://doi.org/https://doi.org/10.1016/j.eswa.2021.115412} {Amfb: Attention based multimodal factorized bilinear pooling for multimodal fake news detection}.
\newblock \emph{Expert Systems with Applications}, 184:115412.

\bibitem[{Kurita et~al.(2019)Kurita, Vyas, Pareek, Black, and Tsvetkov}]{kurita-etal-2019-measuring}
Keita Kurita, Nidhi Vyas, Ayush Pareek, Alan~W Black, and Yulia Tsvetkov. 2019.
\newblock \href {https://doi.org/10.18653/v1/W19-3823} {Measuring bias in contextualized word representations}.
\newblock In \emph{Proceedings of the First Workshop on Gender Bias in Natural Language Processing}, pages 166--172, Florence, Italy. Association for Computational Linguistics.

\bibitem[{Kurpicz-Briki(2020)}]{kurpicz2020cultural}
Mascha Kurpicz-Briki. 2020.
\newblock Cultural differences in bias? origin and gender bias in pre-trained german and french word embeddings.

\bibitem[{Landers and Behrend(2023)}]{landers2023auditing}
Richard~N Landers and Tara~S Behrend. 2023.
\newblock Auditing the ai auditors: A framework for evaluating fairness and bias in high stakes ai predictive models.
\newblock \emph{American Psychologist}, 78(1):36.

\bibitem[{Lauren{\c{c}}on et~al.(2022)Lauren{\c{c}}on, Saulnier, Wang, Akiki, Villanova~del Moral, Le~Scao, Von~Werra, Mou, Gonz{\'a}lez~Ponferrada, Nguyen et~al.}]{laurenccon2022bigscience}
Hugo Lauren{\c{c}}on, Lucile Saulnier, Thomas Wang, Christopher Akiki, Albert Villanova~del Moral, Teven Le~Scao, Leandro Von~Werra, Chenghao Mou, Eduardo Gonz{\'a}lez~Ponferrada, Huu Nguyen, et~al. 2022.
\newblock The bigscience roots corpus: A 1.6 tb composite multilingual dataset.
\newblock \emph{Advances in Neural Information Processing Systems}, 35:31809--31826.

\bibitem[{Lauscher et~al.(2021)Lauscher, Lueken, and Glava{\v{s}}}]{lauscher-etal-2021-sustainable-modular}
Anne Lauscher, Tobias Lueken, and Goran Glava{\v{s}}. 2021.
\newblock \href {https://doi.org/10.18653/v1/2021.findings-emnlp.411} {Sustainable modular debiasing of language models}.
\newblock In \emph{Findings of the Association for Computational Linguistics: EMNLP 2021}, pages 4782--4797, Punta Cana, Dominican Republic. Association for Computational Linguistics.

\bibitem[{Le~Quy et~al.(2022)Le~Quy, Roy, Iosifidis, Zhang, and Ntoutsi}]{le2022survey}
Tai Le~Quy, Arjun Roy, Vasileios Iosifidis, Wenbin Zhang, and Eirini Ntoutsi. 2022.
\newblock A survey on datasets for fairness-aware machine learning.
\newblock \emph{Wiley Interdisciplinary Reviews: Data Mining and Knowledge Discovery}, 12(3):e1452.

\bibitem[{Le~Scao et~al.(2023)Le~Scao, Fan, Akiki, Pavlick, Ili{\'c}, Hesslow, Castagn{\'e}, Luccioni, Yvon, Gall{\'e} et~al.}]{le2023bloom}
Teven Le~Scao, Angela Fan, Christopher Akiki, Ellie Pavlick, Suzana Ili{\'c}, Daniel Hesslow, Roman Castagn{\'e}, Alexandra~Sasha Luccioni, Fran{\c{c}}ois Yvon, Matthias Gall{\'e}, et~al. 2023.
\newblock Bloom: A 176b-parameter open-access multilingual language model.

\bibitem[{Lee et~al.(2024)Lee, Hicke, Yu, Brooks, and Kizilcec}]{lee2024life}
Jinsook Lee, Yann Hicke, Renzhe Yu, Christopher Brooks, and Ren{\'e}~F Kizilcec. 2024.
\newblock The life cycle of large language models in education: A framework for understanding sources of bias.
\newblock \emph{British Journal of Educational Technology}.

\bibitem[{Lee et~al.(2023)Lee, Yasunaga, Meng, Mai, Park, Gupta, Zhang, Narayanan, Teufel, Bellagente, Kang, Park, Leskovec, Zhu, Li, Wu, Ermon, and Liang}]{NEURIPS2023_dd83eada}
Tony Lee, Michihiro Yasunaga, Chenlin Meng, Yifan Mai, Joon~Sung Park, Agrim Gupta, Yunzhi Zhang, Deepak Narayanan, Hannah Teufel, Marco Bellagente, Minguk Kang, Taesung Park, Jure Leskovec, Jun-Yan Zhu, Fei-Fei Li, Jiajun Wu, Stefano Ermon, and Percy~S Liang. 2023.
\newblock \href {https://proceedings.neurips.cc/paper_files/paper/2023/file/dd83eada2c3c74db3c7fe1c087513756-Paper-Datasets_and_Benchmarks.pdf} {Holistic evaluation of text-to-image models}.
\newblock In \emph{Advances in Neural Information Processing Systems}, volume~36, pages 69981--70011. Curran Associates, Inc.

\bibitem[{Lehman et~al.(2023)Lehman, Hernandez, Mahajan, Wulff, Smith, Ziegler, Nadler, Szolovits, Johnson, and Alsentzer}]{lehman2023we}
Eric Lehman, Evan Hernandez, Diwakar Mahajan, Jonas Wulff, Micah~J Smith, Zachary Ziegler, Daniel Nadler, Peter Szolovits, Alistair Johnson, and Emily Alsentzer. 2023.
\newblock Do we still need clinical language models?
\newblock In \emph{Conference on health, inference, and learning}, pages 578--597. PMLR.

\bibitem[{Leteno et~al.(2023)Leteno, Gourru, Laclau, and Gravier}]{leteno2023investigation}
Thibaud Leteno, Antoine Gourru, Charlotte Laclau, and Christophe Gravier. 2023.
\newblock An investigation of structures responsible for gender bias in bert and distilbert.
\newblock In \emph{International Symposium on Intelligent Data Analysis}, pages 249--261. Springer.

\bibitem[{Leventhal(1980)}]{leventhal1980should}
Gerald~S Leventhal. 1980.
\newblock What should be done with equity theory? new approaches to the study of fairness in social relationships.
\newblock In \emph{Social exchange: Advances in theory and research}, pages 27--55. Springer.

\bibitem[{Li et~al.(2021)Li, Ataman, and Sennrich}]{li-etal-2021-vision}
Jiaoda Li, Duygu Ataman, and Rico Sennrich. 2021.
\newblock \href {https://doi.org/10.18653/v1/2021.emnlp-main.673} {Vision matters when it should: Sanity checking multimodal machine translation models}.
\newblock In \emph{Proceedings of the 2021 Conference on Empirical Methods in Natural Language Processing}, pages 8556--8562, Online and Punta Cana, Dominican Republic. Association for Computational Linguistics.

\bibitem[{Li et~al.(2022)Li, Li, Xiong, and Hoi}]{li2022blip}
Junnan Li, Dongxu Li, Caiming Xiong, and Steven Hoi. 2022.
\newblock Blip: Bootstrapping language-image pre-training for unified vision-language understanding and generation.
\newblock In \emph{International conference on machine learning}, pages 12888--12900. PMLR.

\bibitem[{Li et~al.(2024)Li, Zhang, and Zhang}]{10480206}
Yunqi Li, Lanjing Zhang, and Yongfeng Zhang. 2024.
\newblock \href {https://doi.org/10.1109/CISS59072.2024.10480206} {Probing into the fairness of large language models: A case study of chatgpt}.
\newblock In \emph{2024 58th Annual Conference on Information Sciences and Systems (CISS)}, pages 1--6.

\bibitem[{Liang et~al.(2020)Liang, Li, Zheng, Lim, Salakhutdinov, and Morency}]{liang-etal-2020-towards}
Paul~Pu Liang, Irene~Mengze Li, Emily Zheng, Yao~Chong Lim, Ruslan Salakhutdinov, and Louis-Philippe Morency. 2020.
\newblock \href {https://doi.org/10.18653/v1/2020.acl-main.488} {Towards debiasing sentence representations}.
\newblock In \emph{Proceedings of the 58th Annual Meeting of the Association for Computational Linguistics}, pages 5502--5515, Online. Association for Computational Linguistics.

\bibitem[{Liang et~al.(2021{\natexlab{a}})Liang, Lyu, Fan, Wu, Cheng, Wu, Chen, Wu, Lee, Zhu et~al.}]{liang2021multibench}
Paul~Pu Liang, Yiwei Lyu, Xiang Fan, Zetian Wu, Yun Cheng, Jason Wu, Leslie Chen, Peter Wu, Michelle~A Lee, Yuke Zhu, et~al. 2021{\natexlab{a}}.
\newblock Multibench: Multiscale benchmarks for multimodal representation learning.
\newblock \emph{Advances in neural information processing systems}, 2021(DB1):1.

\bibitem[{Liang et~al.(2021{\natexlab{b}})Liang, Wu, Morency, and Salakhutdinov}]{liang2021towards}
Paul~Pu Liang, Chiyu Wu, Louis-Philippe Morency, and Ruslan Salakhutdinov. 2021{\natexlab{b}}.
\newblock Towards understanding and mitigating social biases in language models.
\newblock In \emph{International Conference on Machine Learning}, pages 6565--6576. PMLR.

\bibitem[{Liang et~al.(2021{\natexlab{c}})Liang, Wu, Morency, and Salakhutdinov}]{pmlr-v139-liang21a}
Paul~Pu Liang, Chiyu Wu, Louis-Philippe Morency, and Ruslan Salakhutdinov. 2021{\natexlab{c}}.
\newblock \href {https://proceedings.mlr.press/v139/liang21a.html} {Towards understanding and mitigating social biases in language models}.
\newblock In \emph{Proceedings of the 38th International Conference on Machine Learning}, volume 139 of \emph{Proceedings of Machine Learning Research}, pages 6565--6576. PMLR.

\bibitem[{Liang et~al.(2022{\natexlab{a}})Liang, Zadeh, and Morency}]{liang2022foundations}
Paul~Pu Liang, Amir Zadeh, and Louis-Philippe Morency. 2022{\natexlab{a}}.
\newblock Foundations and trends in multimodal machine learning: Principles, challenges, and open questions.
\newblock \emph{arXiv preprint arXiv:2209.03430}.

\bibitem[{Liang et~al.(2022{\natexlab{b}})Liang, Bommasani, Lee, Tsipras, Soylu, Yasunaga, Zhang, Narayanan, Wu, Kumar et~al.}]{liang2022holistic}
Percy Liang, Rishi Bommasani, Tony Lee, Dimitris Tsipras, Dilara Soylu, Michihiro Yasunaga, Yian Zhang, Deepak Narayanan, Yuhuai Wu, Ananya Kumar, et~al. 2022{\natexlab{b}}.
\newblock Holistic evaluation of language models.
\newblock \emph{arXiv preprint arXiv:2211.09110}.

\bibitem[{Liang et~al.(2022{\natexlab{c}})Liang, Zhang, Kwon, Yeung, and Zou}]{liang2022mind}
Victor~Weixin Liang, Yuhui Zhang, Yongchan Kwon, Serena Yeung, and James~Y Zou. 2022{\natexlab{c}}.
\newblock Mind the gap: Understanding the modality gap in multi-modal contrastive representation learning.
\newblock \emph{Advances in Neural Information Processing Systems}, 35:17612--17625.

\bibitem[{Lin et~al.(2022)Lin, Hilton, and Evans}]{lin-etal-2022-truthfulqa}
Stephanie Lin, Jacob Hilton, and Owain Evans. 2022.
\newblock \href {https://doi.org/10.18653/v1/2022.acl-long.229} {{T}ruthful{QA}: Measuring how models mimic human falsehoods}.
\newblock In \emph{Proceedings of the 60th Annual Meeting of the Association for Computational Linguistics (Volume 1: Long Papers)}, pages 3214--3252, Dublin, Ireland. Association for Computational Linguistics.

\bibitem[{Lin et~al.(2014)Lin, Maire, Belongie, Hays, Perona, Ramanan, Doll{\'a}r, and Zitnick}]{lin2014microsoft}
Tsung-Yi Lin, Michael Maire, Serge Belongie, James Hays, Pietro Perona, Deva Ramanan, Piotr Doll{\'a}r, and C~Lawrence Zitnick. 2014.
\newblock Microsoft coco: Common objects in context.
\newblock In \emph{Computer Vision--ECCV 2014: 13th European Conference, Zurich, Switzerland, September 6-12, 2014, Proceedings, Part V 13}, pages 740--755. Springer.

\bibitem[{Liu et~al.(2024{\natexlab{a}})Liu, Li, Li, and Lee}]{liu2024improved}
Haotian Liu, Chunyuan Li, Yuheng Li, and Yong~Jae Lee. 2024{\natexlab{a}}.
\newblock Improved baselines with visual instruction tuning.
\newblock In \emph{Proceedings of the IEEE/CVF Conference on Computer Vision and Pattern Recognition}, pages 26296--26306.

\bibitem[{Liu et~al.(2024{\natexlab{b}})Liu, Li, Wu, and Lee}]{liu2024visual}
Haotian Liu, Chunyuan Li, Qingyang Wu, and Yong~Jae Lee. 2024{\natexlab{b}}.
\newblock Visual instruction tuning.
\newblock \emph{Advances in neural information processing systems}, 36.

\bibitem[{Liu et~al.(2023{\natexlab{a}})Liu, Wang, Yang, and Zha}]{liu2023fingpt}
Xiao-Yang Liu, Guoxuan Wang, Hongyang Yang, and Daochen Zha. 2023{\natexlab{a}}.
\newblock Fingpt: Democratizing internet-scale data for financial large language models.
\newblock \emph{arXiv preprint arXiv:2307.10485}.

\bibitem[{Liu et~al.(2023{\natexlab{b}})Liu, Yao, Ton, Zhang, Guo, Cheng, Klochkov, Taufiq, and Li}]{liu2023trustworthy}
Yang Liu, Yuanshun Yao, Jean-Francois Ton, Xiaoying Zhang, Ruocheng Guo, Hao Cheng, Yegor Klochkov, Muhammad~Faaiz Taufiq, and Hang Li. 2023{\natexlab{b}}.
\newblock Trustworthy llms: A survey and guideline for evaluating large language models' alignment.
\newblock \emph{arXiv preprint arXiv:2308.05374}.

\bibitem[{Liu et~al.(2023{\natexlab{c}})Liu, Han, Ma, Zhang, Yang, Tian, He, Li, He, Liu et~al.}]{liu2023summary}
Yiheng Liu, Tianle Han, Siyuan Ma, Jiayue Zhang, Yuanyuan Yang, Jiaming Tian, Hao He, Antong Li, Mengshen He, Zhengliang Liu, et~al. 2023{\natexlab{c}}.
\newblock Summary of chatgpt-related research and perspective towards the future of large language models.
\newblock \emph{Meta-Radiology}, page 100017.

\bibitem[{Liu et~al.(2019)Liu, Ott, Goyal, Du, Joshi, Chen, Levy, Lewis, Zettlemoyer, and Stoyanov}]{liu2019roberta}
Yinhan Liu, Myle Ott, Naman Goyal, Jingfei Du, Mandar Joshi, Danqi Chen, Omer Levy, Mike Lewis, Luke Zettlemoyer, and Veselin Stoyanov. 2019.
\newblock Roberta: A robustly optimized bert pretraining approach.
\newblock \emph{arXiv preprint arXiv:1907.11692}.

\bibitem[{Lu et~al.(2020)Lu, Mardziel, Wu, Amancharla, and Datta}]{lu2020gender}
Kaiji Lu, Piotr Mardziel, Fangjing Wu, Preetam Amancharla, and Anupam Datta. 2020.
\newblock Gender bias in neural natural language processing.
\newblock \emph{Logic, language, and security: essays dedicated to Andre Scedrov on the occasion of his 65th birthday}, pages 189--202.

\bibitem[{Luccioni and Viviano(2021)}]{luccioni2021s}
Alexandra Luccioni and Joseph Viviano. 2021.
\newblock What’s in the box? an analysis of undesirable content in the common crawl corpus.
\newblock In \emph{Proceedings of the 59th Annual Meeting of the Association for Computational Linguistics and the 11th International Joint Conference on Natural Language Processing (Volume 2: Short Papers)}, pages 182--189.

\bibitem[{Lui et~al.(2024)Lui, Chia, Berrios, Ross, and Kiela}]{lui2024leveraging}
Nicholas Lui, Bryan Chia, William Berrios, Candace Ross, and Douwe Kiela. 2024.
\newblock Leveraging diffusion perturbations for measuring fairness in computer vision.
\newblock In \emph{Proceedings of the AAAI Conference on Artificial Intelligence}, volume~38, pages 14220--14228.

\bibitem[{Luo et~al.(2024)Luo, Huang, Deng, Liu, Chen, and Liu}]{luo2024bigbench}
Hanjun Luo, Haoyu Huang, Ziye Deng, Xuecheng Liu, Ruizhe Chen, and Zuozhu Liu. 2024.
\newblock Bigbench: A unified benchmark for social bias in text-to-image generative models based on multi-modal llm.
\newblock \emph{arXiv preprint arXiv:2407.15240}.

\bibitem[{Ma et~al.(2015)Ma, Correll, and Wittenbrink}]{ma2015chicago}
Debbie~S Ma, Joshua Correll, and Bernd Wittenbrink. 2015.
\newblock The chicago face database: A free stimulus set of faces and norming data.
\newblock \emph{Behavior research methods}, 47:1122--1135.

\bibitem[{Ma et~al.(2023)Ma, Zhang, Bian, Liu, Zhang, Zhao, Zhang, Fu, Hu, and Wu}]{ma2023fairness}
Huan Ma, Changqing Zhang, Yatao Bian, Lemao Liu, Zhirui Zhang, Peilin Zhao, Shu Zhang, Huazhu Fu, Qinghua Hu, and Bingzhe Wu. 2023.
\newblock Fairness-guided few-shot prompting for large language models.
\newblock \emph{Advances in Neural Information Processing Systems}, 36:43136--43155.

\bibitem[{Malic et~al.(2023)Malic, Kumari, and Liu}]{10411546}
Vincent~Quirante Malic, Anamika Kumari, and Xiaozhong Liu. 2023.
\newblock \href {https://doi.org/10.1109/ICDMW60847.2023.00037} {Racial skew in fine-tuned legal ai language models}.
\newblock In \emph{2023 IEEE International Conference on Data Mining Workshops (ICDMW)}, pages 245--252.

\bibitem[{Mandal et~al.(2023{\natexlab{a}})Mandal, Leavy, and Little}]{mandal2023measuring}
Abhishek Mandal, Susan Leavy, and Suzanne Little. 2023{\natexlab{a}}.
\newblock Measuring bias in multimodal models: Multimodal composite association score.
\newblock In \emph{International Workshop on Algorithmic Bias in Search and Recommendation}, pages 17--30. Springer.

\bibitem[{Mandal et~al.(2023{\natexlab{b}})Mandal, Little, and Leavy}]{10.1145/3577190.3614156}
Abhishek Mandal, Suzanne Little, and Susan Leavy. 2023{\natexlab{b}}.
\newblock \href {https://doi.org/10.1145/3577190.3614156} {Multimodal bias: Assessing gender bias in computer vision models with nlp techniques}.
\newblock In \emph{Proceedings of the 25th International Conference on Multimodal Interaction}, ICMI '23, page 416–424, New York, NY, USA. Association for Computing Machinery.

\bibitem[{May et~al.(2019)May, Wang, Bordia, Bowman, and Rudinger}]{may-etal-2019-measuring}
Chandler May, Alex Wang, Shikha Bordia, Samuel~R. Bowman, and Rachel Rudinger. 2019.
\newblock \href {https://doi.org/10.18653/v1/N19-1063} {On measuring social biases in sentence encoders}.
\newblock In \emph{Proceedings of the 2019 Conference of the North {A}merican Chapter of the Association for Computational Linguistics: Human Language Technologies, Volume 1 (Long and Short Papers)}, pages 622--628, Minneapolis, Minnesota. Association for Computational Linguistics.

\bibitem[{Meade et~al.(2022)Meade, Poole-Dayan, and Reddy}]{meade-etal-2022-empirical}
Nicholas Meade, Elinor Poole-Dayan, and Siva Reddy. 2022.
\newblock \href {https://doi.org/10.18653/v1/2022.acl-long.132} {An empirical survey of the effectiveness of debiasing techniques for pre-trained language models}.
\newblock In \emph{Proceedings of the 60th Annual Meeting of the Association for Computational Linguistics (Volume 1: Long Papers)}, pages 1878--1898, Dublin, Ireland. Association for Computational Linguistics.

\bibitem[{Mehrabi et~al.(2021)Mehrabi, Morstatter, Saxena, Lerman, and Galstyan}]{mehrabi2021survey}
Ninareh Mehrabi, Fred Morstatter, Nripsuta Saxena, Kristina Lerman, and Aram Galstyan. 2021.
\newblock A survey on bias and fairness in machine learning.
\newblock \emph{ACM computing surveys (CSUR)}, 54(6):1--35.

\bibitem[{Mei et~al.(2023)Mei, Fereidooni, and Caliskan}]{10.1145/3593013.3594109}
Katelyn Mei, Sonia Fereidooni, and Aylin Caliskan. 2023.
\newblock \href {https://doi.org/10.1145/3593013.3594109} {Bias against 93 stigmatized groups in masked language models and downstream sentiment classification tasks}.
\newblock In \emph{Proceedings of the 2023 ACM Conference on Fairness, Accountability, and Transparency}, FAccT '23, page 1699–1710, New York, NY, USA. Association for Computing Machinery.

\bibitem[{Meng et~al.(2022{\natexlab{a}})Meng, Trinh, Xu, Enouen, and Liu}]{meng2022interpretability}
Chuizheng Meng, Loc Trinh, Nan Xu, James Enouen, and Yan Liu. 2022{\natexlab{a}}.
\newblock Interpretability and fairness evaluation of deep learning models on mimic-iv dataset.
\newblock \emph{Scientific Reports}, 12(1):7166.

\bibitem[{Meng et~al.(2022{\natexlab{b}})Meng, Huang, Zhang, and Han}]{meng2022generating}
Yu~Meng, Jiaxin Huang, Yu~Zhang, and Jiawei Han. 2022{\natexlab{b}}.
\newblock Generating training data with language models: Towards zero-shot language understanding.
\newblock \emph{Advances in Neural Information Processing Systems}, 35:462--477.

\bibitem[{Menick et~al.(2022)Menick, Trebacz, Mikulik, Aslanides, Song, Chadwick, Glaese, Young, Campbell-Gillingham, Irving et~al.}]{menick2022teaching}
Jacob Menick, Maja Trebacz, Vladimir Mikulik, John Aslanides, Francis Song, Martin Chadwick, Mia Glaese, Susannah Young, Lucy Campbell-Gillingham, Geoffrey Irving, et~al. 2022.
\newblock Teaching language models to support answers with verified quotes.
\newblock \emph{arXiv preprint arXiv:2203.11147}.

\bibitem[{Mesk{\'o} and Topol(2023)}]{mesko2023imperative}
Bertalan Mesk{\'o} and Eric~J Topol. 2023.
\newblock The imperative for regulatory oversight of large language models (or generative ai) in healthcare.
\newblock \emph{NPJ digital medicine}, 6(1):120.

\bibitem[{Meyer et~al.(2023)Meyer, Urbanowicz, Martin, O’Connor, Li, Peng, Bright, Tatonetti, Won, Gonzalez-Hernandez et~al.}]{meyer2023chatgpt}
Jesse~G Meyer, Ryan~J Urbanowicz, Patrick~CN Martin, Karen O’Connor, Ruowang Li, Pei-Chen Peng, Tiffani~J Bright, Nicholas Tatonetti, Kyoung~Jae Won, Graciela Gonzalez-Hernandez, et~al. 2023.
\newblock Chatgpt and large language models in academia: opportunities and challenges.
\newblock \emph{BioData Mining}, 16(1):20.

\bibitem[{Mikolov et~al.(2013)Mikolov, Sutskever, Chen, Corrado, and Dean}]{mikolov2013distributed}
Tomas Mikolov, Ilya Sutskever, Kai Chen, Greg~S Corrado, and Jeff Dean. 2013.
\newblock Distributed representations of words and phrases and their compositionality.
\newblock \emph{Advances in neural information processing systems}, 26.

\bibitem[{Min et~al.(2023)Min, Ross, Sulem, Veyseh, Nguyen, Sainz, Agirre, Heintz, and Roth}]{min2023recent}
Bonan Min, Hayley Ross, Elior Sulem, Amir Pouran~Ben Veyseh, Thien~Huu Nguyen, Oscar Sainz, Eneko Agirre, Ilana Heintz, and Dan Roth. 2023.
\newblock Recent advances in natural language processing via large pre-trained language models: A survey.
\newblock \emph{ACM Computing Surveys}, 56(2):1--40.

\bibitem[{Morales et~al.(2023)Morales, Clarisó, and Cabot}]{10298519}
Sergio Morales, Robert Clarisó, and Jordi Cabot. 2023.
\newblock \href {https://doi.org/10.1109/ASE56229.2023.00018} {Automating bias testing of llms}.
\newblock In \emph{2023 38th IEEE/ACM International Conference on Automated Software Engineering (ASE)}, pages 1705--1707.

\bibitem[{Mozafari et~al.(2020)Mozafari, Farahbakhsh, and Crespi}]{mozafari2020hate}
Marzieh Mozafari, Reza Farahbakhsh, and No{\"e}l Crespi. 2020.
\newblock Hate speech detection and racial bias mitigation in social media based on bert model.
\newblock \emph{PloS one}, 15(8):e0237861.

\bibitem[{Mu et~al.(2022)Mu, Kirillov, Wagner, and Xie}]{norman2022slip}
Norman Mu, Alexander Kirillov, David Wagner, and Saining Xie. 2022.
\newblock \href {https://doi.org/10.1007/978-3-031-19809-0_30} {Slip: Self-supervision meets language-image pre-training}.
\newblock In \emph{Computer Vision – ECCV 2022: 17th European Conference, Tel Aviv, Israel, October 23–27, 2022, Proceedings, Part XXVI}, page 529–544, Berlin, Heidelberg. Springer-Verlag.

\bibitem[{Myers et~al.(2024)Myers, Mohawesh, Chellaboina, Sathvik, Venkatesh, Ho, Henshaw, Alhawawreh, Berdik, and Jararweh}]{myers2024foundation}
Devon Myers, Rami Mohawesh, Venkata~Ishwarya Chellaboina, Anantha~Lakshmi Sathvik, Praveen Venkatesh, Yi-Hui Ho, Hanna Henshaw, Muna Alhawawreh, David Berdik, and Yaser Jararweh. 2024.
\newblock Foundation and large language models: fundamentals, challenges, opportunities, and social impacts.
\newblock \emph{Cluster Computing}, 27(1):1--26.

\bibitem[{Nadeem et~al.(2021)Nadeem, Bethke, and Reddy}]{nadeem-etal-2021-stereoset}
Moin Nadeem, Anna Bethke, and Siva Reddy. 2021.
\newblock \href {https://doi.org/10.18653/v1/2021.acl-long.416} {{S}tereo{S}et: Measuring stereotypical bias in pretrained language models}.
\newblock In \emph{Proceedings of the 59th Annual Meeting of the Association for Computational Linguistics and the 11th International Joint Conference on Natural Language Processing (Volume 1: Long Papers)}, pages 5356--5371, Online. Association for Computational Linguistics.

\bibitem[{Nagrani et~al.(2020)Nagrani, Chung, Xie, and Zisserman}]{nagrani2020voxceleb}
Arsha Nagrani, Joon~Son Chung, Weidi Xie, and Andrew Zisserman. 2020.
\newblock Voxceleb: Large-scale speaker verification in the wild.
\newblock \emph{Computer Speech \& Language}, 60:101027.

\bibitem[{Nakamura et~al.(2024)Nakamura, Mishra, Tedeschi, Chai, Stillerman, Friedrich, Yadav, Laud, Chien, Zhuo et~al.}]{nakamura2024aurora}
Taishi Nakamura, Mayank Mishra, Simone Tedeschi, Yekun Chai, Jason~T Stillerman, Felix Friedrich, Prateek Yadav, Tanmay Laud, Vu~Minh Chien, Terry~Yue Zhuo, et~al. 2024.
\newblock Aurora-m: The first open source multilingual language model red-teamed according to the us executive order.
\newblock \emph{arXiv preprint arXiv:2404.00399}.

\bibitem[{Nashwan and Abujaber(2023)}]{nashwan2023harnessing}
Abdulqadir~J Nashwan and Ahmad~A Abujaber. 2023.
\newblock Harnessing large language models in nursing care planning: opportunities, challenges, and ethical considerations.
\newblock \emph{Cureus}, 15(6).

\bibitem[{Navigli et~al.(2023)Navigli, Conia, and Ross}]{navigli2023biases}
Roberto Navigli, Simone Conia, and Bj{\"o}rn Ross. 2023.
\newblock Biases in large language models: origins, inventory, and discussion.
\newblock \emph{ACM Journal of Data and Information Quality}, 15(2):1--21.

\bibitem[{Nazi and Peng(2024)}]{informatics11030057}
Zabir~Al Nazi and Wei Peng. 2024.
\newblock \href {https://doi.org/10.3390/informatics11030057} {Large language models in healthcare and medical domain: A review}.
\newblock \emph{Informatics}, 11(3).

\bibitem[{Nazir et~al.(2024)Nazir, Chakravarthy, Cecchini, Khajuria, Sharma, Mirik, Kocaman, and Talby}]{NAZIR2024100619}
Arshaan Nazir, Thadaka~Kalyan Chakravarthy, David~Amore Cecchini, Rakshit Khajuria, Prikshit Sharma, Ali~Tarik Mirik, Veysel Kocaman, and David Talby. 2024.
\newblock \href {https://doi.org/https://doi.org/10.1016/j.simpa.2024.100619} {Langtest: A comprehensive evaluation library for custom llm and nlp models}.
\newblock \emph{Software Impacts}, 19:100619.

\bibitem[{Nichol et~al.(2022)Nichol, Dhariwal, Ramesh, Shyam, Mishkin, Mcgrew, Sutskever, and Chen}]{pmlr-v162-nichol22a}
Alexander~Quinn Nichol, Prafulla Dhariwal, Aditya Ramesh, Pranav Shyam, Pamela Mishkin, Bob Mcgrew, Ilya Sutskever, and Mark Chen. 2022.
\newblock \href {https://proceedings.mlr.press/v162/nichol22a.html} {{GLIDE}: Towards photorealistic image generation and editing with text-guided diffusion models}.
\newblock In \emph{Proceedings of the 39th International Conference on Machine Learning}, volume 162 of \emph{Proceedings of Machine Learning Research}, pages 16784--16804. PMLR.

\bibitem[{Niu et~al.(2021{\natexlab{a}})Niu, Tang, Zhang, Lu, Hua, and Wen}]{niu2021counterfactual}
Yulei Niu, Kaihua Tang, Hanwang Zhang, Zhiwu Lu, Xian-Sheng Hua, and Ji-Rong Wen. 2021{\natexlab{a}}.
\newblock Counterfactual vqa: A cause-effect look at language bias.
\newblock In \emph{Proceedings of the IEEE/CVF conference on computer vision and pattern recognition}, pages 12700--12710.

\bibitem[{Niu et~al.(2021{\natexlab{b}})Niu, Tang, Zhang, Lu, Hua, and Wen}]{Niu_2021_CVPR}
Yulei Niu, Kaihua Tang, Hanwang Zhang, Zhiwu Lu, Xian-Sheng Hua, and Ji-Rong Wen. 2021{\natexlab{b}}.
\newblock Counterfactual vqa: A cause-effect look at language bias.
\newblock In \emph{Proceedings of the IEEE/CVF Conference on Computer Vision and Pattern Recognition (CVPR)}, pages 12700--12710.

\bibitem[{Nosek et~al.(2002)Nosek, Banaji, and Greenwald}]{nosek2002harvesting}
Brian~A Nosek, Mahzarin~R Banaji, and Anthony~G Greenwald. 2002.
\newblock Harvesting implicit group attitudes and beliefs from a demonstration web site.
\newblock \emph{Group Dynamics: Theory, research, and practice}, 6(1):101.

\bibitem[{Nozza et~al.(2021)Nozza, Bianchi, and Hovy}]{nozza2021honest}
Debora Nozza, Federico Bianchi, and Dirk Hovy. 2021.
\newblock Honest: Measuring hurtful sentence completion in language models.
\newblock In \emph{The 2021 Conference of the North American Chapter of the Association for Computational Linguistics: Human Language Technologies}. Association for Computational Linguistics.

\bibitem[{Ntoutsi et~al.(2020)Ntoutsi, Fafalios, Gadiraju, Iosifidis, Nejdl, Vidal, Ruggieri, Turini, Papadopoulos, Krasanakis et~al.}]{ntoutsi2020bias}
Eirini Ntoutsi, Pavlos Fafalios, Ujwal Gadiraju, Vasileios Iosifidis, Wolfgang Nejdl, Maria-Esther Vidal, Salvatore Ruggieri, Franco Turini, Symeon Papadopoulos, Emmanouil Krasanakis, et~al. 2020.
\newblock Bias in data-driven artificial intelligence systems—an introductory survey.
\newblock \emph{Wiley Interdisciplinary Reviews: Data Mining and Knowledge Discovery}, 10(3):e1356.

\bibitem[{Ouyang et~al.(2022)Ouyang, Wu, Jiang, Almeida, Wainwright, Mishkin, Zhang, Agarwal, Slama, Ray et~al.}]{ouyang2022training}
Long Ouyang, Jeffrey Wu, Xu~Jiang, Diogo Almeida, Carroll Wainwright, Pamela Mishkin, Chong Zhang, Sandhini Agarwal, Katarina Slama, Alex Ray, et~al. 2022.
\newblock Training language models to follow instructions with human feedback.
\newblock \emph{Advances in neural information processing systems}, 35:27730--27744.

\bibitem[{Ovalle et~al.(2023)Ovalle, Goyal, Dhamala, Jaggers, Chang, Galstyan, Zemel, and Gupta}]{10.1145/3593013.3594078}
Anaelia Ovalle, Palash Goyal, Jwala Dhamala, Zachary Jaggers, Kai-Wei Chang, Aram Galstyan, Richard Zemel, and Rahul Gupta. 2023.
\newblock \href {https://doi.org/10.1145/3593013.3594078} {“i’m fully who i am”: Towards centering transgender and non-binary voices to measure biases in open language generation}.
\newblock FAccT '23, page 1246–1266, New York, NY, USA. Association for Computing Machinery.

\bibitem[{Pagano et~al.(2023)Pagano, Loureiro, Lisboa, Peixoto, Guimar{\~a}es, Cruz, Araujo, Santos, Cruz, Oliveira et~al.}]{pagano2023bias}
Tiago~P Pagano, Rafael~B Loureiro, Fernanda~VN Lisboa, Rodrigo~M Peixoto, Guilherme~AS Guimar{\~a}es, Gustavo~OR Cruz, Maira~M Araujo, Lucas~L Santos, Marco~AS Cruz, Ewerton~LS Oliveira, et~al. 2023.
\newblock Bias and unfairness in machine learning models: a systematic review on datasets, tools, fairness metrics, and identification and mitigation methods.
\newblock \emph{Big data and cognitive computing}, 7(1):15.

\bibitem[{Pagliai et~al.(2024)Pagliai, van Boven, Adewumi, Alkhaled, Gurung, S{\"o}dergren, and Barney}]{pagliai2024data}
Irene Pagliai, Goya van Boven, Tosin Adewumi, Lama Alkhaled, Namrata Gurung, Isabella S{\"o}dergren, and Elisa Barney. 2024.
\newblock Data bias according to bipol: Men are naturally right and it is the role of women to follow their lead.
\newblock \emph{arXiv preprint arXiv:2404.04838}.

\bibitem[{Park et~al.(2018)Park, Shin, and Fung}]{park-etal-2018-reducing}
Ji~Ho Park, Jamin Shin, and Pascale Fung. 2018.
\newblock \href {https://doi.org/10.18653/v1/D18-1302} {Reducing gender bias in abusive language detection}.
\newblock In \emph{Proceedings of the 2018 Conference on Empirical Methods in Natural Language Processing}, pages 2799--2804, Brussels, Belgium. Association for Computational Linguistics.

\bibitem[{Parrish et~al.(2022)Parrish, Chen, Nangia, Padmakumar, Phang, Thompson, Htut, and Bowman}]{parrish-etal-2022-bbq}
Alicia Parrish, Angelica Chen, Nikita Nangia, Vishakh Padmakumar, Jason Phang, Jana Thompson, Phu~Mon Htut, and Samuel Bowman. 2022.
\newblock \href {https://doi.org/10.18653/v1/2022.findings-acl.165} {{BBQ}: A hand-built bias benchmark for question answering}.
\newblock In \emph{Findings of the Association for Computational Linguistics: ACL 2022}, pages 2086--2105, Dublin, Ireland. Association for Computational Linguistics.

\bibitem[{Pe{\~n}a et~al.(2023)Pe{\~n}a, Serna, Morales, Fierrez, Ortega, Herrarte, Alcantara, and Ortega-Garcia}]{pena2023human}
Alejandro Pe{\~n}a, Ignacio Serna, Aythami Morales, Julian Fierrez, Alfonso Ortega, Ainhoa Herrarte, Manuel Alcantara, and Javier Ortega-Garcia. 2023.
\newblock Human-centric multimodal machine learning: Recent advances and testbed on ai-based recruitment.
\newblock \emph{SN Computer Science}, 4(5):434.

\bibitem[{Peng et~al.(2022)Peng, Wei, Deng, Wang, and Hu}]{peng2022balanced}
Xiaokang Peng, Yake Wei, Andong Deng, Dong Wang, and Di~Hu. 2022.
\newblock Balanced multimodal learning via on-the-fly gradient modulation.
\newblock In \emph{Proceedings of the IEEE/CVF conference on computer vision and pattern recognition}, pages 8238--8247.

\bibitem[{Pennington et~al.(2014)Pennington, Socher, and Manning}]{pennington2014glove}
Jeffrey Pennington, Richard Socher, and Christopher~D Manning. 2014.
\newblock Glove: Global vectors for word representation.
\newblock In \emph{Proceedings of the 2014 conference on empirical methods in natural language processing (EMNLP)}, pages 1532--1543.

\bibitem[{Perez et~al.(2022)Perez, Huang, Song, Cai, Ring, Aslanides, Glaese, McAleese, and Irving}]{perez-etal-2022-red}
Ethan Perez, Saffron Huang, Francis Song, Trevor Cai, Roman Ring, John Aslanides, Amelia Glaese, Nat McAleese, and Geoffrey Irving. 2022.
\newblock \href {https://doi.org/10.18653/v1/2022.emnlp-main.225} {Red teaming language models with language models}.
\newblock In \emph{Proceedings of the 2022 Conference on Empirical Methods in Natural Language Processing}, pages 3419--3448, Abu Dhabi, United Arab Emirates. Association for Computational Linguistics.

\bibitem[{Pessach and Shmueli(2022)}]{10.1145/3494672}
Dana Pessach and Erez Shmueli. 2022.
\newblock \href {https://doi.org/10.1145/3494672} {A review on fairness in machine learning}.
\newblock \emph{ACM Comput. Surv.}, 55(3).

\bibitem[{Petryk et~al.(2022)Petryk, Dunlap, Nasseri, Gonzalez, Darrell, and Rohrbach}]{9880269}
Suzanne Petryk, Lisa Dunlap, Keyan Nasseri, Joseph Gonzalez, Trevor Darrell, and Anna Rohrbach. 2022.
\newblock \href {https://doi.org/10.1109/CVPR52688.2022.01756} {On guiding visual attention with language specification}.
\newblock In \emph{2022 IEEE/CVF Conference on Computer Vision and Pattern Recognition (CVPR)}, pages 18071--18081.

\bibitem[{Pettersson et~al.(2024)Pettersson, Hult, Eriksson, and Adewumi}]{pettersson2024generative}
Jenny Pettersson, Elias Hult, Tim Eriksson, and Tosin Adewumi. 2024.
\newblock Generative ai and teachers--for us or against us? a case study.
\newblock \emph{arXiv preprint arXiv:2404.03486}.

\bibitem[{Pfeiffer et~al.(2021)Pfeiffer, Kamath, R{\"u}ckl{\'e}, Cho, and Gurevych}]{pfeiffer-etal-2021-adapterfusion}
Jonas Pfeiffer, Aishwarya Kamath, Andreas R{\"u}ckl{\'e}, Kyunghyun Cho, and Iryna Gurevych. 2021.
\newblock \href {https://doi.org/10.18653/v1/2021.eacl-main.39} {{A}dapter{F}usion: Non-destructive task composition for transfer learning}.
\newblock In \emph{Proceedings of the 16th Conference of the European Chapter of the Association for Computational Linguistics: Main Volume}, pages 487--503, Online. Association for Computational Linguistics.

\bibitem[{Pi{\~n}eiro-Mart{\'\i}n et~al.(2023)Pi{\~n}eiro-Mart{\'\i}n, Garc{\'\i}a-Mateo, Doc{\'\i}o-Fern{\'a}ndez, and Lopez-Perez}]{pineiro2023ethical}
Andr{\'e}s Pi{\~n}eiro-Mart{\'\i}n, Carmen Garc{\'\i}a-Mateo, Laura Doc{\'\i}o-Fern{\'a}ndez, and Maria Del~Carmen Lopez-Perez. 2023.
\newblock Ethical challenges in the development of virtual assistants powered by large language models.
\newblock \emph{Electronics}, 12(14):3170.

\bibitem[{Porgali et~al.(2023)Porgali, Albiero, Ryda, Ferrer, and Hazirbas}]{10208759}
Bilal Porgali, Vítor Albiero, Jordan Ryda, Cristian~Canton Ferrer, and Caner Hazirbas. 2023.
\newblock \href {https://doi.org/10.1109/CVPRW59228.2023.00006} {The casual conversations v2 dataset : A diverse, large benchmark for measuring fairness and robustness in audio/vision/speech models}.
\newblock In \emph{2023 IEEE/CVF Conference on Computer Vision and Pattern Recognition Workshops (CVPRW)}, pages 10--17.

\bibitem[{Pryzant et~al.(2020)Pryzant, Martinez, Dass, Kurohashi, Jurafsky, and Yang}]{pryzant2020automatically}
Reid Pryzant, Richard~Diehl Martinez, Nathan Dass, Sadao Kurohashi, Dan Jurafsky, and Diyi Yang. 2020.
\newblock Automatically neutralizing subjective bias in text.
\newblock In \emph{Proceedings of the aaai conference on artificial intelligence}, volume~34, pages 480--489.

\bibitem[{Qi et~al.(2023)Qi, Zeng, Xie, Chen, Jia, Mittal, and Henderson}]{qi2023fine}
Xiangyu Qi, Yi~Zeng, Tinghao Xie, Pin-Yu Chen, Ruoxi Jia, Prateek Mittal, and Peter Henderson. 2023.
\newblock Fine-tuning aligned language models compromises safety, even when users do not intend to!
\newblock \emph{arXiv preprint arXiv:2310.03693}.

\bibitem[{Radford et~al.(2021)Radford, Kim, Hallacy, Ramesh, Goh, Agarwal, Sastry, Askell, Mishkin, Clark et~al.}]{radford2021learning}
Alec Radford, Jong~Wook Kim, Chris Hallacy, Aditya Ramesh, Gabriel Goh, Sandhini Agarwal, Girish Sastry, Amanda Askell, Pamela Mishkin, Jack Clark, et~al. 2021.
\newblock Learning transferable visual models from natural language supervision.
\newblock In \emph{International conference on machine learning}, pages 8748--8763. PMLR.

\bibitem[{Radford et~al.(2019)Radford, Wu, Child, Luan, Amodei, Sutskever et~al.}]{radford2019language}
Alec Radford, Jeffrey Wu, Rewon Child, David Luan, Dario Amodei, Ilya Sutskever, et~al. 2019.
\newblock Language models are unsupervised multitask learners.
\newblock \emph{OpenAI blog}, 1(8):9.

\bibitem[{Rae et~al.(2021)Rae, Borgeaud, Cai, Millican, Hoffmann, Song, Aslanides, Henderson, Ring, Young et~al.}]{rae2021scaling}
Jack~W Rae, Sebastian Borgeaud, Trevor Cai, Katie Millican, Jordan Hoffmann, Francis Song, John Aslanides, Sarah Henderson, Roman Ring, Susannah Young, et~al. 2021.
\newblock Scaling language models: Methods, analysis \& insights from training gopher.
\newblock \emph{arXiv preprint arXiv:2112.11446}.

\bibitem[{Raffel et~al.(2020)Raffel, Shazeer, Roberts, Lee, Narang, Matena, Zhou, Li, and Liu}]{raffel2020exploring}
Colin Raffel, Noam Shazeer, Adam Roberts, Katherine Lee, Sharan Narang, Michael Matena, Yanqi Zhou, Wei Li, and Peter~J Liu. 2020.
\newblock Exploring the limits of transfer learning with a unified text-to-text transformer.
\newblock \emph{Journal of machine learning research}, 21(140):1--67.

\bibitem[{Rahate et~al.(2022)Rahate, Walambe, Ramanna, and Kotecha}]{rahate2022multimodal}
Anil Rahate, Rahee Walambe, Sheela Ramanna, and Ketan Kotecha. 2022.
\newblock Multimodal co-learning: Challenges, applications with datasets, recent advances and future directions.
\newblock \emph{Information Fusion}, 81:203--239.

\bibitem[{Rahman et~al.(2020)Rahman, Hasan, Lee, Zadeh, Mao, Morency, and Hoque}]{rahman2020integrating}
Wasifur Rahman, Md~Kamrul Hasan, Sangwu Lee, Amir Zadeh, Chengfeng Mao, Louis-Philippe Morency, and Ehsan Hoque. 2020.
\newblock Integrating multimodal information in large pretrained transformers.
\newblock In \emph{Proceedings of the conference. Association for Computational Linguistics. Meeting}, volume 2020, page 2359. NIH Public Access.

\bibitem[{Raj et~al.(2023)Raj, Mukherjee, and Zhu}]{raj2023true}
Chahat Raj, Anjishnu Mukherjee, and Ziwei Zhu. 2023.
\newblock True and fair: Robust and unbiased fake news detection via interpretable machine learning.
\newblock In \emph{Proceedings of the 2023 AAAI/ACM Conference on AI, Ethics, and Society}, pages 962--963.

\bibitem[{Ramesh et~al.(2022)Ramesh, Dhariwal, Nichol, Chu, and Chen}]{ramesh2022hierarchical}
Aditya Ramesh, Prafulla Dhariwal, Alex Nichol, Casey Chu, and Mark Chen. 2022.
\newblock Hierarchical text-conditional image generation with clip latents.
\newblock \emph{arXiv preprint arXiv:2204.06125}, 1(2):3.

\bibitem[{Ramesh et~al.(2021)Ramesh, Pavlov, Goh, Gray, Voss, Radford, Chen, and Sutskever}]{ramesh2021zero}
Aditya Ramesh, Mikhail Pavlov, Gabriel Goh, Scott Gray, Chelsea Voss, Alec Radford, Mark Chen, and Ilya Sutskever. 2021.
\newblock Zero-shot text-to-image generation.
\newblock In \emph{International conference on machine learning}, pages 8821--8831. Pmlr.

\bibitem[{Ramesh et~al.(2023)Ramesh, Sitaram, and Choudhury}]{ramesh-etal-2023-fairness}
Krithika Ramesh, Sunayana Sitaram, and Monojit Choudhury. 2023.
\newblock \href {https://doi.org/10.18653/v1/2023.findings-eacl.157} {Fairness in language models beyond {E}nglish: Gaps and challenges}.
\newblock In \emph{Findings of the Association for Computational Linguistics: EACL 2023}, pages 2106--2119, Dubrovnik, Croatia. Association for Computational Linguistics.

\bibitem[{Rana and Jha(2022)}]{rana2022emotion}
Aneri Rana and Sonali Jha. 2022.
\newblock Emotion based hate speech detection using multimodal learning.
\newblock \emph{arXiv preprint arXiv:2202.06218}.

\bibitem[{Ray(2023)}]{ray2023chatgpt}
Partha~Pratim Ray. 2023.
\newblock Chatgpt: A comprehensive review on background, applications, key challenges, bias, ethics, limitations and future scope.
\newblock \emph{Internet of Things and Cyber-Physical Systems}, 3:121--154.

\bibitem[{Raza et~al.(2024)Raza, Ghuge, Ding, Dolatabadi, and Pandya}]{raza2024fair}
Shaina Raza, Shardul Ghuge, Chen Ding, Elham Dolatabadi, and Deval Pandya. 2024.
\newblock Fair enough: Develop and assess a fair-compliant dataset for large language model training?
\newblock \emph{Data Intelligence}, 6(2):559--585.

\bibitem[{Rombach et~al.(2022)Rombach, Blattmann, Lorenz, Esser, and Ommer}]{rombach2022high}
Robin Rombach, Andreas Blattmann, Dominik Lorenz, Patrick Esser, and Bj{\"o}rn Ommer. 2022.
\newblock High-resolution image synthesis with latent diffusion models.
\newblock In \emph{Proceedings of the IEEE/CVF conference on computer vision and pattern recognition}, pages 10684--10695.

\bibitem[{R{\"o}{\"o}sli et~al.(2022)R{\"o}{\"o}sli, Bozkurt, and Hernandez-Boussard}]{roosli2022peeking}
Eliane R{\"o}{\"o}sli, Selen Bozkurt, and Tina Hernandez-Boussard. 2022.
\newblock Peeking into a black box, the fairness and generalizability of a mimic-iii benchmarking model.
\newblock \emph{Scientific Data}, 9(1):24.

\bibitem[{Ross et~al.(2021)Ross, Katz, and Barbu}]{ross-etal-2021-measuring}
Candace Ross, Boris Katz, and Andrei Barbu. 2021.
\newblock \href {https://doi.org/10.18653/v1/2021.naacl-main.78} {Measuring social biases in grounded vision and language embeddings}.
\newblock In \emph{Proceedings of the 2021 Conference of the North American Chapter of the Association for Computational Linguistics: Human Language Technologies}, pages 998--1008, Online. Association for Computational Linguistics.

\bibitem[{Ruder et~al.(2022)Ruder, Vuli{\'c}, and S{\o}gaard}]{ruder-etal-2022-square}
Sebastian Ruder, Ivan Vuli{\'c}, and Anders S{\o}gaard. 2022.
\newblock \href {https://doi.org/10.18653/v1/2022.findings-acl.184} {Square one bias in {NLP}: Towards a multi-dimensional exploration of the research manifold}.
\newblock In \emph{Findings of the Association for Computational Linguistics: ACL 2022}, pages 2340--2354, Dublin, Ireland. Association for Computational Linguistics.

\bibitem[{Rudinger et~al.(2018)Rudinger, Naradowsky, Leonard, and Van~Durme}]{rudinger-etal-2018-gender}
Rachel Rudinger, Jason Naradowsky, Brian Leonard, and Benjamin Van~Durme. 2018.
\newblock \href {https://doi.org/10.18653/v1/N18-2002} {Gender bias in coreference resolution}.
\newblock In \emph{Proceedings of the 2018 Conference of the North {A}merican Chapter of the Association for Computational Linguistics: Human Language Technologies, Volume 2 (Short Papers)}, pages 8--14, New Orleans, Louisiana. Association for Computational Linguistics.

\bibitem[{Ruggeri and Nozza(2023)}]{ruggeri-nozza-2023-multi}
Gabriele Ruggeri and Debora Nozza. 2023.
\newblock \href {https://doi.org/10.18653/v1/2023.findings-acl.403} {A multi-dimensional study on bias in vision-language models}.
\newblock In \emph{Findings of the Association for Computational Linguistics: ACL 2023}, pages 6445--6455, Toronto, Canada. Association for Computational Linguistics.

\bibitem[{Ruzzante et~al.(2022)Ruzzante, Monachesi, Orabona, and Vaes}]{ruzzante2022sexual}
Daniela Ruzzante, Bianca Monachesi, Noemi Orabona, and Jeroen Vaes. 2022.
\newblock The sexual objectification and emotion database: A free stimulus set and norming data of sexually objectified and non-objectified female targets expressing multiple emotions.
\newblock \emph{Behavior Research Methods}, pages 1--15.

\bibitem[{Salinas et~al.(2023)Salinas, Shah, Huang, McCormack, and Morstatter}]{10.1145/3617694.3623257}
Abel Salinas, Parth Shah, Yuzhong Huang, Robert McCormack, and Fred Morstatter. 2023.
\newblock \href {https://doi.org/10.1145/3617694.3623257} {The unequal opportunities of large language models: Examining demographic biases in job recommendations by chatgpt and llama}.
\newblock In \emph{Proceedings of the 3rd ACM Conference on Equity and Access in Algorithms, Mechanisms, and Optimization}, EAAMO '23, New York, NY, USA. Association for Computing Machinery.

\bibitem[{Sami et~al.(2023)Sami, Sami, and Barclay}]{10336315}
Mansour Sami, Ashkan Sami, and Pete Barclay. 2023.
\newblock \href {https://doi.org/10.1109/ICSME58846.2023.00051} {A case study of fairness in generated images of large language models for software engineering tasks}.
\newblock In \emph{2023 IEEE International Conference on Software Maintenance and Evolution (ICSME)}, pages 391--396.

\bibitem[{Santurkar et~al.(2023)Santurkar, Durmus, Ladhak, Lee, Liang, and Hashimoto}]{santurkar2023whose}
Shibani Santurkar, Esin Durmus, Faisal Ladhak, Cinoo Lee, Percy Liang, and Tatsunori Hashimoto. 2023.
\newblock Whose opinions do language models reflect?
\newblock In \emph{International Conference on Machine Learning}, pages 29971--30004. PMLR.

\bibitem[{Sap et~al.(2020)Sap, Gabriel, Qin, Jurafsky, Smith, and Choi}]{sap-etal-2020-social}
Maarten Sap, Saadia Gabriel, Lianhui Qin, Dan Jurafsky, Noah~A. Smith, and Yejin Choi. 2020.
\newblock \href {https://doi.org/10.18653/v1/2020.acl-main.486} {Social bias frames: Reasoning about social and power implications of language}.
\newblock In \emph{Proceedings of the 58th Annual Meeting of the Association for Computational Linguistics}, pages 5477--5490, Online. Association for Computational Linguistics.

\bibitem[{Saxena et~al.(2024)Saxena, Fletcher, and Pechenizkiy}]{saxena2024fairsna}
Akrati Saxena, George Fletcher, and Mykola Pechenizkiy. 2024.
\newblock Fairsna: Algorithmic fairness in social network analysis.
\newblock \emph{ACM Computing Surveys}, 56(8):1--45.

\bibitem[{Schaeffer et~al.(2024)Schaeffer, Miranda, and Koyejo}]{schaeffer2024emergent}
Rylan Schaeffer, Brando Miranda, and Sanmi Koyejo. 2024.
\newblock Are emergent abilities of large language models a mirage?
\newblock \emph{Advances in Neural Information Processing Systems}, 36.

\bibitem[{Scheuneman(1979)}]{scheuneman1979method}
Janice Scheuneman. 1979.
\newblock A method of assessing bias in test items.
\newblock \emph{Journal of Educational Measurement}, pages 143--152.

\bibitem[{Schick et~al.(2021)Schick, Udupa, and Sch{\"u}tze}]{schick2021self}
Timo Schick, Sahana Udupa, and Hinrich Sch{\"u}tze. 2021.
\newblock Self-diagnosis and self-debiasing: A proposal for reducing corpus-based bias in nlp.
\newblock \emph{Transactions of the Association for Computational Linguistics}, 9:1408--1424.

\bibitem[{Schramowski et~al.(2022)Schramowski, Turan, Andersen, Rothkopf, and Kersting}]{schramowski2022large}
Patrick Schramowski, Cigdem Turan, Nico Andersen, Constantin~A Rothkopf, and Kristian Kersting. 2022.
\newblock Large pre-trained language models contain human-like biases of what is right and wrong to do.
\newblock \emph{Nature Machine Intelligence}, 4(3):258--268.

\bibitem[{Schr{\"o}der et~al.(2023)Schr{\"o}der, Schulz, Tarakanov, Feldhans, and Hammer}]{schroder2023measuring}
Sarah Schr{\"o}der, Alexander Schulz, Ivan Tarakanov, Robert Feldhans, and Barbara Hammer. 2023.
\newblock Measuring fairness with biased data: A case study on the effects of unsupervised data in fairness evaluation.
\newblock In \emph{International Work-Conference on Artificial Neural Networks}, pages 134--145. Springer.

\bibitem[{Schuhmann et~al.(2022)Schuhmann, Beaumont, Vencu, Gordon, Wightman, Cherti, Coombes, Katta, Mullis, Wortsman et~al.}]{schuhmann2022laion}
Christoph Schuhmann, Romain Beaumont, Richard Vencu, Cade Gordon, Ross Wightman, Mehdi Cherti, Theo Coombes, Aarush Katta, Clayton Mullis, Mitchell Wortsman, et~al. 2022.
\newblock Laion-5b: An open large-scale dataset for training next generation image-text models.
\newblock \emph{Advances in Neural Information Processing Systems}, 35:25278--25294.

\bibitem[{Schuhmann et~al.(2021)Schuhmann, Vencu, Beaumont, Kaczmarczyk, Mullis, Katta, Coombes, Jitsev, and Komatsuzaki}]{schuhmann2021laion}
Christoph Schuhmann, Richard Vencu, Romain Beaumont, Robert Kaczmarczyk, Clayton Mullis, Aarush Katta, Theo Coombes, Jenia Jitsev, and Aran Komatsuzaki. 2021.
\newblock Laion-400m: Open dataset of clip-filtered 400 million image-text pairs.
\newblock \emph{arXiv preprint arXiv:2111.02114}.

\bibitem[{Selvaraju et~al.(2017)Selvaraju, Cogswell, Das, Vedantam, Parikh, and Batra}]{8237336}
Ramprasaath~R. Selvaraju, Michael Cogswell, Abhishek Das, Ramakrishna Vedantam, Devi Parikh, and Dhruv Batra. 2017.
\newblock \href {https://doi.org/10.1109/ICCV.2017.74} {Grad-cam: Visual explanations from deep networks via gradient-based localization}.
\newblock In \emph{2017 IEEE International Conference on Computer Vision (ICCV)}, pages 618--626.

\bibitem[{Serapio-Garc{\'\i}a et~al.(2023)Serapio-Garc{\'\i}a, Safdari, Crepy, Sun, Fitz, Romero, Abdulhai, Faust, and Matari{\'c}}]{serapio2023personality}
Greg Serapio-Garc{\'\i}a, Mustafa Safdari, Cl{\'e}ment Crepy, Luning Sun, Stephen Fitz, Peter Romero, Marwa Abdulhai, Aleksandra Faust, and Maja Matari{\'c}. 2023.
\newblock Personality traits in large language models.
\newblock \emph{arXiv preprint arXiv:2307.00184}.

\bibitem[{Serna et~al.(2022)Serna, Morales, Fierrez, and Obradovich}]{serna2022sensitive}
Ignacio Serna, Aythami Morales, Julian Fierrez, and Nick Obradovich. 2022.
\newblock Sensitive loss: Improving accuracy and fairness of face representations with discrimination-aware deep learning.
\newblock \emph{Artificial Intelligence}, 305:103682.

\bibitem[{Seth et~al.(2023)Seth, Hemani, and Agarwal}]{10204258}
Ashish Seth, Mayur Hemani, and Chirag Agarwal. 2023.
\newblock \href {https://doi.org/10.1109/CVPR52729.2023.00659} {Dear: Debiasing vision-language models with additive residuals}.
\newblock In \emph{2023 IEEE/CVF Conference on Computer Vision and Pattern Recognition (CVPR)}, pages 6820--6829.

\bibitem[{Shaikh et~al.(2023)Shaikh, Zhang, Held, Bernstein, and Yang}]{shaikh-etal-2023-second}
Omar Shaikh, Hongxin Zhang, William Held, Michael Bernstein, and Diyi Yang. 2023.
\newblock \href {https://doi.org/10.18653/v1/2023.acl-long.244} {On second thought, let{'}s not think step by step! bias and toxicity in zero-shot reasoning}.
\newblock In \emph{Proceedings of the 61st Annual Meeting of the Association for Computational Linguistics (Volume 1: Long Papers)}, pages 4454--4470, Toronto, Canada. Association for Computational Linguistics.

\bibitem[{Sheng et~al.(2021)Sheng, Chang, Natarajan, and Peng}]{sheng-etal-2021-societal}
Emily Sheng, Kai-Wei Chang, Prem Natarajan, and Nanyun Peng. 2021.
\newblock \href {https://doi.org/10.18653/v1/2021.acl-long.330} {Societal biases in language generation: Progress and challenges}.
\newblock In \emph{Proceedings of the 59th Annual Meeting of the Association for Computational Linguistics and the 11th International Joint Conference on Natural Language Processing (Volume 1: Long Papers)}, pages 4275--4293, Online. Association for Computational Linguistics.

\bibitem[{Sheng et~al.(2019)Sheng, Chang, Natarajan, and Peng}]{sheng-etal-2019-woman}
Emily Sheng, Kai-Wei Chang, Premkumar Natarajan, and Nanyun Peng. 2019.
\newblock \href {https://doi.org/10.18653/v1/D19-1339} {The woman worked as a babysitter: On biases in language generation}.
\newblock In \emph{Proceedings of the 2019 Conference on Empirical Methods in Natural Language Processing and the 9th International Joint Conference on Natural Language Processing (EMNLP-IJCNLP)}, pages 3407--3412, Hong Kong, China. Association for Computational Linguistics.

\bibitem[{Sinha et~al.(2021)Sinha, Jia, Hupkes, Pineau, Williams, and Kiela}]{sinha2021masked}
Koustuv Sinha, Robin Jia, Dieuwke Hupkes, Joelle Pineau, Adina Williams, and Douwe Kiela. 2021.
\newblock Masked language modeling and the distributional hypothesis: Order word matters pre-training for little.
\newblock \emph{arXiv preprint arXiv:2104.06644}.

\bibitem[{Smith et~al.(2022{\natexlab{a}})Smith, Hall, Kambadur, Presani, and Williams}]{smith-etal-2022-im}
Eric~Michael Smith, Melissa Hall, Melanie Kambadur, Eleonora Presani, and Adina Williams. 2022{\natexlab{a}}.
\newblock \href {https://doi.org/10.18653/v1/2022.emnlp-main.625} {{``}{I}{'}m sorry to hear that{''}: Finding new biases in language models with a holistic descriptor dataset}.
\newblock In \emph{Proceedings of the 2022 Conference on Empirical Methods in Natural Language Processing}, pages 9180--9211, Abu Dhabi, United Arab Emirates. Association for Computational Linguistics.

\bibitem[{Smith et~al.(2022{\natexlab{b}})Smith, Patwary, Norick, LeGresley, Rajbhandari, Casper, Liu, Prabhumoye, Zerveas, Korthikanti et~al.}]{smith2022using}
Shaden Smith, Mostofa Patwary, Brandon Norick, Patrick LeGresley, Samyam Rajbhandari, Jared Casper, Zhun Liu, Shrimai Prabhumoye, George Zerveas, Vijay Korthikanti, et~al. 2022{\natexlab{b}}.
\newblock Using deepspeed and megatron to train megatron-turing nlg 530b, a large-scale generative language model.
\newblock \emph{arXiv preprint arXiv:2201.11990}.

\bibitem[{Soares and Angelov(2019)}]{soares2019fair}
Eduardo Soares and Plamen Angelov. 2019.
\newblock Fair-by-design explainable models for prediction of recidivism.
\newblock \emph{arXiv preprint arXiv:1910.02043}.

\bibitem[{Solaiman and Dennison(2021)}]{solaiman2021process}
Irene Solaiman and Christy Dennison. 2021.
\newblock Process for adapting language models to society (palms) with values-targeted datasets.
\newblock \emph{Advances in Neural Information Processing Systems}, 34:5861--5873.

\bibitem[{Srinivasan et~al.(2021)Srinivasan, Raman, Chen, Bendersky, and Najork}]{10.1145/3404835.3463257}
Krishna Srinivasan, Karthik Raman, Jiecao Chen, Michael Bendersky, and Marc Najork. 2021.
\newblock \href {https://doi.org/10.1145/3404835.3463257} {Wit: Wikipedia-based image text dataset for multimodal multilingual machine learning}.
\newblock In \emph{Proceedings of the 44th International ACM SIGIR Conference on Research and Development in Information Retrieval}, SIGIR '21, page 2443–2449, New York, NY, USA. Association for Computing Machinery.

\bibitem[{Srinivasan and Bisk(2022)}]{srinivasan-bisk-2022-worst}
Tejas Srinivasan and Yonatan Bisk. 2022.
\newblock \href {https://doi.org/10.18653/v1/2022.gebnlp-1.10} {Worst of both worlds: Biases compound in pre-trained vision-and-language models}.
\newblock In \emph{Proceedings of the 4th Workshop on Gender Bias in Natural Language Processing (GeBNLP)}, pages 77--85, Seattle, Washington. Association for Computational Linguistics.

\bibitem[{Stoychev and Gunes(2022)}]{stoychev2022effect}
Samuil Stoychev and Hatice Gunes. 2022.
\newblock The effect of model compression on fairness in facial expression recognition.
\newblock In \emph{International Conference on Pattern Recognition}, pages 121--138. Springer.

\bibitem[{Sun et~al.(2019)Sun, Gaut, Tang, Huang, ElSherief, Zhao, Mirza, Belding, Chang, and Wang}]{sun-etal-2019-mitigating}
Tony Sun, Andrew Gaut, Shirlyn Tang, Yuxin Huang, Mai ElSherief, Jieyu Zhao, Diba Mirza, Elizabeth Belding, Kai-Wei Chang, and William~Yang Wang. 2019.
\newblock \href {https://doi.org/10.18653/v1/P19-1159} {Mitigating gender bias in natural language processing: Literature review}.
\newblock In \emph{Proceedings of the 57th Annual Meeting of the Association for Computational Linguistics}, pages 1630--1640, Florence, Italy. Association for Computational Linguistics.

\bibitem[{Swim et~al.(2001)Swim, Hyers, Cohen, and Ferguson}]{swim2001everyday}
Janet~K Swim, Lauri~L Hyers, Laurie~L Cohen, and Melissa~J Ferguson. 2001.
\newblock Everyday sexism: Evidence for its incidence, nature, and psychological impact from three daily diary studies.
\newblock \emph{Journal of Social issues}, 57(1):31--53.

\bibitem[{Talat et~al.(2022)Talat, N{\'e}v{\'e}ol, Biderman, Clinciu, Dey, Longpre, Luccioni, Masoud, Mitchell, Radev, Sharma, Subramonian, Tae, Tan, Tunuguntla, and Van Der~Wal}]{talat-etal-2022-reap}
Zeerak Talat, Aur{\'e}lie N{\'e}v{\'e}ol, Stella Biderman, Miruna Clinciu, Manan Dey, Shayne Longpre, Sasha Luccioni, Maraim Masoud, Margaret Mitchell, Dragomir Radev, Shanya Sharma, Arjun Subramonian, Jaesung Tae, Samson Tan, Deepak Tunuguntla, and Oskar Van Der~Wal. 2022.
\newblock \href {https://doi.org/10.18653/v1/2022.bigscience-1.3} {You reap what you sow: On the challenges of bias evaluation under multilingual settings}.
\newblock In \emph{Proceedings of BigScience Episode {\#}5 -- Workshop on Challenges {\&} Perspectives in Creating Large Language Models}, pages 26--41, virtual+Dublin. Association for Computational Linguistics.

\bibitem[{Tamkin et~al.(2021)Tamkin, Brundage, Clark, and Ganguli}]{tamkin2021understanding}
Alex Tamkin, Miles Brundage, Jack Clark, and Deep Ganguli. 2021.
\newblock Understanding the capabilities, limitations, and societal impact of large language models.
\newblock \emph{arXiv preprint arXiv:2102.02503}.

\bibitem[{Tan and Celis(2019)}]{tan2019assessing}
Yi~Chern Tan and L~Elisa Celis. 2019.
\newblock Assessing social and intersectional biases in contextualized word representations.
\newblock \emph{Advances in neural information processing systems}, 32.

\bibitem[{Tang et~al.(2021)Tang, Du, Li, Liu, Zou, and Hu}]{10.1145/3442381.3449950}
Ruixiang Tang, Mengnan Du, Yuening Li, Zirui Liu, Na~Zou, and Xia Hu. 2021.
\newblock \href {https://doi.org/10.1145/3442381.3449950} {Mitigating gender bias in captioning systems}.
\newblock In \emph{Proceedings of the Web Conference 2021}, WWW '21, page 633–645, New York, NY, USA. Association for Computing Machinery.

\bibitem[{Tay et~al.(2022)Tay, Woo, Hickman, Booth, and D’Mello}]{doi:10.1177/25152459211061337}
Louis Tay, Sang~Eun Woo, Louis Hickman, Brandon~M. Booth, and Sidney D’Mello. 2022.
\newblock \href {https://doi.org/10.1177/25152459211061337} {A conceptual framework for investigating and mitigating machine-learning measurement bias (mlmb) in psychological assessment}.
\newblock \emph{Advances in Methods and Practices in Psychological Science}, 5(1):25152459211061337.

\bibitem[{Teo et~al.(2024)Teo, Abdollahzadeh, and Cheung}]{teo2024measuring}
Christopher Teo, Milad Abdollahzadeh, and Ngai-Man~Man Cheung. 2024.
\newblock On measuring fairness in generative models.
\newblock \emph{Advances in Neural Information Processing Systems}, 36.

\bibitem[{Teubner et~al.(2023)Teubner, Flath, Weinhardt, van~der Aalst, and Hinz}]{teubner2023welcome}
Timm Teubner, Christoph~M Flath, Christof Weinhardt, Wil van~der Aalst, and Oliver Hinz. 2023.
\newblock Welcome to the era of chatgpt et al. the prospects of large language models.
\newblock \emph{Business \& Information Systems Engineering}, 65(2):95--101.

\bibitem[{Thapliyal et~al.(2022)Thapliyal, Pont~Tuset, Chen, and Soricut}]{thapliyal-etal-2022-crossmodal}
Ashish~V. Thapliyal, Jordi Pont~Tuset, Xi~Chen, and Radu Soricut. 2022.
\newblock \href {https://doi.org/10.18653/v1/2022.emnlp-main.45} {Crossmodal-3600: A massively multilingual multimodal evaluation dataset}.
\newblock In \emph{Proceedings of the 2022 Conference on Empirical Methods in Natural Language Processing}, pages 715--729, Abu Dhabi, United Arab Emirates. Association for Computational Linguistics.

\bibitem[{Thirunavukarasu et~al.(2023)Thirunavukarasu, Ting, Elangovan, Gutierrez, Tan, and Ting}]{thirunavukarasu2023large}
Arun~James Thirunavukarasu, Darren Shu~Jeng Ting, Kabilan Elangovan, Laura Gutierrez, Ting~Fang Tan, and Daniel Shu~Wei Ting. 2023.
\newblock Large language models in medicine.
\newblock \emph{Nature medicine}, 29(8):1930--1940.

\bibitem[{Thomee et~al.(2016)Thomee, Shamma, Friedland, Elizalde, Ni, Poland, Borth, and Li}]{thomee2016yfcc100m}
Bart Thomee, David~A Shamma, Gerald Friedland, Benjamin Elizalde, Karl Ni, Douglas Poland, Damian Borth, and Li-Jia Li. 2016.
\newblock Yfcc100m: The new data in multimedia research.
\newblock \emph{Communications of the ACM}, 59(2):64--73.

\bibitem[{Thoppilan et~al.(2022)Thoppilan, De~Freitas, Hall, Shazeer, Kulshreshtha, Cheng, Jin, Bos, Baker, Du et~al.}]{thoppilan2022lamda}
Romal Thoppilan, Daniel De~Freitas, Jamie Hall, Noam Shazeer, Apoorv Kulshreshtha, Heng-Tze Cheng, Alicia Jin, Taylor Bos, Leslie Baker, Yu~Du, et~al. 2022.
\newblock Lamda: Language models for dialog applications.
\newblock \emph{arXiv preprint arXiv:2201.08239}.

\bibitem[{Touvron et~al.(2023)Touvron, Martin, Stone, Albert, Almahairi, Babaei, Bashlykov, Batra, Bhargava, Bhosale et~al.}]{touvron2023llama}
Hugo Touvron, Louis Martin, Kevin Stone, Peter Albert, Amjad Almahairi, Yasmine Babaei, Nikolay Bashlykov, Soumya Batra, Prajjwal Bhargava, Shruti Bhosale, et~al. 2023.
\newblock Llama 2: Open foundation and fine-tuned chat models.
\newblock \emph{arXiv preprint arXiv:2307.09288}.

\bibitem[{Van~der Wal et~al.(2024)Van~der Wal, Bachmann, Leidinger, van Maanen, Zuidema, and Schulz}]{van2024undesirable}
Oskar Van~der Wal, Dominik Bachmann, Alina Leidinger, Leendert van Maanen, Willem Zuidema, and Katrin Schulz. 2024.
\newblock Undesirable biases in nlp: Addressing challenges of measurement.
\newblock \emph{Journal of Artificial Intelligence Research}, 79:1--40.

\bibitem[{Vanmassenhove et~al.(2018)Vanmassenhove, Hardmeier, and Way}]{vanmassenhove-etal-2018-getting}
Eva Vanmassenhove, Christian Hardmeier, and Andy Way. 2018.
\newblock \href {https://doi.org/10.18653/v1/D18-1334} {Getting gender right in neural machine translation}.
\newblock In \emph{Proceedings of the 2018 Conference on Empirical Methods in Natural Language Processing}, pages 3003--3008, Brussels, Belgium. Association for Computational Linguistics.

\bibitem[{Vaswani et~al.(2017)Vaswani, Shazeer, Parmar, Uszkoreit, Jones, Gomez, Kaiser, and Polosukhin}]{vaswani2017attention}
Ashish Vaswani, Noam Shazeer, Niki Parmar, Jakob Uszkoreit, Llion Jones, Aidan~N Gomez, {\L}ukasz Kaiser, and Illia Polosukhin. 2017.
\newblock Attention is all you need.
\newblock \emph{Advances in neural information processing systems}, 30.

\bibitem[{Vig et~al.(2020)Vig, Gehrmann, Belinkov, Qian, Nevo, Singer, and Shieber}]{vig2020investigating}
Jesse Vig, Sebastian Gehrmann, Yonatan Belinkov, Sharon Qian, Daniel Nevo, Yaron Singer, and Stuart Shieber. 2020.
\newblock Investigating gender bias in language models using causal mediation analysis.
\newblock \emph{Advances in neural information processing systems}, 33:12388--12401.

\bibitem[{Wambsganss et~al.(2022)Wambsganss, Swamy, Rietsche, and K{\"a}ser}]{wambsganss2022bias}
Thiemo Wambsganss, Vinitra Swamy, Roman Rietsche, and Tanja K{\"a}ser. 2022.
\newblock Bias at a second glance: A deep dive into bias for german educational peer-review data modeling.
\newblock \emph{arXiv preprint arXiv:2209.10335}.

\bibitem[{Wang et~al.(2021)Wang, Liu, and Wang}]{wang-etal-2021-gender}
Jialu Wang, Yang Liu, and Xin Wang. 2021.
\newblock \href {https://doi.org/10.18653/v1/2021.emnlp-main.151} {Are gender-neutral queries really gender-neutral? mitigating gender bias in image search}.
\newblock In \emph{Proceedings of the 2021 Conference on Empirical Methods in Natural Language Processing}, pages 1995--2008, Online and Punta Cana, Dominican Republic. Association for Computational Linguistics.

\bibitem[{Wang et~al.(2022{\natexlab{a}})Wang, Liu, and Wang}]{wang-etal-2022-assessing}
Jialu Wang, Yang Liu, and Xin Wang. 2022{\natexlab{a}}.
\newblock \href {https://doi.org/10.18653/v1/2022.findings-acl.211} {Assessing multilingual fairness in pre-trained multimodal representations}.
\newblock In \emph{Findings of the Association for Computational Linguistics: ACL 2022}, pages 2681--2695, Dublin, Ireland. Association for Computational Linguistics.

\bibitem[{Wang et~al.(2023{\natexlab{a}})Wang, Li, Chen, Cai, Zhu, Lin, Cao, Liu, Liu, and Sui}]{wang2023large}
Peiyi Wang, Lei Li, Liang Chen, Zefan Cai, Dawei Zhu, Binghuai Lin, Yunbo Cao, Qi~Liu, Tianyu Liu, and Zhifang Sui. 2023{\natexlab{a}}.
\newblock Large language models are not fair evaluators.
\newblock \emph{arXiv preprint arXiv:2305.17926}.

\bibitem[{Wang et~al.(2024)Wang, Wang, Zhou, Dong, Tan, and Li}]{wang2024ceb}
Song Wang, Peng Wang, Tong Zhou, Yushun Dong, Zhen Tan, and Jundong Li. 2024.
\newblock Ceb: Compositional evaluation benchmark for fairness in large language models.
\newblock \emph{arXiv preprint arXiv:2407.02408}.

\bibitem[{Wang et~al.(2023{\natexlab{b}})Wang, Bao, Dong, Bjorck, Peng, Liu, Aggarwal, Mohammed, Singhal, Som et~al.}]{wang2023image}
Wenhui Wang, Hangbo Bao, Li~Dong, Johan Bjorck, Zhiliang Peng, Qiang Liu, Kriti Aggarwal, Owais~Khan Mohammed, Saksham Singhal, Subhojit Som, et~al. 2023{\natexlab{b}}.
\newblock Image as a foreign language: Beit pretraining for vision and vision-language tasks.
\newblock In \emph{Proceedings of the IEEE/CVF Conference on Computer Vision and Pattern Recognition}, pages 19175--19186.

\bibitem[{Wang et~al.(2022{\natexlab{b}})Wang, He, Jin, Yang, Wang, and Qu}]{9552921}
Xingbo Wang, Jianben He, Zhihua Jin, Muqiao Yang, Yong Wang, and Huamin Qu. 2022{\natexlab{b}}.
\newblock \href {https://doi.org/10.1109/TVCG.2021.3114794} {M2lens: Visualizing and explaining multimodal models for sentiment analysis}.
\newblock \emph{IEEE Transactions on Visualization and Computer Graphics}, 28(1):802--812.

\bibitem[{Wang et~al.(2020)Wang, Qinami, Karakozis, Genova, Nair, Hata, and Russakovsky}]{wang2020towards}
Zeyu Wang, Klint Qinami, Ioannis~Christos Karakozis, Kyle Genova, Prem Nair, Kenji Hata, and Olga Russakovsky. 2020.
\newblock Towards fairness in visual recognition: Effective strategies for bias mitigation.
\newblock In \emph{Proceedings of the IEEE/CVF conference on computer vision and pattern recognition}, pages 8919--8928.

\bibitem[{Wang et~al.(2022{\natexlab{c}})Wang, Dong, Xue, Zhang, Chiu, Wei, and Ren}]{wang2022fairness}
Zhibo Wang, Xiaowei Dong, Henry Xue, Zhifei Zhang, Weifeng Chiu, Tao Wei, and Kui Ren. 2022{\natexlab{c}}.
\newblock Fairness-aware adversarial perturbation towards bias mitigation for deployed deep models.
\newblock In \emph{Proceedings of the IEEE/CVF conference on computer vision and pattern recognition}, pages 10379--10388.

\bibitem[{Webster et~al.(2020)Webster, Wang, Tenney, Beutel, Pitler, Pavlick, Chen, Chi, and Petrov}]{webster2020measuring}
Kellie Webster, Xuezhi Wang, Ian Tenney, Alex Beutel, Emily Pitler, Ellie Pavlick, Jilin Chen, Ed~Chi, and Slav Petrov. 2020.
\newblock Measuring and reducing gendered correlations in pre-trained models.
\newblock \emph{arXiv preprint arXiv:2010.06032}.

\bibitem[{Weidinger et~al.(2021)Weidinger, Mellor, Rauh, Griffin, Uesato, Huang, Cheng, Glaese, Balle, Kasirzadeh et~al.}]{weidinger2021ethical}
Laura Weidinger, John Mellor, Maribeth Rauh, Conor Griffin, Jonathan Uesato, Po-Sen Huang, Myra Cheng, Mia Glaese, Borja Balle, Atoosa Kasirzadeh, et~al. 2021.
\newblock Ethical and social risks of harm from language models.
\newblock \emph{arXiv preprint arXiv:2112.04359}.

\bibitem[{Weidinger et~al.(2022)Weidinger, Uesato, Rauh, Griffin, Huang, Mellor, Glaese, Cheng, Balle, Kasirzadeh et~al.}]{weidinger2022taxonomy}
Laura Weidinger, Jonathan Uesato, Maribeth Rauh, Conor Griffin, Po-Sen Huang, John Mellor, Amelia Glaese, Myra Cheng, Borja Balle, Atoosa Kasirzadeh, et~al. 2022.
\newblock Taxonomy of risks posed by language models.
\newblock In \emph{Proceedings of the 2022 ACM Conference on Fairness, Accountability, and Transparency}, pages 214--229.

\bibitem[{Weinberg(2022)}]{weinberg2022rethinking}
Lindsay Weinberg. 2022.
\newblock Rethinking fairness: An interdisciplinary survey of critiques of hegemonic ml fairness approaches.
\newblock \emph{Journal of Artificial Intelligence Research}, 74:75--109.

\bibitem[{Welbl et~al.(2021)Welbl, Glaese, Uesato, Dathathri, Mellor, Hendricks, Anderson, Kohli, Coppin, and Huang}]{welbl-etal-2021-challenges-detoxifying}
Johannes Welbl, Amelia Glaese, Jonathan Uesato, Sumanth Dathathri, John Mellor, Lisa~Anne Hendricks, Kirsty Anderson, Pushmeet Kohli, Ben Coppin, and Po-Sen Huang. 2021.
\newblock \href {https://doi.org/10.18653/v1/2021.findings-emnlp.210} {Challenges in detoxifying language models}.
\newblock In \emph{Findings of the Association for Computational Linguistics: EMNLP 2021}, pages 2447--2469, Punta Cana, Dominican Republic. Association for Computational Linguistics.

\bibitem[{West et~al.(2019)West, Whittaker, and Crawford}]{west2019discriminating}
Sarah~Myers West, Meredith Whittaker, and Kate Crawford. 2019.
\newblock Discriminating systems.
\newblock \emph{AI Now}, pages 1--33.

\bibitem[{Wick et~al.(2019)Wick, Tristan et~al.}]{wick2019unlocking}
Michael Wick, Jean-Baptiste Tristan, et~al. 2019.
\newblock Unlocking fairness: a trade-off revisited.
\newblock \emph{Advances in neural information processing systems}, 32.

\bibitem[{Wolfe and Caliskan(2022{\natexlab{a}})}]{wolfe2022american}
Robert Wolfe and Aylin Caliskan. 2022{\natexlab{a}}.
\newblock American== white in multimodal language-and-image ai.
\newblock In \emph{Proceedings of the 2022 AAAI/ACM Conference on AI, Ethics, and Society}, pages 800--812.

\bibitem[{Wolfe and Caliskan(2022{\natexlab{b}})}]{wolfe2022markedness}
Robert Wolfe and Aylin Caliskan. 2022{\natexlab{b}}.
\newblock Markedness in visual semantic ai.
\newblock In \emph{Proceedings of the 2022 ACM Conference on Fairness, Accountability, and Transparency}, pages 1269--1279.

\bibitem[{Wolfe et~al.(2023)Wolfe, Yang, Howe, and Caliskan}]{wolfe2023contrastive}
Robert Wolfe, Yiwei Yang, Bill Howe, and Aylin Caliskan. 2023.
\newblock Contrastive language-vision ai models pretrained on web-scraped multimodal data exhibit sexual objectification bias.
\newblock In \emph{Proceedings of the 2023 ACM Conference on Fairness, Accountability, and Transparency}, pages 1174--1185.

\bibitem[{Wright et~al.(2020)Wright, Shaikh, Park, Epperson, Ahmed, Pinel, Yang, and Chau}]{wright2020recast}
Austin~P Wright, Omar Shaikh, Haekyu Park, Will Epperson, Muhammed Ahmed, Stephane Pinel, Diyi Yang, and Duen~Horng Chau. 2020.
\newblock Recast: Interactive auditing of automatic toxicity detection models.
\newblock In \emph{Proceedings of the Eighth International Workshop of Chinese CHI}, pages 80--82.

\bibitem[{Xu et~al.(2024)Xu, Hou, Pang, Deng, Xu, Shen, and Cheng}]{10.1145/3626772.3657750}
Shicheng Xu, Danyang Hou, Liang Pang, Jingcheng Deng, Jun Xu, Huawei Shen, and Xueqi Cheng. 2024.
\newblock \href {https://doi.org/10.1145/3626772.3657750} {Invisible relevance bias: Text-image retrieval models prefer ai-generated images}.
\newblock In \emph{Proceedings of the 47th International ACM SIGIR Conference on Research and Development in Information Retrieval}, SIGIR '24, page 208–217, New York, NY, USA. Association for Computing Machinery.

\bibitem[{Xu et~al.(2020)Xu, White, Kalkan, and Gunes}]{xu2020investigating}
Tian Xu, Jennifer White, Sinan Kalkan, and Hatice Gunes. 2020.
\newblock Investigating bias and fairness in facial expression recognition.
\newblock In \emph{Computer Vision--ECCV 2020 Workshops: Glasgow, UK, August 23--28, 2020, Proceedings, Part VI 16}, pages 506--523. Springer.

\bibitem[{Xue et~al.(2023)Xue, Wang, Wei, Liu, Woo, and Kuo}]{xue2023bias}
Jintang Xue, Yun-Cheng Wang, Chengwei Wei, Xiaofeng Liu, Jonghye Woo, and C-C~Jay Kuo. 2023.
\newblock Bias and fairness in chatbots: An overview.
\newblock \emph{arXiv preprint arXiv:2309.08836}.

\bibitem[{Yan et~al.(2020)Yan, Huang, and Soleymani}]{yan2020mitigating}
Shen Yan, Di~Huang, and Mohammad Soleymani. 2020.
\newblock Mitigating biases in multimodal personality assessment.
\newblock In \emph{Proceedings of the 2020 International Conference on Multimodal Interaction}, pages 361--369.

\bibitem[{Yang et~al.(2022)Yang, Bai, Wang, Lin, Lin, Qin, and He}]{yang2022fine}
Aimin Yang, Qifeng Bai, Jigang Wang, Nankai Lin, Xiaotian Lin, Guanqiu Qin, and Junheng He. 2022.
\newblock A fine-grained social bias measurement framework for open-domain dialogue systems.
\newblock In \emph{CCF International Conference on Natural Language Processing and Chinese Computing}, pages 240--251. Springer.

\bibitem[{Yao et~al.(2023)Yao, Shah, Sun, Cho, and Huang}]{yao2023end}
Barry~Menglong Yao, Aditya Shah, Lichao Sun, Jin-Hee Cho, and Lifu Huang. 2023.
\newblock End-to-end multimodal fact-checking and explanation generation: A challenging dataset and models.
\newblock In \emph{Proceedings of the 46th International ACM SIGIR Conference on Research and Development in Information Retrieval}, pages 2733--2743.

\bibitem[{Yee et~al.(2021)Yee, Tantipongpipat, and Mishra}]{yee2021image}
Kyra Yee, Uthaipon Tantipongpipat, and Shubhanshu Mishra. 2021.
\newblock Image cropping on twitter: Fairness metrics, their limitations, and the importance of representation, design, and agency.
\newblock \emph{Proceedings of the ACM on Human-Computer Interaction}, 5(CSCW2):1--24.

\bibitem[{Yoo et~al.(2021)Yoo, Park, Kang, Lee, and Park}]{yoo-etal-2021-gpt3mix-leveraging}
Kang~Min Yoo, Dongju Park, Jaewook Kang, Sang-Woo Lee, and Woomyoung Park. 2021.
\newblock \href {https://doi.org/10.18653/v1/2021.findings-emnlp.192} {{GPT}3{M}ix: Leveraging large-scale language models for text augmentation}.
\newblock In \emph{Findings of the Association for Computational Linguistics: EMNLP 2021}, pages 2225--2239, Punta Cana, Dominican Republic. Association for Computational Linguistics.

\bibitem[{Yu et~al.(2023)Yu, Zhuang, Zhang, Meng, Ratner, Krishna, Shen, and Zhang}]{NEURIPS2023_ae9500c4}
Yue Yu, Yuchen Zhuang, Jieyu Zhang, Yu~Meng, Alexander~J Ratner, Ranjay Krishna, Jiaming Shen, and Chao Zhang. 2023.
\newblock \href {https://proceedings.neurips.cc/paper_files/paper/2023/file/ae9500c4f5607caf2eff033c67daa9d7-Paper-Datasets_and_Benchmarks.pdf} {Large language model as attributed training data generator: A tale of diversity and bias}.
\newblock In \emph{Advances in Neural Information Processing Systems}, volume~36, pages 55734--55784. Curran Associates, Inc.

\bibitem[{Yuan et~al.(2021)Yuan, Chen, Wang, Yu, Shi, Jiang, Tay, Feng, and Yan}]{yuan2021tokens}
Li~Yuan, Yunpeng Chen, Tao Wang, Weihao Yu, Yujun Shi, Zi-Hang Jiang, Francis~EH Tay, Jiashi Feng, and Shuicheng Yan. 2021.
\newblock Tokens-to-token vit: Training vision transformers from scratch on imagenet.
\newblock In \emph{Proceedings of the IEEE/CVF international conference on computer vision}, pages 558--567.

\bibitem[{Zafar et~al.(2017)Zafar, Valera, Gomez~Rodriguez, and Gummadi}]{zafar2017fairness}
Muhammad~Bilal Zafar, Isabel Valera, Manuel Gomez~Rodriguez, and Krishna~P Gummadi. 2017.
\newblock Fairness beyond disparate treatment \& disparate impact: Learning classification without disparate mistreatment.
\newblock In \emph{Proceedings of the 26th international conference on world wide web}, pages 1171--1180.

\bibitem[{Zayed et~al.(2024)Zayed, Mordido, Shabanian, Baldini, and Chandar}]{zayed2024fairness}
Abdelrahman Zayed, Gon{\c{c}}alo Mordido, Samira Shabanian, Ioana Baldini, and Sarath Chandar. 2024.
\newblock Fairness-aware structured pruning in transformers.
\newblock In \emph{Proceedings of the AAAI Conference on Artificial Intelligence}, volume~38, pages 22484--22492.

\bibitem[{Zerveas et~al.(2022)Zerveas, Rekabsaz, Cohen, and Eickhoff}]{zerveas2022mitigating}
George Zerveas, Navid Rekabsaz, Daniel Cohen, and Carsten Eickhoff. 2022.
\newblock Mitigating bias in search results through contextual document reranking and neutrality regularization.
\newblock In \emph{Proceedings of the 45th International ACM SIGIR Conference on Research and Development in Information Retrieval}, pages 2532--2538.

\bibitem[{Zhang et~al.(2018)Zhang, Lemoine, and Mitchell}]{zhang2018mitigating}
Brian~Hu Zhang, Blake Lemoine, and Margaret Mitchell. 2018.
\newblock Mitigating unwanted biases with adversarial learning.
\newblock In \emph{Proceedings of the 2018 AAAI/ACM Conference on AI, Ethics, and Society}, pages 335--340.

\bibitem[{Zhang et~al.(2024{\natexlab{a}})Zhang, Yu, Dong, Li, Su, Chu, and Yu}]{zhang-etal-2024-mm}
Duzhen Zhang, Yahan Yu, Jiahua Dong, Chenxing Li, Dan Su, Chenhui Chu, and Dong Yu. 2024{\natexlab{a}}.
\newblock \href {https://aclanthology.org/2024.findings-acl.738} {{MM}-{LLM}s: Recent advances in {M}ulti{M}odal large language models}.
\newblock In \emph{Findings of the Association for Computational Linguistics ACL 2024}, pages 12401--12430, Bangkok, Thailand and virtual meeting. Association for Computational Linguistics.

\bibitem[{Zhang et~al.(2022{\natexlab{a}})Zhang, Dullerud, Roth, Oakden-Rayner, Pfohl, and Ghassemi}]{pmlr-v174-zhang22a}
Haoran Zhang, Natalie Dullerud, Karsten Roth, Lauren Oakden-Rayner, Stephen Pfohl, and Marzyeh Ghassemi. 2022{\natexlab{a}}.
\newblock \href {https://proceedings.mlr.press/v174/zhang22a.html} {Improving the fairness of chest x-ray classifiers}.
\newblock In \emph{Proceedings of the Conference on Health, Inference, and Learning}, volume 174 of \emph{Proceedings of Machine Learning Research}, pages 204--233. PMLR.

\bibitem[{Zhang et~al.(2020)Zhang, Lu, Abdalla, McDermott, and Ghassemi}]{zhang2020hurtful}
Haoran Zhang, Amy~X Lu, Mohamed Abdalla, Matthew McDermott, and Marzyeh Ghassemi. 2020.
\newblock Hurtful words: quantifying biases in clinical contextual word embeddings.
\newblock In \emph{proceedings of the ACM Conference on Health, Inference, and Learning}, pages 110--120.

\bibitem[{Zhang et~al.(2024{\natexlab{b}})Zhang, Li, Zhang, Pu, Cahyono, Hu, Liu, Zhang, Yang, Li et~al.}]{zhang2024lmms}
Kaichen Zhang, Bo~Li, Peiyuan Zhang, Fanyi Pu, Joshua~Adrian Cahyono, Kairui Hu, Shuai Liu, Yuanhan Zhang, Jingkang Yang, Chunyuan Li, et~al. 2024{\natexlab{b}}.
\newblock Lmms-eval: Reality check on the evaluation of large multimodal models.
\newblock \emph{arXiv preprint arXiv:2407.12772}.

\bibitem[{Zhang et~al.(2022{\natexlab{b}})Zhang, Roller, Goyal, Artetxe, Chen, Chen, Dewan, Diab, Li, Lin et~al.}]{zhang2022opt}
Susan Zhang, Stephen Roller, Naman Goyal, Mikel Artetxe, Moya Chen, Shuohui Chen, Christopher Dewan, Mona Diab, Xian Li, Xi~Victoria Lin, et~al. 2022{\natexlab{b}}.
\newblock Opt: Open pre-trained transformer language models.
\newblock \emph{arXiv preprint arXiv:2205.01068}.

\bibitem[{Zhang et~al.(2022{\natexlab{c}})Zhang, Wang, and Sang}]{zhang2022counterfactually}
Yi~Zhang, Junyang Wang, and Jitao Sang. 2022{\natexlab{c}}.
\newblock Counterfactually measuring and eliminating social bias in vision-language pre-training models.
\newblock In \emph{Proceedings of the 30th ACM International Conference on Multimedia}, pages 4996--5004.

\bibitem[{Zhao et~al.(2017)Zhao, Wang, Yatskar, Ordonez, and Chang}]{zhao-etal-2017-men}
Jieyu Zhao, Tianlu Wang, Mark Yatskar, Vicente Ordonez, and Kai-Wei Chang. 2017.
\newblock \href {https://doi.org/10.18653/v1/D17-1323} {Men also like shopping: Reducing gender bias amplification using corpus-level constraints}.
\newblock In \emph{Proceedings of the 2017 Conference on Empirical Methods in Natural Language Processing}, pages 2979--2989, Copenhagen, Denmark. Association for Computational Linguistics.

\bibitem[{Zhao et~al.(2018{\natexlab{a}})Zhao, Wang, Yatskar, Ordonez, and Chang}]{zhao-etal-2018-gender}
Jieyu Zhao, Tianlu Wang, Mark Yatskar, Vicente Ordonez, and Kai-Wei Chang. 2018{\natexlab{a}}.
\newblock \href {https://doi.org/10.18653/v1/N18-2003} {Gender bias in coreference resolution: Evaluation and debiasing methods}.
\newblock In \emph{Proceedings of the 2018 Conference of the North {A}merican Chapter of the Association for Computational Linguistics: Human Language Technologies, Volume 2 (Short Papers)}, pages 15--20, New Orleans, Louisiana. Association for Computational Linguistics.

\bibitem[{Zhao et~al.(2018{\natexlab{b}})Zhao, Zhou, Li, Wang, and Chang}]{zhao-etal-2018-learning}
Jieyu Zhao, Yichao Zhou, Zeyu Li, Wei Wang, and Kai-Wei Chang. 2018{\natexlab{b}}.
\newblock \href {https://doi.org/10.18653/v1/D18-1521} {Learning gender-neutral word embeddings}.
\newblock In \emph{Proceedings of the 2018 Conference on Empirical Methods in Natural Language Processing}, pages 4847--4853, Brussels, Belgium. Association for Computational Linguistics.

\bibitem[{Zhou et~al.(2021)Zhou, Huang, Fries, Youssef, Amrhein, Chang, Banerjee, Rubin, Xing, Shah et~al.}]{zhou2021radfusion}
Yuyin Zhou, Shih-Cheng Huang, Jason~Alan Fries, Alaa Youssef, Timothy~J Amrhein, Marcello Chang, Imon Banerjee, Daniel Rubin, Lei Xing, Nigam Shah, et~al. 2021.
\newblock Radfusion: Benchmarking performance and fairness for multimodal pulmonary embolism detection from ct and ehr.
\newblock \emph{arXiv preprint arXiv:2111.11665}.

\bibitem[{Ziems et~al.(2024)Ziems, Held, Shaikh, Chen, Zhang, and Yang}]{ziems2024can}
Caleb Ziems, William Held, Omar Shaikh, Jiaao Chen, Zhehao Zhang, and Diyi Yang. 2024.
\newblock Can large language models transform computational social science?
\newblock \emph{Computational Linguistics}, 50(1):237--291.

\bibitem[{Zong et~al.(2022)Zong, Yang, and Hospedales}]{zong2022medfair}
Yongshuo Zong, Yongxin Yang, and Timothy Hospedales. 2022.
\newblock Medfair: Benchmarking fairness for medical imaging.
\newblock \emph{arXiv preprint arXiv:2210.01725}.

\end{thebibliography}
\bibliographystyle{acl_natbib}

\appendix

\section{Appendix}

\subsection{Google Scholar Papers}

\subsubsection{Multimodal\footnote{scholar.google.com/scholar?hl=en\&as\_sdt=0\%2C5\&q=Fairness+ and+Bias+in+Large+Multimodal+Models\&btnG=}}

\paragraph{\acrshort{ieee}}: \cite{wang2020towards}, 
\cite{niu2021counterfactual}, \cite{joshi2021review}, 
\cite{wang2022fairness}, \cite{10204258}, \cite{karkkainen2021fairface}, \cite{peng2022balanced},
\cite{cho2023dall}, \cite{9552921}, \cite{Niu_2021_CVPR}, \cite{Hirota_2022_CVPR}

\paragraph{Springer}: \cite{xu2020investigating}, \cite{pena2023human}, \cite{stoychev2022effect}, \cite{georgopoulos2021mitigating}, \cite{cai2024locating}, \cite{alwahaby2022evidence}, \cite{10.1007/978-3-030-58574-7_34}

\paragraph{\acrshort{acm}}: \cite{booth2021bias}, \cite{yan2020mitigating}, \cite{wolfe2022american}, \cite{goyal2022fairness}, \cite{cheong2024fairrefuse}, \cite{wolfe2022markedness}, \cite{liang2022foundations}, \cite{yao2023end}, \cite{hirota2022gender}, \cite{yee2021image}, \cite{rahman2020integrating}, \cite{10.1145/3494672}, \cite{10.1145/3626772.3657750}, \cite{10.1145/3442381.3449950}

\paragraph{arXiv}: \cite{luo2024bigbench}, \cite{zhou2021radfusion}, \cite{wang2024ceb}, \cite{birhane2021multimodal}, \cite{friedrich2023fair}, \cite{goyal2022vision}, \cite{zhang2024lmms}, \cite{rana2022emotion}, \cite{chen2024mj}, \cite{soares2019fair}, \cite{zong2022medfair}

\paragraph{PubMed}: \cite{liang2021multibench}

\paragraph{\acrshort{mdpi}}: \cite{ferrara2023fairness}, \cite{informatics11030057}

\paragraph{\acrshort{acl}}: \cite{wang-etal-2021-gender}, \cite{srinivasan-bisk-2022-worst}, \cite{li-etal-2021-vision}, \cite{zhang-etal-2024-mm}, \cite{ross-etal-2021-measuring}

\paragraph{Nature}: \cite{meng2022interpretability}, \cite{roosli2022peeking}, \cite{fei2022towards}

\paragraph{\acrshort{plos}}: \cite{jiang2024evaluating}

\paragraph{\acrshort{spie}}: \cite{drukker2023toward}

\paragraph{De Gruyter}: \cite{jenks2024communicating}

\paragraph{\acrshort{neurips}}: \cite{gadre2024datacomp}, \cite{liang2022mind}, \cite{koh2024generating}, \cite{NEURIPS2023_dd83eada}

\paragraph{Elsevier}: \cite{rahate2022multimodal}, \cite{serna2022sensitive}, \cite{KUMARI2021115412}

\paragraph{\acrshort{pkp}}: \cite{jaiswal2020privacy}

\paragraph{\acrshort{mlrp}}: \cite{pmlr-v32-kiros14}, \cite{pmlr-v174-zhang22a}

\paragraph{Wiley}: \cite{ntoutsi2020bias}

\paragraph{Sage}: \cite{doi:10.1177/25152459211061337}

\subsubsection{Language\footnote{scholar.google.com/scholar?hl=en\&as\_sdt=0\%2C5\&q= Fairness+and+Bias+in+Large+Language+Models\&btnG=}}

\paragraph{\acrshort{neurips}}: \cite{ma2023fairness}, \cite{vig2020investigating}, \cite{kojima2022large}, \cite{solaiman2021process}, \cite{wick2019unlocking}, \cite{tan2019assessing}, \cite{ouyang2022training}, \cite{schaeffer2024emergent}, \cite{meng2022generating}, \cite{brown2020language}, \cite{NEURIPS2023_ae9500c4}

\paragraph{\acrshort{acm}}: \cite{bender2021dangers}, \cite{chang2024survey}, \cite{dhamala2021bold}, \cite{weidinger2022taxonomy}, \cite{mehrabi2021survey}, \cite{garg2019counterfactual}, \cite{ganguli2022predictability}, \cite{min2023recent}, \cite{guo2021detecting}, \cite{zhang2020hurtful}, \cite{dodge2019explaining}, \cite{navigli2023biases},
\cite{kotek2023gender}

\paragraph{\acrshort{pkp}}: \cite{pryzant2020automatically}

\paragraph{\acrshort{mit}}: \cite{ziems2024can}, \cite{jiang2021can}, \cite{schick2021self}, \cite{raza2024fair}

\paragraph{Springer}: \cite{freiberger2024fairness}, \cite{meyer2023chatgpt}, \cite{teubner2023welcome}, \cite{hou2024large}, \cite{lu2020gender}

\paragraph{\acrshort{mlrp}}: \cite{liang2021towards}, \cite{biderman2023pythia}, \cite{aher2023using}, \cite{lehman2023we}, \cite{liang2021towards}

\paragraph{\acrshort{jmlr}}: \cite{chung2024scaling}, \cite{chowdhery2023palm}

\paragraph{Cambridge}: \cite{argyle2023out}

\paragraph{arXiv}: \cite{ferrara2023should}, \cite{wang2023large}, \cite{tamkin2021understanding}, \cite{chen2021evaluating}, \cite{liang2022holistic}, 
\cite{weidinger2021ethical}, \cite{srivastava2023beyond}, \cite{rae2021scaling}, \cite{eloundou2023gpts}, \cite{serapio2023personality}, \cite{qi2023fine}, \cite{smith2022using}, \cite{huang2023look}, \cite{liu2023fingpt}, \cite{liu2023trustworthy}, \cite{hu2021lora}, \cite{zhang2022opt}, \cite{sinha2021masked}, \cite{thoppilan2022lamda}, \cite{menick2022teaching}, \cite{doan2024fairness},
\cite{weidinger2019ethical}

\paragraph{Elsevier}: \cite{kasneci2023chatgpt}, \cite{liu2023summary}, \cite{ray2023chatgpt}

\paragraph{\acrshort{acl}}: \cite{welbl-etal-2021-challenges-detoxifying}, \cite{nadeem-etal-2021-stereoset}, \cite{gehman-etal-2020-realtoxicityprompts}, \cite{delobelle-etal-2020-robbert}, \cite{sap-etal-2020-social}, \cite{park-etal-2018-reducing}, \cite{perez-etal-2022-red}, \cite{blodgett-etal-2020-language}, \cite{52742}, \cite{agrawal-etal-2022-large}, \cite{sun-etal-2019-mitigating}, \cite{liang-etal-2020-towards}, \cite{dinan-etal-2020-queens}, \cite{blodgett2021stereotyping}, \cite{chen2024quantifying}, \cite{gao2020making}, \cite{kurita-etal-2019-measuring}, \cite{yoo-etal-2021-gpt3mix-leveraging}, \cite{lin-etal-2022-truthfulqa}, \cite{hutchinson-etal-2020-social}, \cite{ramesh-etal-2023-fairness},
\cite{smith-etal-2022-im}, \cite{ruder-etal-2022-square}

\paragraph{Nature}: \cite{schramowski2022large}, \cite{clusmann2023future}, \cite{hager2024evaluation}, \cite{mesko2023imperative}, \cite{thirunavukarasu2023large}

\paragraph{\acrshort{plos}}: \cite{mozafari2020hate}

\paragraph{Wiley}: \cite{hovy2021five}

\paragraph{\acrshort{mdpi}}: \cite{garrido2021survey}

\paragraph{Preprints}: \cite{kumar2024bias}

\paragraph{PubMed}: \cite{karabacak2023embracing}

\paragraph{Academic Pinnacle}: \cite{desai2023large}

\vskip 0.4in

\subsection{\acrfull{wos} Papers}

\subsubsection{Multimodal\footnote{https://www.webofscience.com/wos/woscc/summary/889a9d49-d906-408c-91b6-ffefdff1880d-fed7fb8c/relevance/1}}

\paragraph{\acrshort{ieee}}: \cite{10204258}, \cite{10208759}, \cite{10336315}, \cite{9880269}

\paragraph{\acrshort{acm}}:
\cite{edenberg2023disambiguating}, \cite{10.1145/3577190.3614156}, \cite{alam2022college}

\paragraph{\acrshort{pkp}}: \cite{lui2024leveraging}

\subsubsection{Language\footnote{www.webofscience.com/wos/woscc/summary/79cd811f-beb4-40f2-844f-e0ca9edf1fe2-fed7f83e/relevance/1}}

\paragraph{\acrshort{ieee}}: \cite{10411546}, \cite{10480206}, \cite{10298519},
\cite{10388308}, \cite{10459082}

\paragraph{Elsevier}: \cite{NAZIR2024100619}, \cite{fan2020spatial}

\paragraph{\acrshort{acm}}: \cite{10.1145/3442442.3452325}, \cite{10.1145/3593013.3594078}, \cite{10.1145/3465416.3483299}, \cite{10.1145/3593013.3594109}, \cite{10.1145/3617694.3623257}, \cite{dhamala2021bold}, \cite{gadiraju2023wouldn}, \cite{saxena2024fairsna}, \cite{jamil2024equity}, \cite{zerveas2022mitigating}, \cite{wright2020recast}, \cite{giner2023datadoc}, \cite{raj2023true}

\paragraph{Springer}: \cite{dolci2023improving}, \cite{delobelle2022fairdistillation}, \cite{schroder2023measuring}, \cite{leteno2023investigation}, \cite{yang2022fine}, 

\paragraph{\acrshort{neurips}}: \cite{ma2023fairness}

\paragraph{Nature}: \cite{haltaufderheide2024ethics}

\paragraph{\acrshort{mdpi}}: \cite{bevara2024scaling}, \cite{pineiro2023ethical}, \cite{da2021could}

\paragraph{\acrshort{mlrp}}:
\cite{pmlr-v139-liang21a}

\paragraph{Now}: \cite{xue2023bias}

\paragraph{Wiley}: \cite{lee2024life}, \cite{hao2024transforming}

\paragraph{\acrshort{acl}}: \cite{kumar-etal-2023-parameter}, \cite{lauscher-etal-2021-sustainable-modular}, \cite{sheng-etal-2021-societal}, \cite{fatemi-etal-2023-improving}, \cite{guo-etal-2022-auto}, \cite{vanmassenhove-etal-2018-getting}, \cite{huang-etal-2020-reducing}, \cite{felkner-etal-2023-winoqueer}, \cite{talat-etal-2022-reap}, \cite{escude-font-costa-jussa-2019-equalizing}

\paragraph{\acrshort{aiaf}}: \cite{van2024undesirable}

\end{document}